\documentclass[journal]{IEEEtran}
\usepackage{graphicx}
\usepackage{epstopdf}
\usepackage{array}
\ifCLASSOPTIONcompsoc
  \usepackage[caption=false,font=normalsize,labelfont=sf,textfont=sf]{subfig}
\else
  \usepackage[caption=false,font=footnotesize]{subfig}
\fi
\usepackage{cite}
\usepackage{float}
\usepackage{bm,amssymb,amsthm,amsmath}
\usepackage[ruled,vlined]{algorithm2e}
\usepackage{enumerate}
\usepackage{color}
\definecolor{darkgreen}{rgb}{0, 0.5, 0}
\usepackage{url}
\usepackage{booktabs,colortbl}
\usepackage{mfirstuc}
\usepackage{multirow}
\usepackage{lettrine}
\usepackage{pxfonts,mathpazo}
\usepackage[export]{adjustbox}

\newcommand\etal{\textit{et al.}}
\newcommand\ie{\textit{i.e.}}
\newcommand\eg{\textit{e.g.}}
\newcommand\etc{\textit{etc.}}
\newcommand\algoabbr{CLIA}
\newcommand\clusteringprocessfullname{Cascade Clustering}

\newcommand\learningprocessfullname{Reference Point Incremental Learning}

\newcommand\titletext{A Many-Objective Evolutionary Algorithm With Two Interacting Processes: \clusteringprocessfullname{} and \learningprocessfullname{}}
\newcommand\scriptO{\mathcal{O}}

\begin{document}
\title{\textit{\titletext{}}}
\author{Hongwei~Ge$^{*}$, Mingde~Zhao$^{*}$,\\
        Liang Sun, Zhen Wang, Guozhen Tan, Qiang Zhang and C. L. Philip Chen
\thanks{$^{*}$Equal contribution. This is an UPDATED preprint of \url{https://ieeexplore.ieee.org/document/8485382}. Hongwei Ge, Liang Sun, Guozhen Tan and Qiang Zhang are with the College of Computer Science and Technology, Dalian University of Technology, Dalian, China; Mingde Zhao is in the School of Computer Science, McGill University and Mila (Montr\'eal Institute of Learning Algorithms, Qu\'ebec AI Institute), Montr\'eal, Canada; Z. Wang is with the School of Mathematical Sciences, Dalian University of Technology, Dalian, China; C. L. P. Chen is with the Department of Computer and Information Science, University of Macau, Macau, China.}
}
\markboth{UPDATED PREPRINT (20190820)}%
{GE AND ZHAO \etal{}: \titletext{}}
\maketitle
\begin{abstract}
Researches have shown difficulties in obtaining proximity while maintaining diversity for many-objective optimization problems. Complexities of the true Pareto front pose challenges for the reference vector-based algorithms for their insufficient adaptability to the diverse characteristics with no priors. This paper proposes a many-objective optimization algorithm with two interacting processes: \MakeLowercase{\clusteringprocessfullname{}} and \MakeLowercase{\learningprocessfullname{}} (\algoabbr{}). In the population selection process based on \MakeLowercase{\clusteringprocessfullname{}} (CC), using the reference vectors provided by the process based on incremental learning, the nondominated and the dominated individuals are clustered and sorted with different manners in a cascade style and are selected by round-robin for better proximity and diversity. In the reference vector adaptation process based on \MakeLowercase{\learningprocessfullname{}}, using the feedbacks from the process based on CC, proper distribution of reference points is gradually obtained by incremental learning. Experimental studies on several benchmark problems show that \algoabbr{} is competitive compared with the state-of-the-art algorithms and has impressive efficiency and versatility using only the interactions between the two processes without incurring extra evaluations.
\end{abstract}
\begin{IEEEkeywords}
Clustering, incremental machine learning, interacting processes, many-objective optimization, reference vector.
\end{IEEEkeywords}
\IEEEpeerreviewmaketitle
\bstctlcite{IEEEexample:BSTcontrol}
\section{Introduction}
\IEEEPARstart{M}{any} real-world problems involve optimization of conflicting objectives \cite{louafi2017multi,li2017quantum}. With the population-based features, multiobjective evolutionary algorithms (MOEAs) have shown promising performance within reasonable runtime by simultaneously evolving towards various parts of the true Pareto front (PF) \cite{zhang2015knee}.
\par
Aiming to find a representative subset of individuals of the true PF, MOEAs pursue two goals: 1) \emph{obtaining proximity} and 2) \emph{maintaining diversity}. Proximity ensures that the subset is close to the true PF in the objective space and diversity ensures that the finite individuals in the subset can well represent the distribution of the true PF in the entire extent. Traditional Pareto dominance-based algorithms such as NSGA-II \cite{deb2002fast} perform well with two or three objectives. However, their performance noticeably deteriorates with more involving objectives \cite{purshouse2007pareto, wang2017diversity}. In \cite{teytaud2007hardness}, it has been shown that the performance of a simply designed random search algorithm is asymptotically roughly equivalent to these algorithms with large number of objectives.
\par
The deterioration in performance greatly motivates the researchers to excavate new ideas 
to tackle many-objective optimization problems (MaOPs), since the real-world applications with many objectives appear widely in industry \cite{sabioni2018robust, cheng2017evolutionary}. These ideas can be summarized as \emph{dominance enhancement}, \emph{indicators}, and \emph{divide and conquer}.
\par
Algorithms based on the idea of \emph{dominance enhancement} try to increase the evolution pressure via modifications on the dominance relation and/or combination with new assistant mechanisms. The attempts for modification, including relaxing, controlling the Pareto dominance relation or devising new partial relations such as $\epsilon$-dominance \cite{hadka2013borg}, $\alpha$-dominance \cite{dai2016improved}, $\theta$-dominance \cite{yuan2016dominance}, $\epsilon$-box dominance \cite{kowatari2012study}, fuzzy-based Pareto dominance \cite{he2014fuzzy}, RP-dominance \cite{elarbi2017decomposition}, \etc{}, have been conducted. However, the way to determine the proper relaxation degree remains as an open issue \cite{li2013comparative}. To combine with the dominance relation, efforts for developing new assistant mechanisms including distance-based rank \cite{fabre2010two}, density adjustment strategies \cite{li2014shift}, clustering-ranking \cite{cai2015clustering}, \etc{}, have also been made. Yet researches have shown that the dominance-based approaches may push evolution toward one or several subspaces and may fail to produce solutions along the entire extent of the PF \cite{li2015bigoal}. Acknowledging this, researches such as average ranking \cite{li2010enhancing}, preference order rank \cite{wang2013preference} which aim to address such partial coverage have been investigated. 
\par
Quite different from the dominance-based ideas, the \emph{indicator}-based algorithms map the proximity and diversity of the population into designed indicators, such as hypervolume (HV) \cite{rostami2017fast, jiang2015simple}, averaged Hausdorff distance \cite{schutze2012averaged}, R2-indicator \cite{gomez2015improved} and IGD-NS indicator \cite{tian2017indicator}. On one hand, researches have shown that the extensive computation for the indicators in high-dimensional objective spaces still remains as a bottleneck \cite{cheng2016reference}, though some contributions have been made to alleviate the computational burden \cite{sun2018IGD, bader2011hype}. On the other hand, the designs of the mappings from the population to the indicators determine the capabilities of providing appropriate and sufficient evolution pressures. Designing of the effective indicators remains as a challenging task.
\par
The algorithms that employ the idea of \emph{divide-and-conquer} aim to downsize the scale of the problem and transform the original MaOP to smaller subproblems. Recognized as the reference vector \cite{cheng2016reference}-based algorithms, some MOEAs use reference vectors as weight vectors to aggregate the objectives into subproblems or use reference vectors as neighborhood axes to convert the original problem into subproblems of the corresponding subspaces \cite{cheng2016reference}. MOEA/D \cite{zhang2007MOEAD}, a representative MOEA of this kind, decomposes the original problem into scalar ones by using the reference vectors as weight vectors. In NSGA-III \cite{deb2013evolutionary}, another representative MOEA, the objective space is segmented into neighborhoods around the reference vectors and the environmental selection is conducted with the help of niching techniques. There are also many other interesting ideas including different kinds of region division approaches or other techniques to downsize the problem scale. In GrEA \cite{yang2013grid}, individuals are put into the subspaces of grids for ranking and selection. In NSGA-III-OSD \cite{bi2017improved}, the objective space is first divided into several high-level subspaces and the individuals are accordingly divided into the subpopulations. Inside these high-level subspaces, there are also low-level neighborhoods formed by the reference vectors. In \cite{yuan2018objective}, the dimension of the objective space of an original problem is reduced by solving designed multiobjective optimization problems. Overall, despite these MOEAs achieve encouraging performance, the approaches to convert the original problem to the subproblems, the way to deal the subproblems and the dependence on the quality of reference vectors or space segments are still major bottlenecks \cite{cheng2017benchmark}.
\par
Recently, the trends of hybridizing more than one of the aforementioned ideas emerge, with the purpose to exploit the merits of the employed ideas altogether. In MOEA/DD \cite{li2015evolutionary}, decomposition and Pareto dominance are together utilized and it has demonstrated superiority over many contemporary MOEAs. In Two\_Arch2 \cite{wang2015improved}, to obtain competitive performance, the convergence archive utilizes an indicator and the diversity archive is maintained by enhancing the Pareto dominance with $L_{1/M}$ norm. In \cite{dai2018indicator}, IREA makes use of the merits of the indicator and reference vectors to achieve competitive performance. Though encouraging performance has been achieved by the complementary contributions of the ideas, these hybrid MOEAs are still in a way haunted by the inherent bottlenecks of the strategies they have employed, such as the dependence on the quality of reference vectors and the extensive computational burdens.
\par
In this paper, we propose a dominance and divide-and-conquer-based MOEA with two interactive processes: \MakeLowercase{\clusteringprocessfullname{}} and \MakeLowercase{\learningprocessfullname{}} (\algoabbr{}). In the proposed selection operator of \MakeLowercase{\clusteringprocessfullname{}}, the nondominated individuals are guided by the reference vectors and the dominated ones are guided by the elites among the nondominated individuals. The clustering for the nondominated individuals proceeds first and the clustering for the dominated individuals succeeds in a cascade style. This process is designed to evenly distribute the individuals over the whole extent of the current PF while providing sufficient evolution pressure. In the proposed reference vector adaptation mechanism based on \MakeLowercase{\learningprocessfullname{}}, an incremental SVM is used to gradually provide better reference vectors to increase the versatility on diverse problems.
\par
The rest of this paper is organized as follows. Section II gives the reviews of the reference vector-based algorithms and the motivations of this paper. Section III gives the details of \algoabbr{} focusing on the two interacting processes. Section IV presents the experimental studies, which include investigating the characteristics and comparing the performance with state-of-the-art algorithms. Section V concludes this paper.
\section{Reference Vector-Based Algorithms}
Algorithms utilizing reference vectors have many attractive properties. One one hand, reference vectors could generally guide the evolution toward proximity and diversity. On the other hand, they are able to be articulated with user preferences \cite{cheng2016reference} or conveniently applied to complicated scenarios \cite{chugh2018surrogate}. 
In this section, we first review the reference vector-based MOEAs. Then we analyze the existing challenges and discuss the motivations of this paper.
\subsection{Representative Works}
Recently, many reference vector-based MOEAs have been contributed. MOEA/D is a representative reference vector-based MOEA designed for multiobjective optimization, which utilizes the reference vectors as the weight vectors to aggregate the proximity and the diversity into scalar values \cite{zhang2007MOEAD} and such approach is normally recognized as ``decomposition.'' Each solution is associated with a subproblem, and the subproblems are optimized collaboratively. Such design demonstrated superiority over many contemporary algorithms. NSGA-III is another representative algorithm based on reference vectors \cite{deb2013evolutionary}. It emphasizes individuals that are nondominated and close to the reference vectors. NSGA-III follows different framework compared to MOEA/D, where all individuals in the population are together selected and evolved, not separately. With such design, NSGA-III can achieve satisfactory results in short time. In \cite{cheng2016reference}, an efficient and effective algorithm RVEA is proposed, which uses the adaptive angle-penalized distance (APD) to decompose the original MaOP into scalar subproblems. RVEA achieves competitive performance among many reference vector-based state-of-the-art algorithms. Despite the encouraging performance, there are still two major challenges for the performance of the reference vector-based algorithms.
\subsection{Existing Challenges}
The first major challenge for the current reference vector-based algorithms is the decrease in evolution pressure in the high-dimensional objective spaces. Many researches have been conducted into devising better selection approaches that could in a way address the insufficient evolution pressure. Some efforts have been put into investigating better subproblem conversion approaches \cite{jiang2017scalarizing,cheng2016reference,wang2016decomposition,wang2018localized}. However in \cite{jiang2017scalarizing} and \cite{wang2018localized}, it is demonstrated that the current approaches may still be unsatisfactory since they may have problems with either proximity or diversity. Some efforts have also been made to enhance the evolution pressure by integrating new mechanisms. In \cite{li2015evolutionary}, the proposed algorithm MOEA/DD utilizes Pareto dominance and decomposition comprehensively. It has demonstrated superior performance given the properly set reference vectors. In \cite{dai2018indicator}, IREA utilizes the $I_{\epsilon+}$ indicator and the reference vectors together, demonstrating competitiveness among many state-of-the-art MOEAs. However, the increment in the complexity for MOEA/DD and IREA hybridizing decomposition with other ideas causes heavy computational burden. The problems for evolution pressure are still to be addressed.
\par
\begin{figure}[!t]
\centering
\captionsetup{justification = centering}
\includegraphics[width=0.26\textwidth]{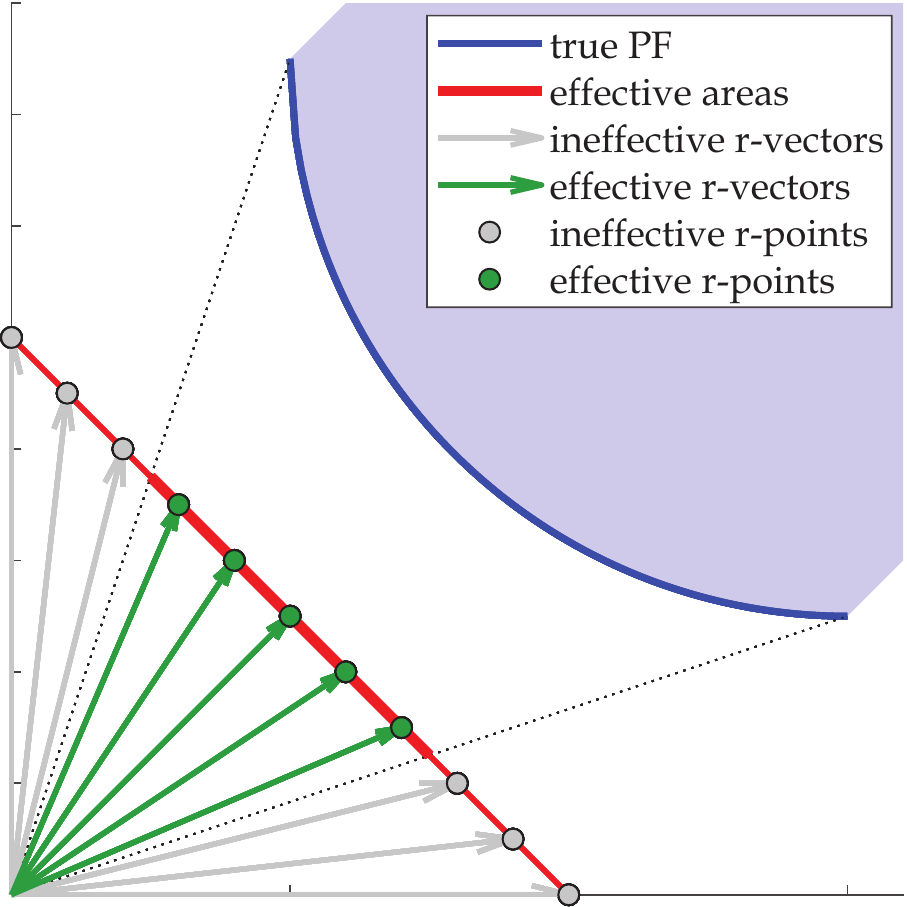}
\caption{Demonstration for the central projection of the true PF covering only a part of the unit simplex.}
\label{fig:projection}
\end{figure}

The second major challenge lies in the fact that the performance of these MOEAs is sensitive to the distribution of the reference vectors: they rely on the uniformity of reference vectors to ensure the quality of the subproblem construction, in some cases that could mean diversity. To assure such uniformity, the algorithms such as \cite{deb2013evolutionary} often employ predefined uniformly distributed reference vectors with the same amount as the population size, hoping that each of these reference vectors could ultimately guide the population to be distributed evenly over the true PF. However, this simple configuration is inappropriate for the following two reasons.
\begin{enumerate}
\item
Difference in Curvature: The curvature of the true PF may be irregular, \eg{} convex, concave or more complicated, so even if the reference vectors are originally uniformly generated, their intersection points with the true PF hypersurface may not be uniformly distributed \cite{cheng2016reference, tian2017indicator}.
\item
Partial PFs: In the problems with a certain characteristic, the true PFs are not wide-spread to fully cover the hypersurface they are located.
\end{enumerate}
The characteristic can be captured by employing a central projection of the true PF onto the unit simplex as demonstrated in Fig. \ref{fig:projection}. If the unit simplex is just partially covered by the projection\footnote{For convenience, we call this type of PF ``partially spread'' or ``partial' and the other type ``fully spread'' or ``full.''}, the aforementioned generation strategy may lead the individuals to coincide on certain points on the true PF and thus cause decrease in diversity. To apply reference vector-based MOEAs on the real-world occasions with diverse characteristics, many efforts have been contributed to prior-free reference vector adaptation. Some methods try to replace the noncontributing reference vectors. In A-NSGA-III \cite{jain2014evolutionary}, the undesirable reference vectors are continuously deleted, and new reference vectors are added according to the distribution of contemporary individuals along the whole evolutionary process. However, catering to the contemporary individuals in the objective space may have the risks of evolving toward a part of the true PF, not the entire extent. In RVEA* \cite{cheng2016reference}, an additional reference set is adjusted adaptively to handle the irregularities of the true PF. Such additional set always disturbs the uniformity of the reference vectors and may not lead to ideal diversity. In MOEA/D-AWA \cite{qi2014adaptive}, an adaptation based on deleting the overcrowded subproblems and inserting new subproblems in the sparse areas is used. Such insertion may not satisfactorily adapt to the problems with disconnected PFs. There are also many other interesting ideas. In PICEA-w \cite{wang2013weights}, a set of reference vectors co-evolves with the population during the evolutionary process. Nevertheless, such co-evolution is yet to be proved viable on the many-objective problems. PaS \cite{wang2016decomposition} avoids the estimations of the PF curvature and adjusts the way of decomposition by adaptively changing the scalarization method. However, the adaptation for the curvatures of the PF cannot address the deterioration of performance caused by the partially spread PFs. Besides, the pervasive adaptation methods neglect the identification for the appropriate moments for adaptation and thus may disturb the uniformity of reference vectors on the problems with regular PFs. Such disturbance may ultimately cause degradation in performance \cite{cheng2016reference,tian2017indicator}. Anyhow, the existing reference vector adaptation methods still can be improved to adapt to the diverse PFs more satisfactorily. 
\subsection{Motivation}
To address the challenges, we devise a reference vector-based algorithm with a selection operator that is expected to provide sufficient evolution pressure. Also, a reference vector adaptation strategy should be devised which does not disturb the uniformity of reference vectors when initiated and provides satisfactory versatility. And importantly, these two processes should cooperate well.
\par
The merits of the current reference vector-based state-of-the-art algorithms motivate us to make use of the knowledge in the related researches. Besides pursuing the quality of the selected population, the wide spread of the selected population should be considered, since in \cite{jain2014evolutionary} and \cite{cheng2016reference}, it is indicated that the wide-spread populations are not only beneficial for the population diversity but also for the reference vector adaptation. Considering all the points, in this paper, we propose a selection operator of CC, which is featured with two levels of clusterings, as the first level clustering for the nondominated individuals determines the second level clustering for the dominated individuals. The two levels of individuals are selected under different criteria. A round-robin picking is embedded to balance the number of individuals near each reference vector for the wide-spread of the population.
\par
The existing adaptation methods change the local density of reference points on the unit simplex. Such changes may not ensure the versatility and may also disturb the uniformity \cite{cheng2016reference,tian2017indicator}. Moreover, the frequent periodic adaptations seem not appropriate, for they cannot judge the need for the adaptations, especially for the problems with full PFs where adaptations may push the population toward a part of the true PF \cite{wang2017effect,liu2017reference}. Thus, it is desirable to devise an adaptation method that initiates only when adjustments are needed and hardly disturbs the uniformity. In this paper, a status sampler is designed to capture the appropriate moments for adaptation and incremental learning is used to capture the effective areas for reference point generation. Such adaptation always generates uniformly distributed reference points and reduces the ineffective ones based on the accumulated knowledge. With the coordination of the proposed mechanisms, the problems of the disturbance for diversity should be in a way addressed.
\section{\algoabbr{}}
In this section, \algoabbr{} will be explained with focus on the two interacting processes. The first process is the selection operator based on \MakeLowercase{\clusteringprocessfullname{}}, employed to guide the evolution of the individuals using the reference vectors provided by the second process. The second is the reference vector adaptation mechanism based on \MakeLowercase{\learningprocessfullname{}}, employed to generate properly distributed reference vectors using the feedbacks from the first process.
\subsection{\clusteringprocessfullname{}}
In this subsection, we present the details of the proposed selection process named \clusteringprocessfullname{} (CC).
\subsubsection{Frontier Individual Identification}
Instead of employing Non-Dominated Sort (NDSort) \cite{tian2017effectiveness} fully, a \emph{frontier individual identification} mechanism is employed, which terminates the NDsort when the first front is identified, thus only picking out the first front as the \emph{frontier individuals}.
\begin{algorithm}[!t]
\small
\caption{CC}
\label{alg:clustering}
\KwIn{$Z$ (set of reference vectors), $P$ (potential population), $N$ (population size for the next generation)}
\KwOut{$P$ (population for the next generation)}

\textcolor{darkgreen}{//Frontier Solution Identification}\\

$[F, NF] \leftarrow $ FS\_Identification($P$)\;

\textcolor{darkgreen}{//Attach frontiers to reference vectors, return the clusters}\\
$C \leftarrow $Attach($F$, $Z$, 'point2vector')\;

\For{each cluster $c_i \in C$}{
    \For{each frontier $\bm{f_j}$ in $c_i$}{
        $PDM(\bm{f_j}) \leftarrow mean(\bm{f_j}) + sin(\bm{z_i}, \bm{f_j})$;
    }
    $c_i.F \leftarrow $ sort($c_i.F$, $PDM(c_i.F)$, ascend)\;
    Pick out ${c}_i.{f}_j$ with the smallest PDM as $c_i.center$;
}

\textcolor{darkgreen}{//Attach nonfrontiers to clusters}\\

$C  \leftarrow $Attach($NF$, $C$, 'point2center')\;

\For{each cluster $c_i \in C$}{
    $c_i.NF \leftarrow$ sort($c_i.NF$, $d(c_i.NF, c_i.center)$, ascend)\;
    Create selection queue $c_i.S \leftarrow \langle c_i.F, c_i.NF \rangle$\;
}
\textcolor{darkgreen}{//Round-robin Picking}\\
$i \leftarrow 1$\;
$P \leftarrow \emptyset$\;
\While{$|P| < N$}{
    $P \leftarrow P \cup Pop(c_i.S)$\;
    $i \leftarrow mod(i, |C|) + 1$;
}
\end{algorithm}
\subsubsection{Bi-level Clustering and Intraclass Sorting}
Each frontier individual is attached to its nearest reference vector by calculating the sine values of their included angles with each reference vector. Reference vectors with frontier individuals attached are recognized as \emph{active reference vectors}. The frontier individuals attached to the same reference vector are intraclassly sorted using a proximity and diversity metric (PDM) in ascending order.
\begin{equation}
\label{formula:PDM}
\begin{split}
PDM(\bm{o},\bm{z}) & \equiv PM(\bm{o}) + DM(\bm{o},\bm{z}) \equiv mean(\bm{o}) + \alpha {\|\bm{o}\|}_2 sin(\bm{o}, \bm{z})\ \\
& \equiv \frac{\bm{o}^T \bm{1}}{M} + \alpha\frac{\sqrt{{\|\bm{o}\|}_2^2 {\|\bm{z}\|}_2^2 - {(\bm{o}^T \bm{z})} ^ 2}}{{\|\bm{z}\|}_2}
\end{split}
\end{equation}
where $\bm{o}$ is the frontier individual, $\bm{1}$ is the all-one vector, $\bm{z}$ is the nearest reference vector to $\bm{o}$, and $\alpha$ is a coefficient leveraging proximity and diversity and is chosen to be $5$ as it is in PBI \cite{zhang2007MOEAD}. It should be noticed that the actual calculation is not such complex as the formula, for $sin(\bm{o},\bm{z})$ is pre-calculated while attaching the frontiers to the reference vectors.
\par
$PM(\bm{o})$ is a term designed to reflect the proximity of an individual. It calculates the mean of the values on each objective. The smaller the $PM(\bm{o})$, the closer the frontier individual is to the true PF. Different from the similar proximity term used in BiGE \cite{li2015bigoal}, we avoid the changes of the relative positions between reference vectors and the current PF caused by normalization \cite{cheng2016reference} that have influences on the process of \MakeLowercase{\learningprocessfullname{}} which is to be discussed later. $DM(\bm{o}, \bm{z})$ expresses the distance from a frontier $\bm{o}$ to the nearest reference line $\bm{z}$ is located on, which represents the distribution error between $\bm{z}$ and $\bm{o}$. The smaller the $DM(\bm{o}, \bm{z})$, the closer the frontier individual $\bm{o}$ is to the reference vector $\bm{z}$. So, by using properly distributed reference vectors, $DM(\bm{o}, \bm{z})$ guides the evolution of individuals toward better diversity.
\begin{figure*}[!t]
\centering
\subfloat[][PBI proximity term]{
\captionsetup{justification = centering}
\includegraphics[width=0.18\textwidth]{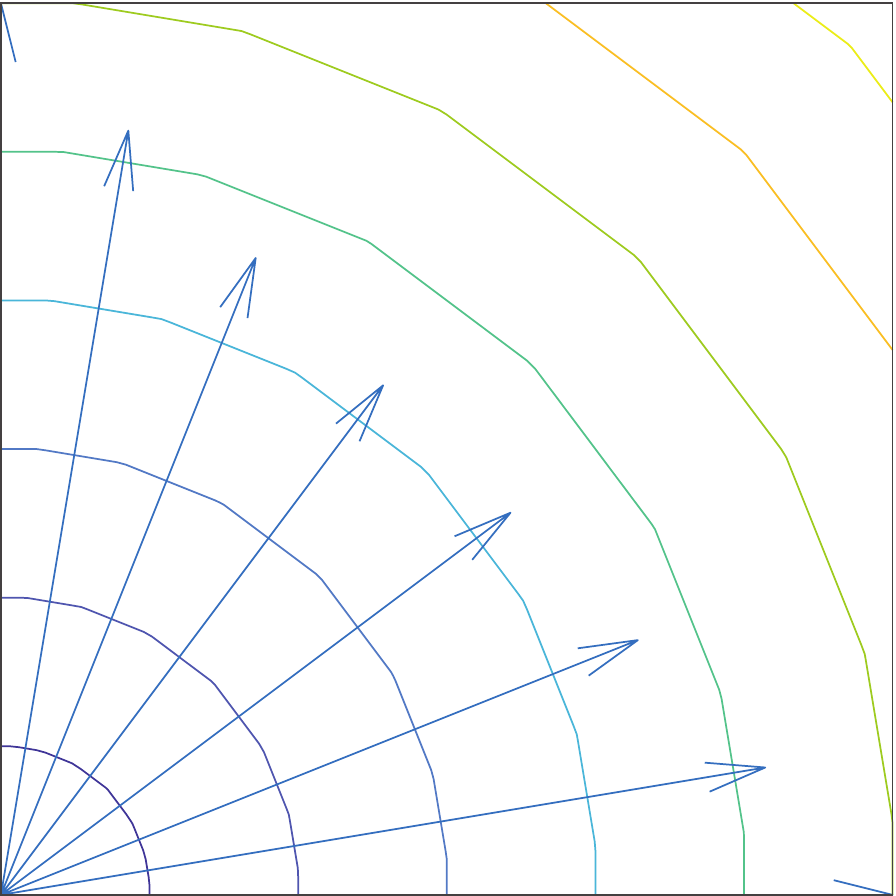}}%
\subfloat[PM, proximity term]{
\captionsetup{justification = centering}
\includegraphics[width=0.18\textwidth]{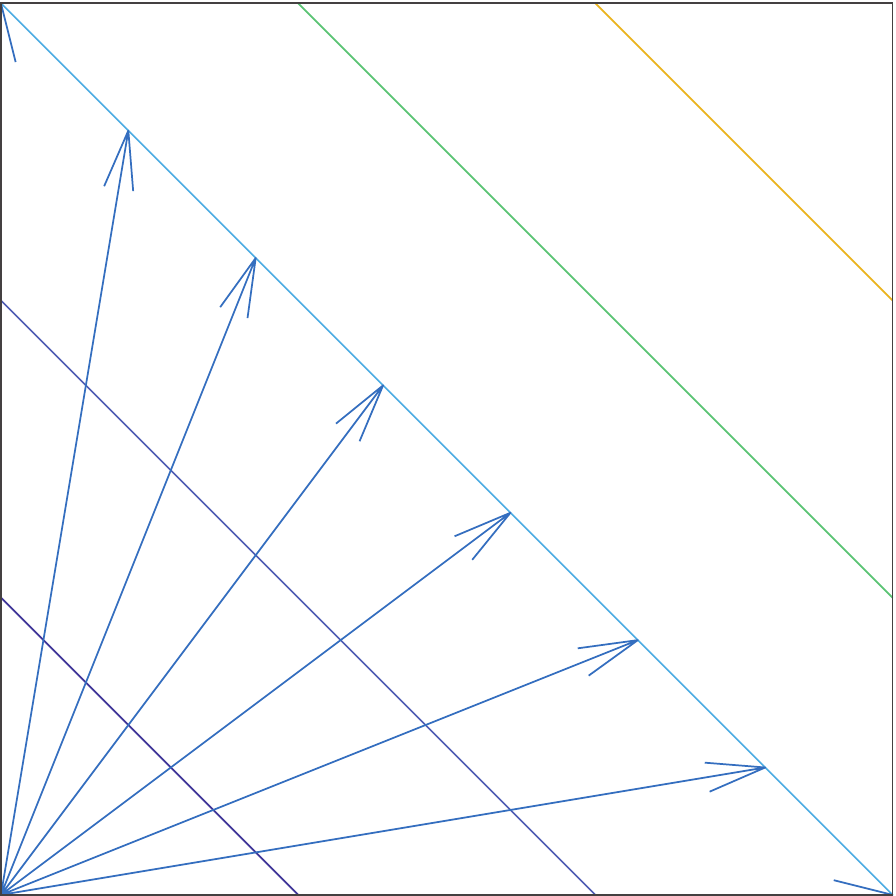}}%
\subfloat[DM, shared diversity term]{
\captionsetup{justification = centering}
\includegraphics[width=0.18\textwidth]{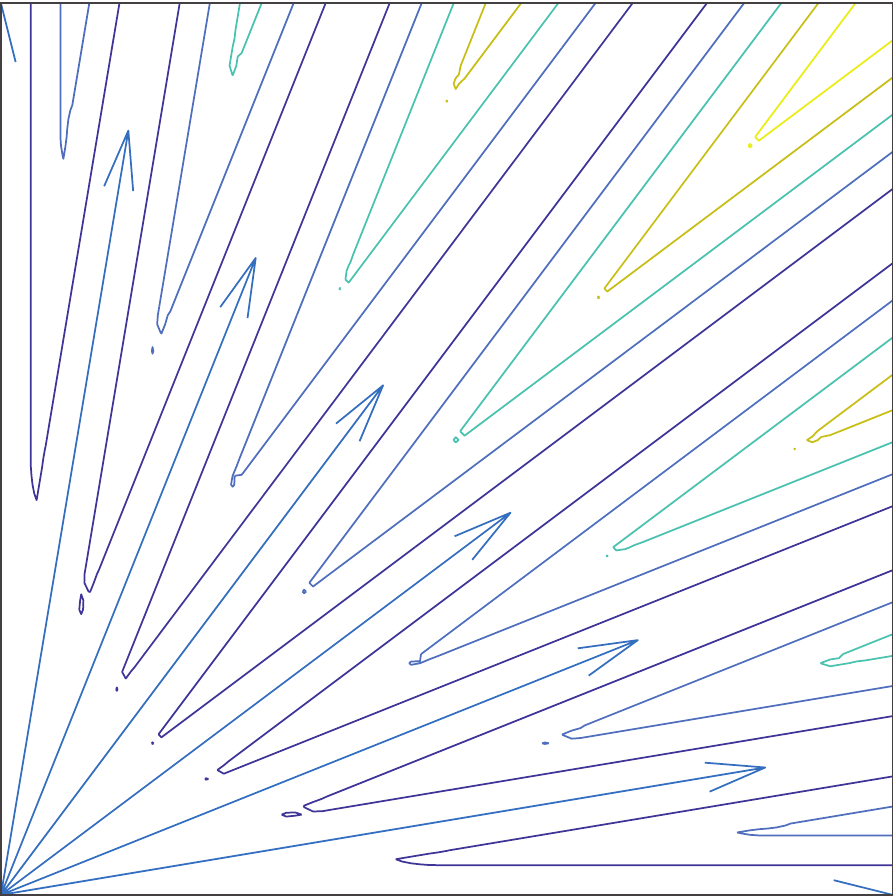}}%
\subfloat[PBI]{
\captionsetup{justification = centering}
\includegraphics[width=0.18\textwidth]{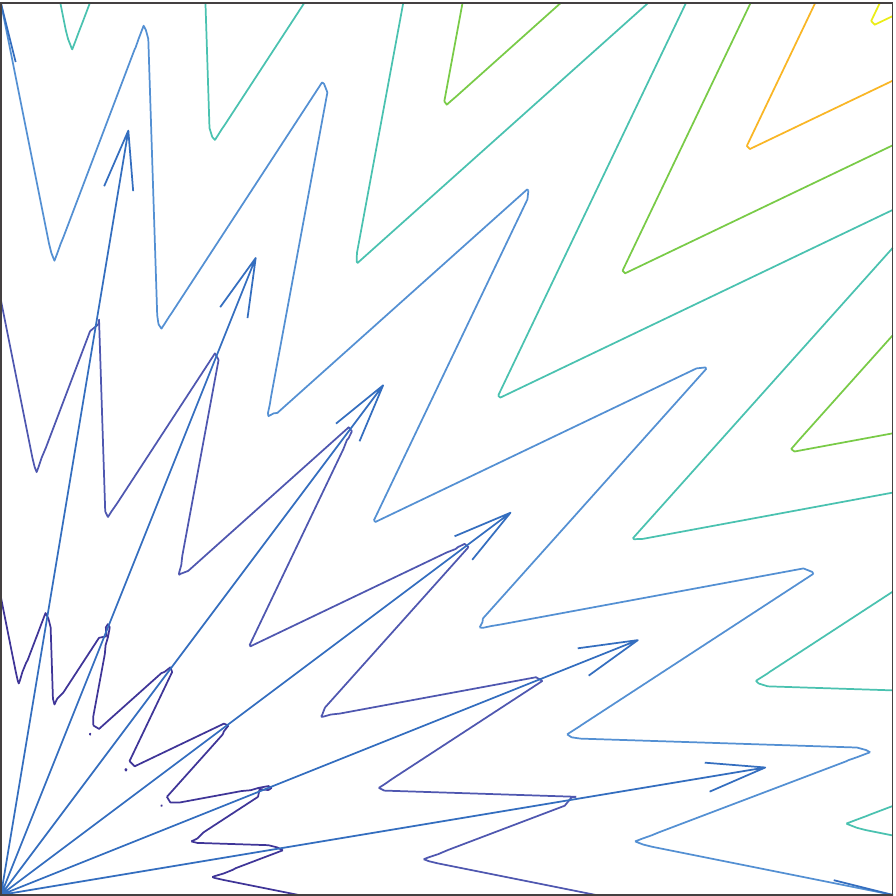}}%
\subfloat[PDM]{
\captionsetup{justification = centering}
\includegraphics[width=0.18\textwidth]{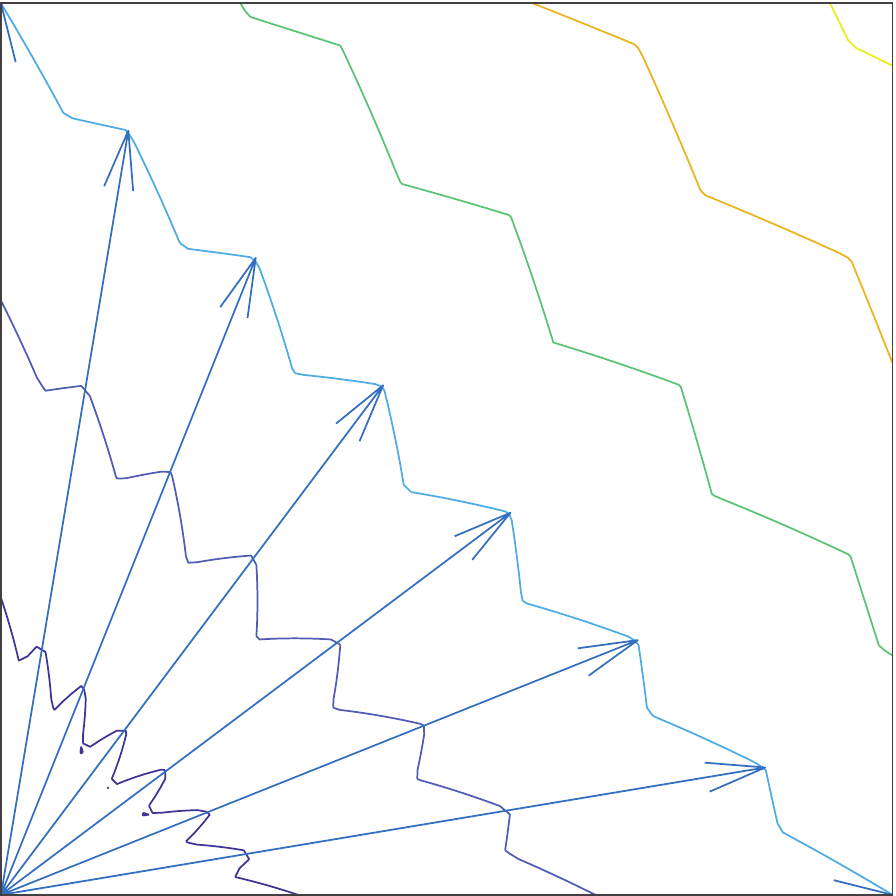}}%
\caption{Fields of PBI function and PDM function in the objective space. We exclude other factors and thus the evolution pressure can be treated as gradients. In (a) and (b), the contours for the proximity term of PBI and PDM is respectively drawn; In (c), the contours for the shared diversity term are, respectively, drawn. In (d) and (e), the contours of the combined PBI field and PDM field are, respectively, illustrated. The contours with the same PBI values fluctuate around concave concentric arcs, while the contours of PDM fluctuate around the parallel lines. PDM has no preference for the PF curvature.}
\label{fig:metric_comparison}
\end{figure*}
\par
There are many designs for evaluating individuals with the help of reference vectors, such as the APD in RVEA \cite{cheng2016reference} that dynamically balances convergence and diversity or the penalty-based boundary intersection approach (PBI) in MOEA/D \cite{zhang2007MOEAD}. In this paper, the design of PDM is inspired from the fitness assignment method of PBI in \cite{zhang2007MOEAD}. They both evaluate a preferred individual using the nearest reference vector with a scalar value. The difference between them is the proximity term. As visually demonstrated in Fig. \ref{fig:metric_comparison}(a) and (d), the proximity term for PBI, which calculates the length of the projection of the individual on the nearest reference vector, guides the evolution of individuals toward the origin. In contrast, PDM, which calculates the mean value, guides the evolution of individuals in a direction perpendicular to the unit simplex. They both have the similar effects of pushing the individuals toward the true PF. As we can observe, the contours with the same PBI values are fluctuating around concave concentric arcs, which may indicate that PBI has a natural preference for concave PFs. Such characteristic may have negative effects on the problems with linear PFs or worse effects on the convex ones. In contrast, in Fig. \ref{fig:metric_comparison}(e), the contours of PDM are fluctuating around the parallel lines. It should be difficult to perfectly approximate a complex PF using a fixed scalarizing function for all reference vectors. Though such design accordingly has the preference for linear PFs, it is a compromised solution with the expectation to perform versatilely on problems with diverse PF curvatures. In the experimental studies, we will validate the versatility and efficiency of PDM compared to PBI.
\par
Then the frontier individuals attached to each reference vector are gathered as clusters. Also, the frontier individual with the best PDM is taken as the center of the corresponding cluster. Each nonfrontier individual is assigned to the clusters with the nearest cluster center. For each cluster, the nonfrontier individuals are sorted by their Euclidean distances to the corresponding cluster centers in ascending orders. In this way, the nearest frontier individuals are used to guide the evolution of the nonfrontier individuals. With the guidance of the elites (cluster centers) of the elites (frontiers), the nonfrontier individuals are under intense pressure toward better proximity and diversity. After the CC, two intraclassly sorted queues are created for each cluster: the sorted frontier queue and the sorted nonfrontier queue.

\subsubsection{Round-Robin Selection}
To ensure the inheritance of the desirable frontier individuals as well as to maintain the diversity by making the selected individuals evenly distributed near the current PF, a round-robin picking method is employed. For each cluster, a selection queue is created in advance by concatenating the sorted nonfrontier queue after the sorted frontier queue. In each round, the head of each selection queue is popped out and added to the next generation until the size of the next generation reaches $N$. It should be noticed that, unless the number of clusters is more than $N$ or replaced by better frontier individuals, all cluster centers will be kept. Also, the selected nonfrontiers are the nearest to the cluster centers, which mean that they are expected to be the ones with the best proximity and diversity.
\par
\begin{figure}[!t]
\centering

\captionsetup{justification = centering}
\includegraphics[width=0.25\textwidth]{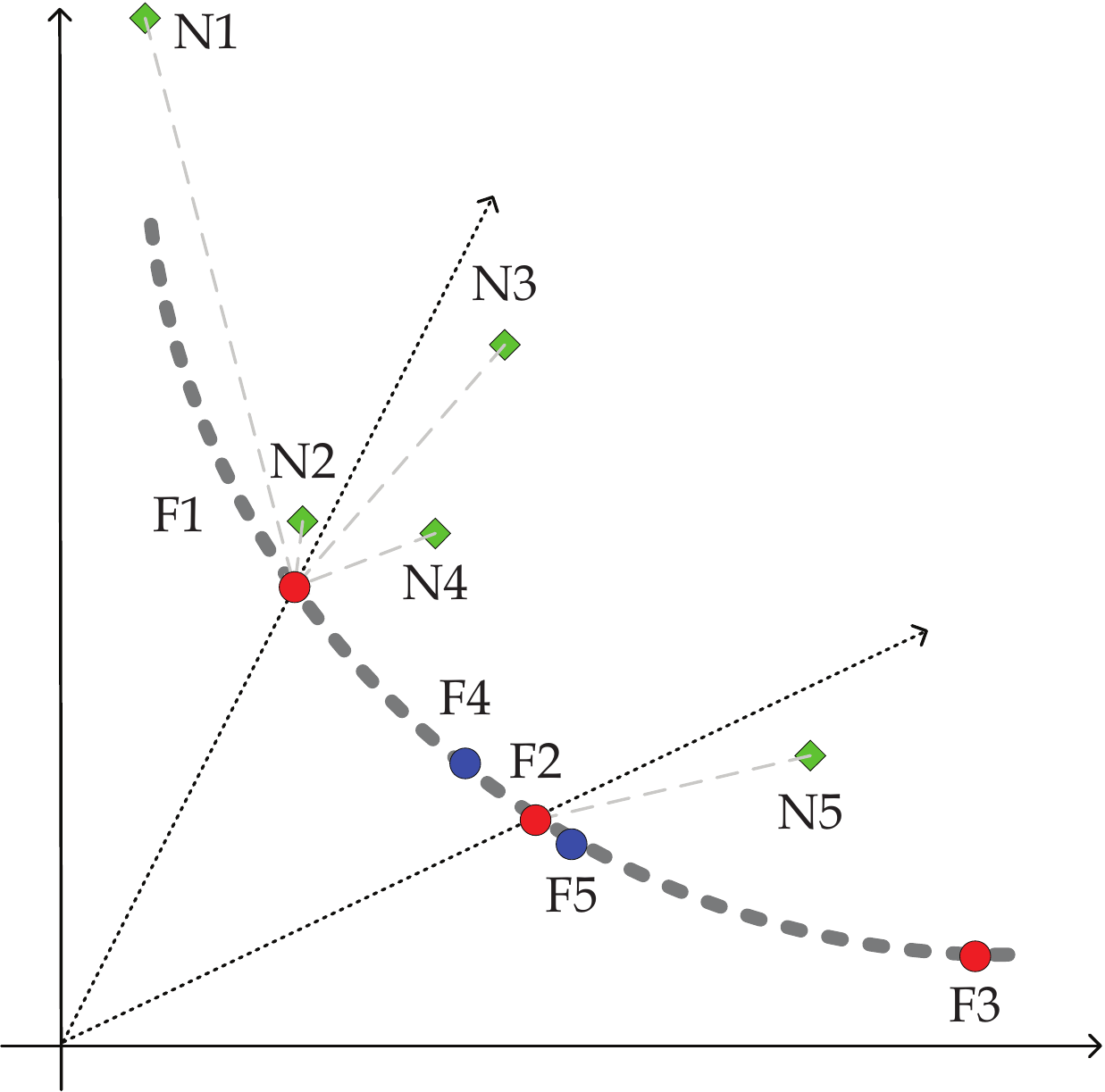}

\caption{Demonstration for \MakeLowercase{\clusteringprocessfullname{}}: Step 1. Frontier identification. The identification divides the population into frontiers (F1 to F5) and nonfrontiers (N1 to N5); Step 2. Cluster and sort the frontiers, locate the cluster centers (F1, F2 and F3). The frontier individuals attached to a mutual reference vector are clustered and sorted using PDM; Step 3. Cluster and sort the nonfrontiers. The nonfrontiers are attached to the clusters and sorted using their distances to the cluster centers; Step 4. Round-robin picking. The next population (F1, F2, F3, N2, N5) is selected by popping the selection queues.}
\label{fig:cc}
\end{figure}

The essence of \MakeLowercase{\clusteringprocessfullname{}} lies in the fact that close numbers of individuals are selected from the evenly distributed clusters over the current PF. The selection guarantees that the selected high-quality individuals (no matter frontiers or nonfrontiers) are evenly distributed near the current PF. Thus it is able to obtain populations with good proximity and diversity. The process of \MakeLowercase{\clusteringprocessfullname{}} is visualized in Fig. \ref{fig:cc}.
\par
With $M$ objectives and $N$ individuals, frontier individuals are picked out in $\scriptO{}(MN^2)$ at worst. And $\scriptO{}(MN^2)$ is costed on other operations including the bi-level clustering and intraclass sorting. Thus the overall runtime complexity for \MakeLowercase{\clusteringprocessfullname{}} is $O(MN^2)$, which is at the same level of the nondominated sorting-based algorithms with $\scriptO{}(MN^2)$ costs at worst \cite{tian2017effectiveness}. Though they share the same runtime complexity, the efficiency that can be observed in the experiments of \MakeLowercase{\clusteringprocessfullname{}} is worthy of being highlighted. Only identifying the frontier individuals generally saves time no matter which NDSort algorithm is employed; Though we have not used parallelization in the experiments for fair comparison, operations in each cluster are potentially parallelizable including the PDM calculation, intraclass sorting for the frontiers and nonfrontiers.
\par
The pseudo code of the CC process is shown in Algorithm \ref{alg:clustering}. The activities (active or not) of reference vectors in the clustering process are additionally gathered and handed to the \MakeLowercase{\learningprocessfullname{}} process.
\subsection{\learningprocessfullname{}}
\begin{figure*}[!t]
\centering
\subfloat[$1^{st}$ sample]{
\captionsetup{justification = centering}
\includegraphics[width=0.2\textwidth]{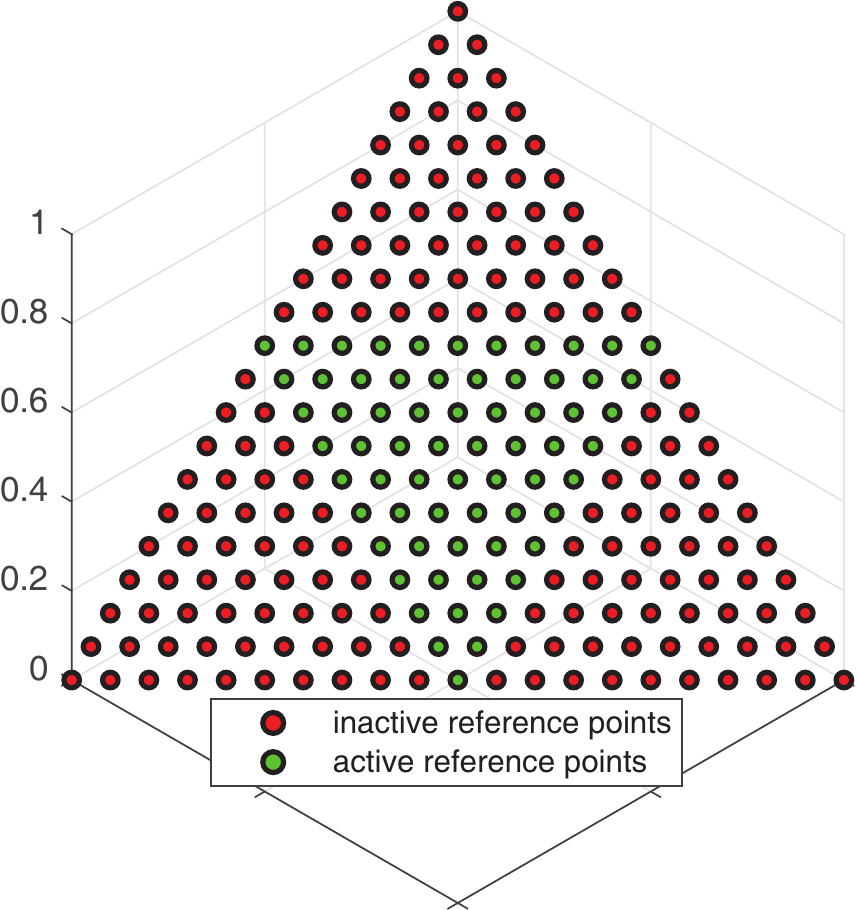}}
\hfill
\subfloat[$1^{st}$ learning]{
\captionsetup{justification = centering}
\includegraphics[width=0.2\textwidth]{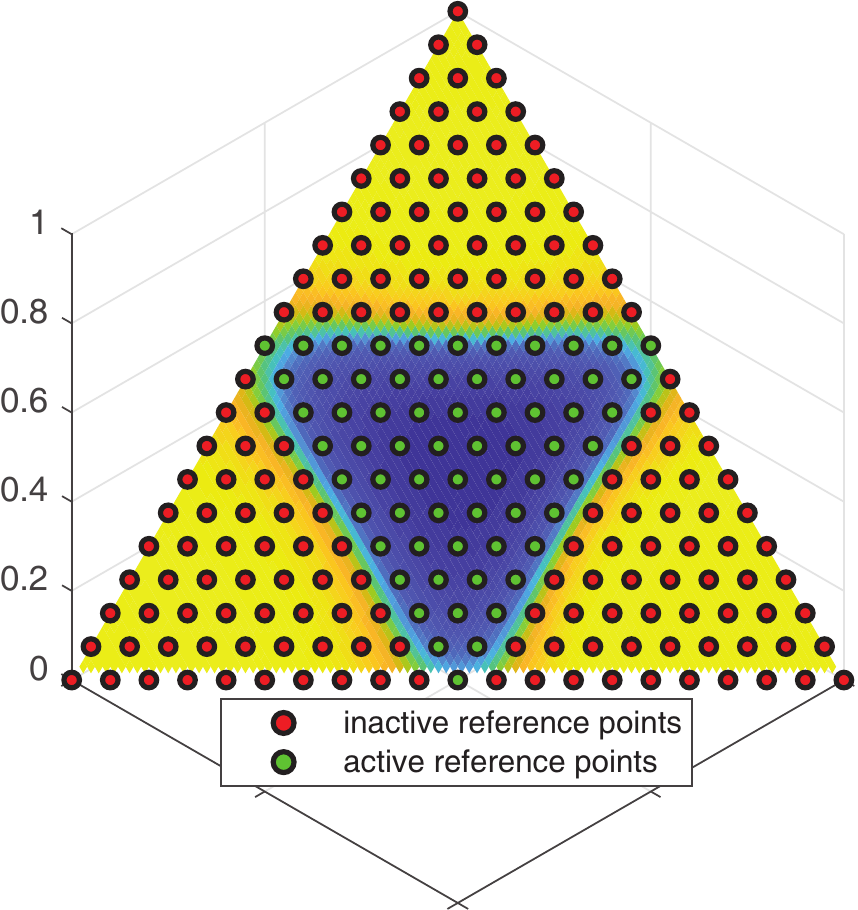}}
\hfill
\subfloat[$1^{st}$ reduction]{
\captionsetup{justification = centering}
\includegraphics[width=0.2\textwidth]{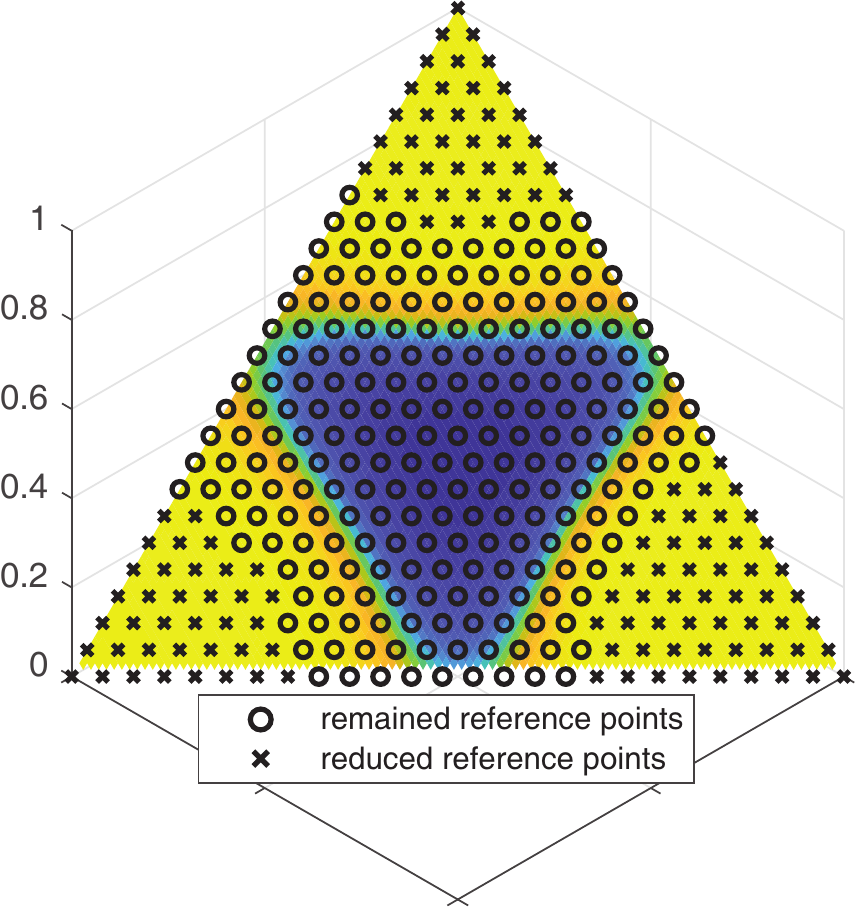}}
\hfill
\subfloat[$2^{nd}$ sample]{
\captionsetup{justification = centering}
\includegraphics[width=0.2\textwidth]{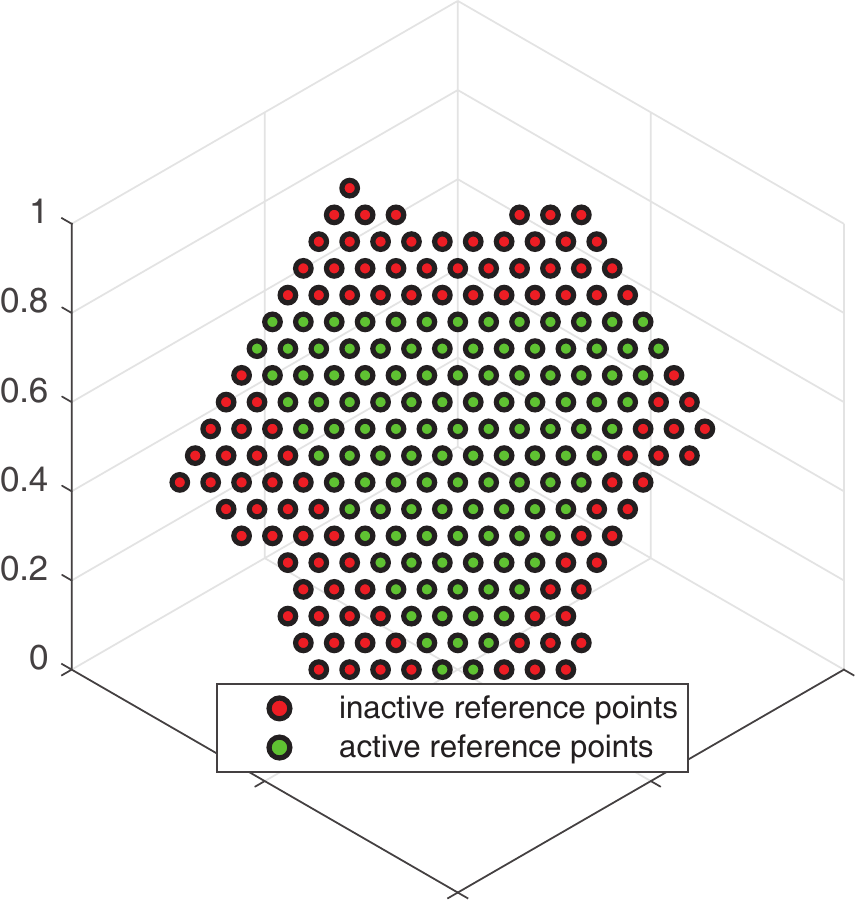}}

\subfloat[$2^{nd}$ learning]{
\captionsetup{justification = centering}
\includegraphics[width=0.2\textwidth]{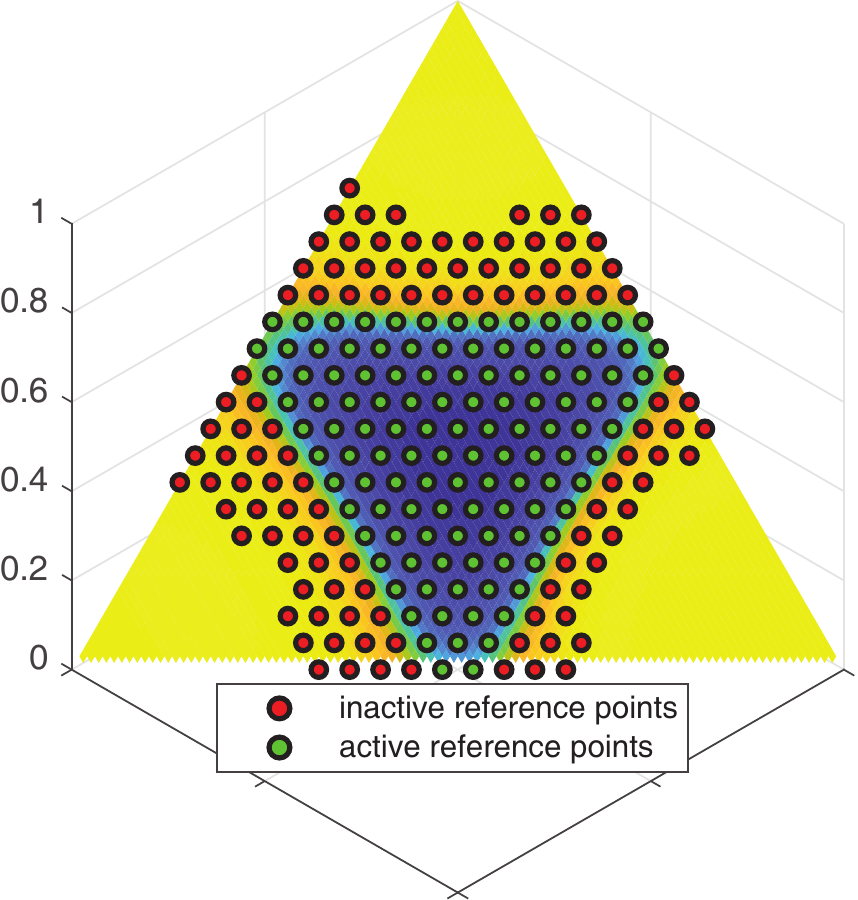}}
\hfill
\subfloat[$2^{nd}$ reduction]{
\captionsetup{justification = centering}
\includegraphics[width=0.2\textwidth]{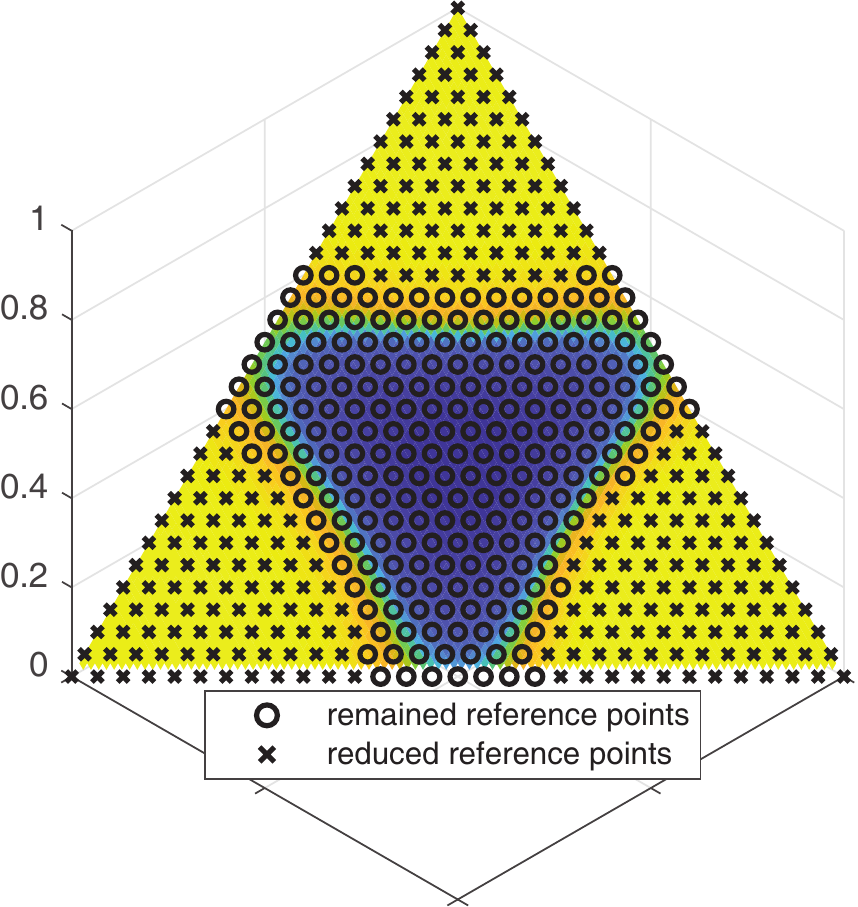}}
\hfill
\subfloat[$3^{rd}$ sample]{
\captionsetup{justification = centering}
\includegraphics[width=0.2\textwidth]{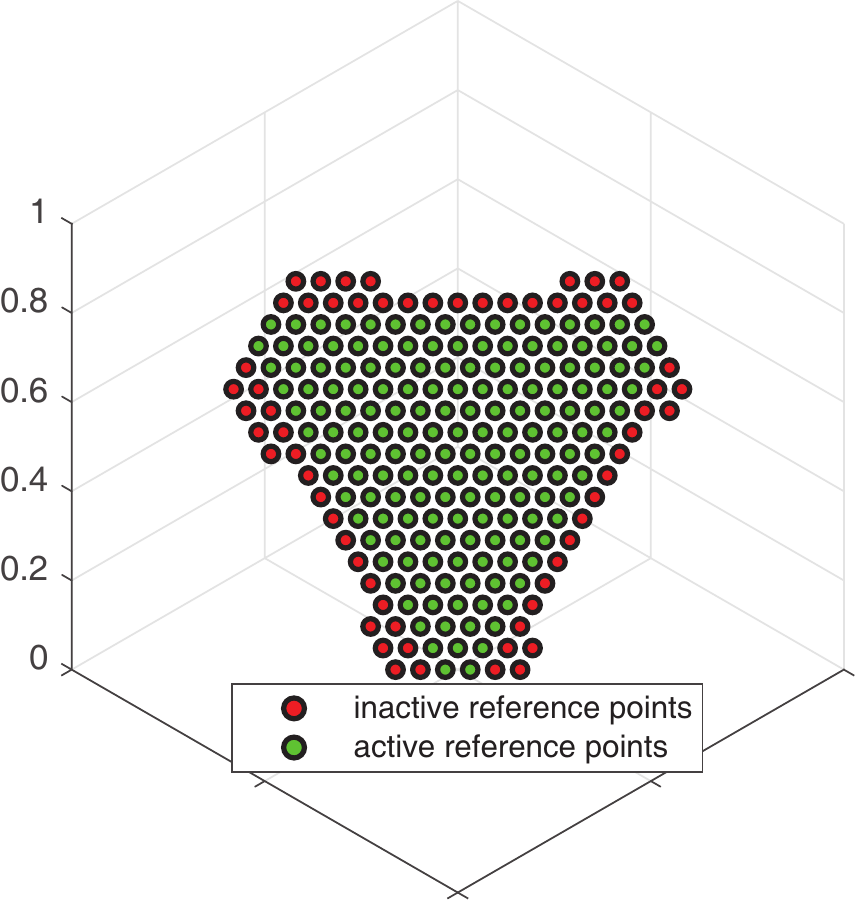}}
\hfill
\subfloat[Last sample]{
\captionsetup{justification = centering}
\includegraphics[width=0.2\textwidth]{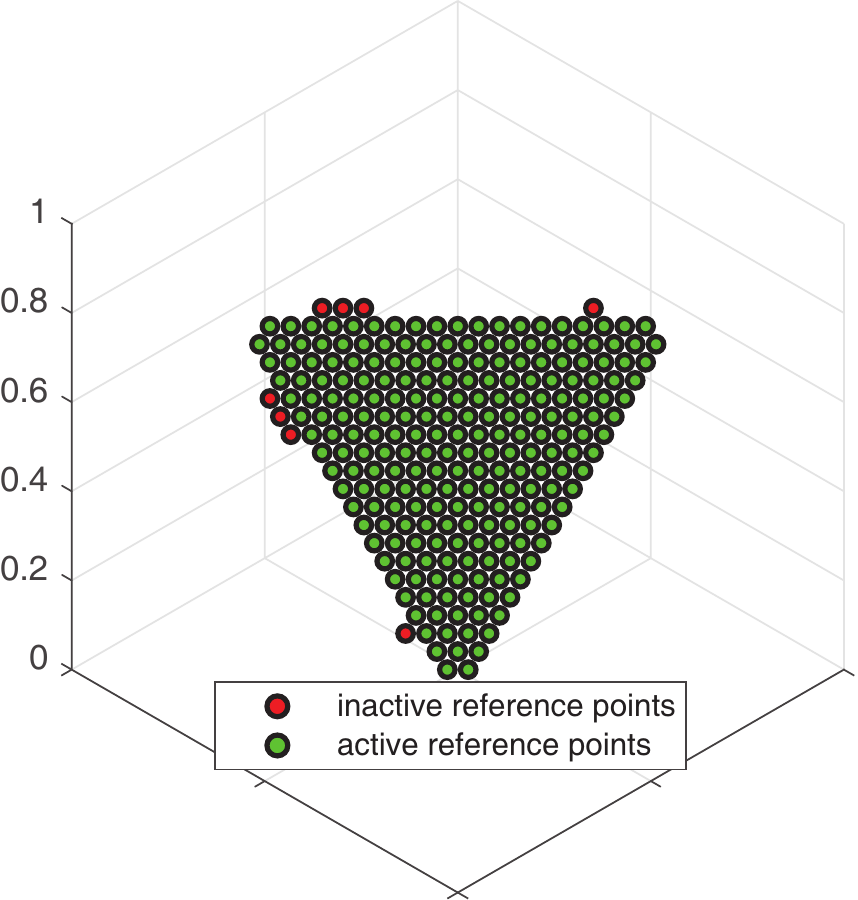}}
\caption{Demonstration for the \MakeLowercase{\learningprocessfullname{}} process: (a). The status sampler reports that the activities of the reference vectors are stable, and the first batch of samples for learning is obtained; (b). The incremental learner is initiated to learn the samples from (a), the background surface shows the scores of each location on the unit simplex, the more blue the higher potential, the more yellow the lower; (c). The old reference points are all discarded and denser reference points are generated on the unit simplex. The reference points with scores lower than $\delta$ are reduced. In this example, $\delta=0.33$; (d). With the remaining newly generated potential reference vectors, clustering process keep pulling the population close to them. After a few generations, the second sample set for learning is obtained; (e, f, g, h). After several sampling-learning-reducing cycles, the reference vectors are repositioned to fit the true PF as the boundaries for the effective areas become clearer.}
\label{fig:redistribution}
\end{figure*}
Though the reference vectors that do not intersect with the true PF can be useful during the evolutionary process \cite{ishibuchi2017performance}, such usefulness for the population will become trivial when the population is proximate to the true PF since then the pivot mission is to make the population wide-spread for better diversity. For problems with partial PFs, when individuals are proximate to the true PF, some reference vectors that do not intersect with the true PF are hard to be activated since they are too far to be attached with frontier individuals. Under such circumstance, there will be fewer than $N$ reference vectors guiding the evolution. Though aforementioned analyses on CC indicate that the selected population will be roughly evenly distributed near the current PF, it is impossible to reach the final state of \textit{one frontier attached to one reference vector} as it can be in the problems with full PFs. So, the diversity of the population should be further improved.
\par
When the number of active reference vectors are always insufficient, a natural solution would be generating more reference vectors corresponding to higher density of reference points over unit simplex. However, this leads to huge amount of reference vectors participating the clustering and a lot of them are actually impossible to be activated, \ie{} they are ineffective. For efficiency, the number of ineffective reference vectors must be controlled to an acceptable level.
\par
An example is given in Fig. \ref{fig:projection}. Reference vectors are generated by using the uniformly distributed reference points over the true PF. In the example, only a part of reference vectors can be attached by the frontier individuals on the true PF. No matter how dense the reference vectors are, we can observe that reference points corresponding to the active reference vectors hardly locate outside the projection areas of the true PF. This indicates that the appropriate reference points for populations with good proximity should be located in the projection areas of the true PF, \ie{} effective areas on the unit simplex. If we successfully generate more evenly distributed reference points inside these effective areas and reduce the outsiders, there will be more reference vectors intersecting with the true PF while the efficiency can be maintained by reducing the ineffective outsiders. These appropriate reference vectors can distribute the population more evenly over the true PF, where the individuals were once overcrowded near the insufficient active reference vectors.
\par
This motivates us to find proper ways to estimate the effective areas on the unit simplex. However, since the characteristics of the true PF are not known priors, they are hard to capture. Also, the effective areas may be disconnected, irregular or more sophisticated, fitting them in a definitive model is also hard.
\par
For the populations with good proximity selected by \MakeLowercase{\clusteringprocessfullname{}}, the activities of reference vectors can reflect the distribution of the true PF in the objective space. Each time a reference vector set is generated, there will be active and inactive reference vectors. Therefore, we can use the activities of reference points to train a classifier for identifying the effective and ineffective reference points and therefore estimate the effective areas on the unit simplex. If a point locates inside the areas of the positive training samples, it will be scored high and vice versa. When reference points of higher densities are generated, the trained classifier will evaluate the effectiveness of the points by scoring them.
\par
The activities of reference vectors are effective samples for training only when the current PF is close enough to the true PF and the individuals are stably distributed in the existing clusters. We notice that even if historically most of the reference vectors have once been activated, for the problems with partial PFs, when the population gradually reaches the true PF, the number of continuously active reference vectors still falls below the population size $N$. The phenomena inspire our solution of employing a status sampler. It compares the historical activities of the reference vectors with the current ones. The status sampler will report ``stable'' status, only when the activities of all reference vectors do not change for a certain $\theta$ generations. In \cite{cheng2016reference}, it is stated that difficulties for adaptation include the moment it should be carried out, because even for a regular PF, it is still likely that some reference vectors can be occasionally inactive. The design of status sampler lowers the possibilities for the inaccuracies. For such stable status of reference vector activities show that in $\theta$ generations the current PF have been stably distributed around the active reference vectors and no significant shrinking or expanding for the extent of the current PF has been observed. Such stableness is considered as the indicator for the appropriate moment to learn.
\par
After the stable status sampling, we train a classifier in the aforementioned way. After the training, denser reference points on the unit simplex will be generated using the same way\footnote{For $M < 8$, we increase the lattice density. Else, we increase the density of reference points in the boundary layer and the inner layers.} as NSGA-III \cite{deb2013evolutionary} and they will be scored by the trained classifier. Then, the potentially effective ones with the scores greater than threshold $\delta$ will be kept.
\par
Usually, the effective areas are small compared to the whole unit simplex. Thus the one-shot learning of the reference points cannot guarantee the accurate identification. With the increase of the density of reference points, the distribution of the samples will reflect the boundaries of the effective areas clearer. Also, increasing the generation density only once may not generate enough active reference vectors. These all indicate the need for the sampling-learning-reducing procedure to run multiple times. If we simply store all the samples and each time retrain the classifier with all the data, it will result in drastic growth for the demand of computational resources. In this paper, the strategy of incremental learning is employed to train the model upon the prior knowledge \cite{diehl2003SVM}. In each learning iteration, the reserved training samples near the margin samples from the previous learning iterations will participate the new iteration together with the new training samples and the learning is iteratively conducted upon the prior knowledge. Such method can increase the accuracy for boundary identification and boost the training speed when there are packs of data to be learned \cite{diehl2003SVM}. In Fig. \ref{fig:redistribution}, we investigate into a real-time running process of the cycles of sampling-learning-reducing on the problem of MaF1, $M=3$ with an inverted triangle partial true PF.
\par
The threshold $\delta$ is an important factor for the selection of potentially effective reference points. It should not be fixed constant for the two following reasons. First, the accuracy of the classifier identifying the effective areas is generally increasing with the density of the training samples. Thus the fixed $\delta$ may lead to the reduction of the effective reference vectors. Second, the scores of the boundaries vary as classifier learns incrementally. Thus the fixed $\delta$ may lead to over-reduction, \ie{} the number of the reference vectors left is even lower than $N$. Under such circumstance, the expected state of one frontier on one reference vector cannot be reached. The higher the score, the more likely the corresponding vector can be activated. A good choice is to keep $n$ ($n > N$) reference points with the best scores after the reduction. Thus $\delta$ is adaptively set to the $n$-th highest score of the scored reference points. As long as $n$ is relatively small, this setting will ensure the robustness of the incremental learning while maintaining the efficiency. In the experiments, we have chosen $n=2N$ based on the experimental results.
\par
The computational complexity of the incremental SVM is at worst the same as a normal SVM, which runs at $O(max\{n, d\} \cdot {min\{n, d\}}^2)$ \cite{chapelle2007training}, where $n$ is the number of samples and $d$ is the dimension of the sample space. In \algoabbr{}, with the preprocess of Gram-Schmidt, we lower $d=M$ to $d=M-1$. Since the number of samples $n$ are far greater than $d$, that will lead to $O(max\{n, d\} \cdot {min\{n, d\}}^2)=O(nM^2)$. In our implementation, the worst runtime complexity is $O(nM^2)=O(NM^2)$. This reference vector adaptation method is attractive since its training is fast where no extra burden on the objective function fitness evaluations is incurred and only the feedbacks of the clustering process is used. The pseudo-code for the adaptation process is in Algorithm \ref{alg:learning}.
\par
It should be noticed that the design of the incremental learning process is highly compatible with the adaptation methods that adapts to the curvature of the true PF. The incremental learning deals with the problems with partial PFs, combining with other adaptation techniques, it could boost the performance for the reference vector-based algorithms.
\begin{algorithm}[t]
\small
\caption{\learningprocessfullname{}}
\label{alg:learning}
\KwIn{$Z_{A}$ (active reference points), $Z_{I}$ (inactive reference points), $S$ (sampler), $D$ (old reference point generation density), $M$ (number of objectives), $N$ (population size), $C$ (classifier)}
\KwOut{$Z$ (new reference point set), $D$ (new generation density), $C$ (classifier) }
\If{isStable($S$.status) and $|Z_{A}| < N$}{
    \textcolor{darkgreen}{//Project points to one dimension lower and assemble training set $\Delta$ with labels ``active'' or ``inactive''}\\

    $\Delta \leftarrow$ Assemble(Projection($Z_{A}$), Projection($Z_{I}$))\;

    \eIf{exist($C$)}{
%
        $C$.learn($\Delta$, 'incremental');
    }{
        $C \leftarrow$ new(incrementalLearner)\;
        $C$.learn($\Delta$);
    }
    \textcolor{darkgreen}{//Generate denser reference points}\\
    $D \leftarrow D + 1$\;
    $Z \leftarrow$ generateReference($D$, $M$)\;
    \textcolor{darkgreen}{//get $\delta$, which depends on a specific strategy}\\
    $\delta \leftarrow$ getDelta($\cdot$)\;
    \textcolor{darkgreen}{//pick the potential points whose $p \geq \delta$}\\
    $\bm{y} \leftarrow C$.predict(Projection($Z$))\;
    $Z \leftarrow$ find($Z$, $\bm{y}$, $\delta$);
}
\end{algorithm}
\subsection{Proposed Framework}
\begin{figure}[!t]
  \centering
  \includegraphics[width=0.35\textwidth]{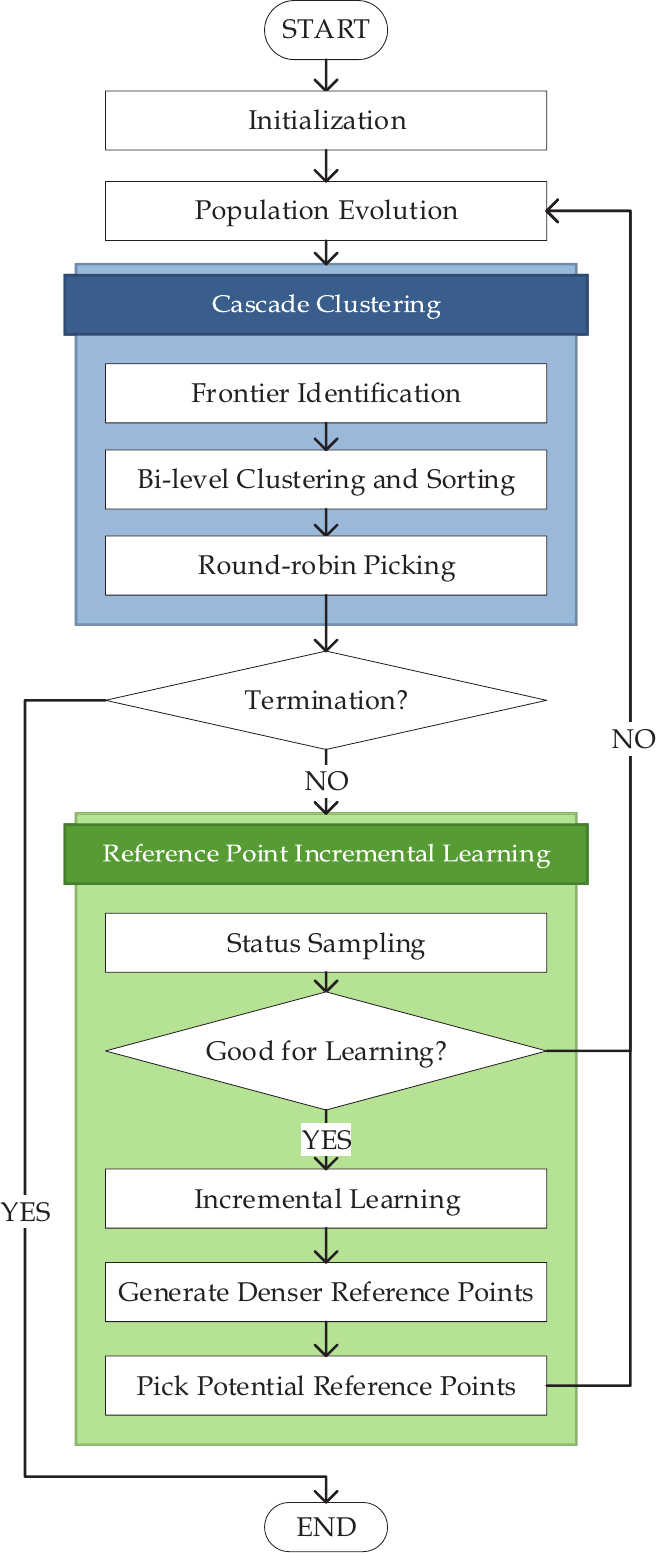}\\
  \caption{Framework of \algoabbr{}}
  \label{fig:framework}
\end{figure}

The framework of \algoabbr{} is presented in Fig. \ref{fig:framework}.
\par
First, to initialize, $N$ individuals are randomly generated with $N$ uniformly distributed reference vectors. Then, the main loop cycles until the criteria of termination are satisfied: evolve the population to get the offsprings and combine them as the potential population with the size of $2N$. Then, using the proposed CC, the new population is picked out of the potential population. The activities of reference vectors are then delivered to the status sampler. If the sampler assess the current status stable and the number of active reference vectors is not enough, \MakeLowercase{\learningprocessfullname{}} will learn incrementally to estimate the effective areas, and then generate proper reference vectors. The algorithm proceeds along with the interactions of the two processes.
\section{Experimental Studies}
\subsection{Experimental Settings}
To validate and analyze the performance and behavior of the proposed \algoabbr{} and its components, the following five sets of experiments are conducted.
\begin{enumerate}
\item
Comparative analyses on the MaF benchmark suite \cite{cheng2017benchmark}: Compare the overall performance with the state-of-the-art algorithms to demonstrate the capabilities of \algoabbr{} handling diverse characteristics.
\item
Component analysis for the frontier metric PDM: Validate the characteristics of PDM on the problems with diverse true PF curvatures by comparing it to PBI \cite{zhang2007MOEAD};
\item
Component analysis for cascade clustering: Validate the performance of \MakeLowercase{\clusteringprocessfullname{}} on problems with full PFs by comparing with the counterparts in the state-of-the-art reference vector-based algorithms;
\item
Component analysis for \MakeLowercase{\learningprocessfullname{}}: Validate the effectiveness of the reference vector adaptation by comparing the real-time reference vector activities of \algoabbr{} with the counterparts in the state-of-the-art algorithms on problems with partial PFs.
\item
Behavior analyses for the interacting processes on partial PFs: Validate the feasibility and the effectiveness of the clustering-learning interactions, demonstrate the effectiveness of the whole design.
\end{enumerate}
\par
Experiments were conducted via PlatEMO \cite{PlatEMO} with MATLAB R2018a on Intel Core i7-8700k (4.70GHz).
\par
The setting for each test case is identical to the standard for CEC'2018 MaOP competition \cite{cheng2017benchmark}. The number of fitness evaluations is set to be $maxFEs=max(1e5, D \times 1e4)$, where $D$ is the default dimension of the solution space that corresponds to a certain number of objectives $M$. The evolution operator is the same for every algorithm, which is the combination of simulated binary crossover (SBX) and the polynomial mutation, with crossover distribution index ${\eta}_{c}=20$, crossover probability $p_c=1$, mutation distribution index ${\eta}_{m}=20$ and mutation probability $p_m=1/D$ \cite{cheng2017benchmark}.
\par
The performance metric adopted in this section is Inverted Generational Distance (IGD), a prevailing metric which measures both diversity and proximity of an obtained population. The smaller the IGD, the better the proximity and diversity.
\subsection{Parameter Settings}
For \algoabbr{}, we expect to find an appropriate parameter set for all the test cases without the manual parameter adjustments by trial-and-error. Parameter selections are based on parameter sensitivity analyses, which can be found in the supplementary file.
\par
The choice of the status sampling threshold $\theta$ should be effective, reasonable and good for generalization. In the experiments, we found that $\theta\!\geq\!5$ is required to prevent inaccurate samplings and that the distribution of the population can be stable enough when $\theta\!=\!20$. Also, if $maxFEs$ are sufficient enough, $\theta$ should be increased for more careful samplings. Considering these, we have adopted $\theta\!=\!min\{\!20,\!max\{5,\!ceil(maxFEs\!/\!2e4)\}\}$.
\par
\begin{figure}[!t]
\centering
\captionsetup{justification = centering}
\includegraphics[height=0.25\textwidth]{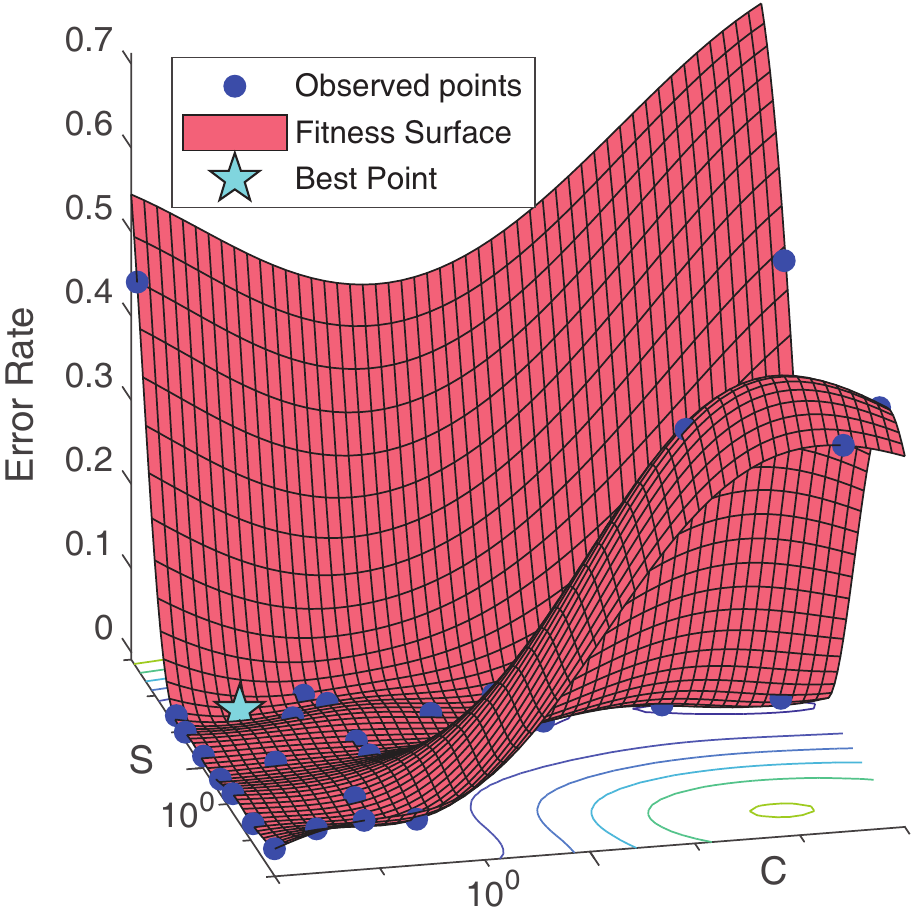}
\caption{Visualization for SVM hyperparameters selection}
\label{fig:parameters}
\end{figure}

For the selection of the hyper-parameters $\langle{}S,C\rangle$ ($S$ is the kernel scale of the Gaussian kernel, $C$ is the soft-margin regularization parameter) in the incremental SVM learner, we use different pairs of hyper-parameters $\langle{}S,C\rangle$ on DTLZ7, which is a complicated test case with partial disconnected true PF. After each run is finished, each final state of the incremental SVM is tested with many points both inside and outside the actual effective areas on the unit simplex. The error rate is optimized by an automatic heuristic-based optimizer in MATLAB to find an appropriate value of the hyper-parameters. From the results, we have adopted $\langle{}S,C\rangle=\langle{}0.056,10\rangle$. The hyper-parameter optimization results are visualized in Fig. \ref{fig:parameters}.
\par
For the threshold $\delta(n)$ for potentially effective reference points, we have tested the closest integers to each number of the set $\{1.25N,1.5N,1.75N,2N,3N,4N,5N\}$ and found that once the number reaches $2N$, the differences in performance become trivial. Thus, we choose $n=2N$, \ie{} after the reduction, there will be at least $2N$ points left (if the total number is below $2N$, all will be kept).
\par
The settings of the fixed SVM hyperparameters and the auto-selected $\theta$ and the adaptive $\delta$ are set globally for all the test cases, which is beneficial for generalization.
\subsection{Comprehensive Capabilities Analyses}
MaF benchmark, designed for CEC'2018 MaOP competition, is with diverse properties for a systematic study of the MOEAs. The performance on the benchmark suite can demonstrate the overall capabilities to complexities which represent various real-world scenarios \cite{cheng2017benchmark}. In the following parts, we go detail into the performance of \algoabbr{} on the problems with different characteristics.
\subsubsection{Overall Results on CEC'2018 Competition}
$15$ problems are provided in the MaF benchmark suite. The competition standard includes solving each problems with $\langle\!N,M\!\rangle\!\in\!\{\langle\!210,5\!\rangle,\langle\!230,10\!\rangle,\langle\!240,15\!\rangle\} $, constituting $45$ different test cases in total. In Table \ref{tab:IGD}, the averaged IGD performance is compared with the state-of-the-art algorithms including AR-MOEA \cite{tian2017indicator}, A-NSGA-III \cite{jain2014evolutionary}, Two\_Arch2 \cite{wang2015improved}, GrEA \cite{yang2013grid}, KnEA\cite{zhang2015knee}, MOEA/DD \cite{li2015evolutionary} and RVEA* \cite{cheng2016reference}. All data presented are averaged over $20$ independent runs. For each test case, the best averaged IGD result is shaded gray.
\begin{table*}[htbp]
\setlength{\tabcolsep}{1.25pt}
\renewcommand\arraystretch{0.8}
\scriptsize
  \centering
  \caption{The IGD Results for CEC'2018 MaOP Competition Test Cases with Results of $t$-tests and Friedman Tests}
    \begin{tabular}{cp{1.5em}cccccccccccccccc}
    \toprule
    \toprule
    \multicolumn{2}{c}{\multirow{2}[1]{*}{}} & \multicolumn{2}{c}{\algoabbr{}} & \multicolumn{2}{c}{AR-MOEA} & \multicolumn{2}{c}{A-NSGA-III} & \multicolumn{2}{c}{Two\_Arch2} & \multicolumn{2}{c}{GrEA} & \multicolumn{2}{c}{KnEA} & \multicolumn{2}{c}{MOEA/DD} & \multicolumn{2}{c}{RVEA*} \\
    \multicolumn{2}{c}{} & Mean  & Std   & Mean  & Std   & Mean  & Std   & Mean  & Std   & Mean  & Std   & Mean  & Std   & Mean  & Std   & Mean  & Std \\
    \multicolumn{1}{c}{\multirow{3}[0]{*}{MaF1}} & \multicolumn{1}{c}{5} & \cellcolor[rgb]{ .859,  .859,  .859}1.1e-1 & 7.55e-4 & 1.41e-1 & 1.89e-3 & 1.58e-1 & 1.45e-2 & 1.23e-1 & 1.44e-3 & 1.12e-1 & 6.68e-4 & 1.17e-1 & 1.78e-3 & 3.01e-1 & 1.28e-2 & 1.46e-1 & 8.94e-3 \\
          & \multicolumn{1}{c}{10} & 2.39e-1 & 3.09e-3 & 2.52e-1 & 1.29e-3 & 2.73e-1 & 1.11e-2 & \cellcolor[rgb]{ .859,  .859,  .859}2.37e-1 & 2.32e-3 & 2.27e-1 & 8.04e-3 & 2.35e-1 & 3.98e-3 & 3.89e-1 & 2.45e-2 & 3.47e-1 & 2.86e-2 \\
          & \multicolumn{1}{c}{15} & \cellcolor[rgb]{ .859,  .859,  .859}2.63e-1 & 2.81e-3 & 2.88e-1 & 1.25e-2 & 3.15e-1 & 6.29e-3 & 2.83e-1 & 3.04e-3 & 2.95e-1 & 1.09e-2 & 3.0e-1 & 4.65e-3 & 5.34e-1 & 2.93e-2 & 4.32e-1 & 3.34e-2 \\
    \multirow{3}[0]{*}{MaF2} & \multicolumn{1}{c}{5} & \cellcolor[rgb]{ .859,  .859,  .859}9.6e-2 & 1.67e-3 & 1.11e-1 & 1.03e-3 & 1.06e-1 & 1.8e-3 & 1.07e-1 & 1.81e-3 & 9.65e-2 & 9.91e-4 & 4.25e-1 & 5.61e-2 & 1.4e-1 & 1.03e-2 & 1.09e-1 & 1.95e-3 \\
          & \multicolumn{1}{c}{10} & \cellcolor[rgb]{ .859,  .859,  .859}1.6e-1 & 1.62e-3 & 2.06e-1 & 8.73e-3 & 2.32e-1 & 2.45e-2 & 1.66e-1 & 2.05e-3 & 1.7e-1 & 1.61e-3 & 1.64e-1 & 9.09e-3 & 2.14e-1 & 1.65e-2 & 2.71e-1 & 1.19e-2 \\
          & \multicolumn{1}{c}{15} & 1.66e-1 & 2.52e-3 & 2.26e-1 & 1.02e-2 & 2.2e-1 & 1.28e-2 & \cellcolor[rgb]{ .859,  .859,  .859}1.65e-1 & 1.7e-3 & 2.03e-1 & 5.68e-3 & 1.93e-1 & 1.57e-3 & 4.06e-1 & 4.89e-2 & 2.84e-1 & 2.12e-2 \\
    \multicolumn{1}{c}{\multirow{3}[0]{*}{MaF3}} & \multicolumn{1}{c}{5} & \cellcolor[rgb]{ .859,  .859,  .859}6.36e-2 & 1.12e-3 & 9.71e-2 & 1.74e-3 & 8.3e-2 & 2.45e-2 & 1.08e-1 & 1.64e-2 & 6.07e-1 & 2.72e-1 & 1.92e-1 & 1.24e-1 & 1.17e-1 & 1.42e-3 & 7.68e-2 & 2.65e-3 \\
          & \multicolumn{1}{c}{10} & \cellcolor[rgb]{ .859,  .859,  .859}7.45e-2 & 2.6e-3 & 4.08e0 & 1.18e1 & 5.61e6 & 2.01e7 & 7.14e-1 & 9.63e-1 & 2.69e-1 & 2.59e-1 & 1.89e-1 & 2.37e-3 & 1.16e-1 & 2.72e-3 & 7.73e0 & 8.66e0 \\
          & \multicolumn{1}{c}{15} & \cellcolor[rgb]{ .859,  .859,  .859}8.56e-2 & 6.07e-4 & 5.48e1 & 1.51e2 & 2.35e2 & 6.16e2 & 1.85e6 & 5.35e6 & 4.52e0 & 1.7e1 & 3.86e1 & 1.72e2 & 1.09e-1 & 1.99e-3 & 2.66e-1 & 6.95e-1 \\
    \multirow{3}[0]{*}{MaF4} & \multicolumn{1}{c}{5} & \cellcolor[rgb]{ .859,  .859,  .859}1.9e0 & 6.47e-2 & 2.45e0 & 9.25e-2 & 2.35e0 & 1.49e-1 & 1.92e0 & 2.47e-2 & 2.45e0 & 1.93e0 & 2.45e0 & 2.01e-1 & 7.6e0 & 3.43e-1 & 2.09e0 & 9.62e-2 \\
          & \multicolumn{1}{c}{10} & 1.0e2 & 2.36e0 & 9.68e1 & 6.42e0 & 9.74e1 & 6.05e0 & \cellcolor[rgb]{ .859,  .859,  .859}5.38e1 & 3.03e0 & 6.72e1 & 4.35e1 & 7.8e1 & 1.05e1 & 3.98e2 & 1.64e1 & 7.97e1 & 7.62e0 \\
          & \multicolumn{1}{c}{15} & 3.62e3 & 1.62e1 & 4.01e3 & 5.26e2 & 3.74e3 & 3.39e2 & \cellcolor[rgb]{ .859,  .859,  .859}1.48e3 & 2.42e2 & 4.1e3 & 7.45e2 & 1.64e3 & 1.7e2 & 1.52e4 & 1.52e3 & 2.77e3 & 2.46e2 \\
    \multicolumn{1}{c}{\multirow{3}[0]{*}{MaF5}} & \multicolumn{1}{c}{5} & \cellcolor[rgb]{ .859,  .859,  .859}1.93e0 & 8.36e-3 & 2.39e0 & 8.01e-1 & 1.97e0 & 4.9e-3 & 1.95e0 & 2.83e-2 & 1.77e0 & 2.64e-2 & 2.19e0 & 5.82e-2 & 6.18e0 & 1.28e0 & 2.26e0 & 8.41e-1 \\
          & \multicolumn{1}{c}{10} & 7.8e1 & 1.03e0 & 9.87e1 & 4.11e0 & 8.08e1 & 4.1e0 & 4.93e1 & 1.3e0 & \cellcolor[rgb]{ .859,  .859,  .859}4.52e1 & 1.05e0 & 7.66e1 & 5.15e0 & 2.87e2 & 1.4e1 & 7.03e1 & 1.39e1 \\
          & \multicolumn{1}{c}{15} & 2.29e3 & 2.05e2 & 3.31e3 & 4.44e2 & 2.32e3 & 6.54e2 & \cellcolor[rgb]{ .859,  .859,  .859}1.14e3 & 4.93e1 & 1.29e3 & 6.75e1 & 1.93e3 & 1.75e2 & 7.31e3 & 1.7e1 & 2.54e3 & 7.97e2 \\
    \multirow{3}[0]{*}{MaF6} & \multicolumn{1}{c}{5} & \cellcolor[rgb]{ .859,  .859,  .859}2.4e-3 & 3.32e-4 & 4.22e-3 & 5.63e-5 & 1.82e-2 & 1.14e-2 & 6.56e-3 & 5.3e-4 & 2.32e-2 & 1.22e-3 & 2.71e-3 & 1.84e-4 & 7.81e-2 & 1.99e-3 & 2.37e-2 & 4.98e-3 \\
          & \multicolumn{1}{c}{10} & \cellcolor[rgb]{ .859,  .859,  .859}3.4e-2 & 1.28e-4 & 1.08e-1 & 1.4e-1 & 3.02e0 & 4.22e0 & 5.68e-1 & 2.6e-1 & 2.87e-1 & 6.89e-2 & 2.28e0 & 6.86e0 & 1.13e-1 & 5.45e-3 & 4.39e-2 & 4.22e-2 \\
          & \multicolumn{1}{c}{15} & \cellcolor[rgb]{ .859,  .859,  .859}4.24e-2 & 1.5e-2 & 2.38e-1 & 1.25e-1 & 1.55e1 & 1.37e1 & 7.42e-1 & 3.77e-6 & 4.33e-1 & 3.12e-1 & 5.02e-1 & 1.97e-1 & 1.31e-1 & 5.15e-3 & 1.68e-1 & 1.91e-1 \\
    \multicolumn{1}{c}{\multirow{3}[0]{*}{MaF7}} & \multicolumn{1}{c}{5} & 2.7e-1 & 6.11e-3 & 3.29e-1 & 7.55e-3 & 2.81e-1 & 1.68e-2 & 2.98e-1 & 4.14e-2 & 2.39e-1 & 5.41e-3 & \cellcolor[rgb]{ .859,  .859,  .859}2.52e-1 & 1.08e-2 & 2.9e0 & 4.31e-1 & 2.11e-1 & 3.92e-3 \\
          & \multicolumn{1}{c}{10} & 8.53e-1 & 3.69e-2 & 1.57e0 & 9.55e-2 & 1.18e0 & 9.98e-2 & \cellcolor[rgb]{ .859,  .859,  .859}7.8e-1 & 1.71e-2 & 1.32e0 & 6.79e-2 & 8.2e-1 & 2.91e-2 & 2.44e0 & 5.29e-1 & 9.09e-1 & 1.36e-1 \\
          & \multicolumn{1}{c}{15} & 2.16e0 & 1.67e-1 & 4.06e0 & 6.97e-1 & 3.17e0 & 4.49e-1 & \cellcolor[rgb]{ .859,  .859,  .859}1.5e0 & 6.32e-2 & 4.89e0 & 2.14e-1 & 1.68e0 & 9.87e-2 & 3.41e0 & 3.06e-2 & 1.77e0 & 4.43e-1 \\
    \multirow{3}[0]{*}{MaF8} & \multicolumn{1}{c}{5} & \cellcolor[rgb]{ .859,  .859,  .859}7.88e-2 & 2.65e-3 & 1.29e-1 & 4.45e-3 & 1.5e-1 & 1.74e-2 & 1.13e-1 & 2.71e-3 & 9.8e-2 & 2.33e-3 & 1.45e-1 & 3.41e-2 & 3.32e-1 & 2.85e-2 & 2.67e-1 & 5.42e-2 \\
          & \multicolumn{1}{c}{10} & 1.46e-1 & 4.02e-3 & 1.37e-1 & 4.17e-3 & 3.28e-1 & 6.06e-2 & \cellcolor[rgb]{ .859,  .859,  .859}1.16e-1 & 2.46e-3 & 1.35e-1 & 3.02e-3 & 1.48e-1 & 1.24e-2 & 9.01e-1 & 1.19e-2 & 9.74e-1 & 1.68e-1 \\
          & \multicolumn{1}{c}{15} & 1.9e-1 & 9.46e-3 & 1.73e-1 & 6.24e-3 & 3.61e-1 & 6.02e-2 & \cellcolor[rgb]{ .859,  .859,  .859}1.12e-1 & 1.55e-3 & 1.57e-1 & 2.35e-3 & 1.81e-1 & 9.64e-3 & 1.33e0 & 1.7e-2 & 1.45e0 & 3.2e-1 \\
    \multicolumn{1}{c}{\multirow{3}[0]{*}{MaF9}} & \multicolumn{1}{c}{5} & \cellcolor[rgb]{ .859,  .859,  .859}7.87e-2 & 5.9e-3 & 1.27e-1 & 8.24e-3 & 2.43e-1 & 1.13e-1 & 1.21e-1 & 2.18e-2 & 1.27e0 & 3.76e-1 & 5.66e-1 & 2.22e-1 & 2.52e-1 & 6.98e-3 & 1.87e-1 & 2.98e-2 \\
          & \multicolumn{1}{c}{10} & 3.38e-1 & 6.58e-2 & \cellcolor[rgb]{ .859,  .859,  .859}1.77e-1 & 8.2e-3 & 5.98e-1 & 2.14e-1 & 1.06e0 & 8.55e-2 & 1.37e0 & 1.43e-1 & 1.33e2 & 2.15e2 & 4.44e-1 & 1.93e-2 & 9.31e-1 & 2.16e-1 \\
          & \multicolumn{1}{c}{15} & 3.48e-1 & 1.45e-1 & 1.51e-1 & 5.4e-3 & 2.65e0 & 4.63e0 & \cellcolor[rgb]{ .859,  .859,  .859}1.07e-1 & 1.18e-3 & 3.0e0 & 3.81e0 & 1.13e0 & 2.69e0 & 9.52e-1 & 1.34e-2 & 1.14e0 & 2.02e-1 \\
    \multirow{3}[0]{*}{MaF10} & \multicolumn{1}{c}{5} & \cellcolor[rgb]{ .859,  .859,  .859}3.73e-1 & 8.63e-3 & 4.71e-1 & 1.09e-2 & 4.59e-1 & 3.35e-2 & 4.37e-1 & 1.25e-2 & 4.01e-1 & 1.0e-2 & 4.11e-1 & 1.18e-2 & 7.69e-1 & 1.2e-1 & 6.3e-1 & 7.66e-2 \\
          & \multicolumn{1}{c}{10} & 1.05e0 & 3.73e-2 & 1.2e0 & 5.94e-2 & 1.07e0 & 4.19e-2 & \cellcolor[rgb]{ .859,  .859,  .859}9.69e-1 & 1.51e-2 & 1.06e0 & 3.77e-2 & 1.17e0 & 8.8e-2 & 1.94e0 & 4.57e-2 & 1.35e0 & 9.24e-2 \\
          & \multicolumn{1}{c}{15} & 1.52e0 & 6.54e-2 & 1.96e0 & 8.53e-2 & 1.76e0 & 5.59e-1 & \cellcolor[rgb]{ .859,  .859,  .859}1.42e0 & 2.35e-2 & 1.82e0 & 4.02e-2 & 1.69e0 & 1.01e-1 & 2.4e0 & 1.05e-1 & 2.0e0 & 6.39e-2 \\
    \multicolumn{1}{c}{\multirow{3}[0]{*}{MaF11}} & \multicolumn{1}{c}{5} & \cellcolor[rgb]{ .859,  .859,  .859}6.35e-1 & 1.26e-2 & 8.25e-1 & 1.72e-2 & 1.04e0 & 5.0e-1 & 6.77e-1 & 4.53e-2 & 1.13e0 & 4.29e-1 & 1.54e0 & 5.71e-1 & 4.13e0 & 3.9e-1 & 2.42e0 & 8.17e-1 \\
          & \multicolumn{1}{c}{10} & 3.0e0 & 1.28e0 & 2.56e0 & 4.39e-1 & 5.7e0 & 6.26e-1 & \cellcolor[rgb]{ .859,  .859,  .859}2.26e0 & 2.02e-1 & 3.39e0 & 8.11e-1 & 5.67e0 & 1.1e0 & 1.53e1 & 2.98e-2 & 1.06e1 & 2.2e0 \\
          & \multicolumn{1}{c}{15} & 8.52e0 & 2.79e0 & 4.68e-1 & 6.22e-1 & 1.29e1 & 1.6e0 & \cellcolor[rgb]{ .859,  .859,  .859}2.18e-1 & 2.08e-1 & 5.56e0 & 9.24e-1 & 9.66e0 & 1.91e0 & 2.57e1 & 3.58e-1 & 1.5e1 & 2.36e0 \\
    \multirow{3}[0]{*}{MaF12} & \multicolumn{1}{c}{5} & \cellcolor[rgb]{ .859,  .859,  .859}9.33e-1 & 2.51e-3 & 1.12e0 & 7.95e-3 & 9.4e-1 & 8.33e-3 & 1.09e0 & 1.15e-2 & 9.38e-1 & 6.34e-3 & 1.06e0 & 1.72e-2 & 1.29e0 & 1.19e-2 & 9.68e-1 & 9.86e-3 \\
          & \multicolumn{1}{c}{10} & 4.35e0 & 2.06e-2 & 4.69e0 & 1.44e-2 & 4.55e0 & 2.34e-1 & 4.25e0 & 3.09e-2 & \cellcolor[rgb]{ .859,  .859,  .859}3.99e0 & 2.36e-2 & 4.65e0 & 5.44e-2 & 6.6e0 & 1.99e-1 & 4.49e0 & 4.47e-2 \\
          & \multicolumn{1}{c}{15} & 7.56e0 & 1.58e-1 & 7.62e0 & 1.66e-1 & 8.48e0 & 3.55e-1 & 7.7e0 & 9.24e-2 & \cellcolor[rgb]{ .859,  .859,  .859}7.08e0 & 7.4e-2 & 7.5e0 & 1.81e-1 & 8.71e0 & 3.19e-1 & 8.4e0 & 1.47e-1 \\
    \multicolumn{1}{c}{\multirow{3}[0]{*}{MaF13}} & \multicolumn{1}{c}{5} & \cellcolor[rgb]{ .859,  .859,  .859}1.2e-1 & 1.19e-2 & 1.3e-1 & 6.74e-3 & 1.63e-1 & 1.72e-2 & 1.6e-1 & 1.83e-2 & 1.67e-1 & 2.87e-2 & 1.58e-1 & 2.32e-2 & 2.34e-1 & 3.1e-2 & 4.51e-1 & 9.0e-2 \\
          & \multicolumn{1}{c}{10} & 2.17e-1 & 1.15e-2 & \cellcolor[rgb]{ .859,  .859,  .859}1.22e-1 & 7.21e-3 & 2.3e-1 & 2.0e-2 & 1.62e-1 & 1.44e-2 & 1.85e-1 & 1.92e-2 & 1.9e-1 & 3.0e-2 & 3.4e-1 & 2.85e-2 & 4.52e-1 & 9.87e-2 \\
          & \multicolumn{1}{c}{15} & 2.49e-1 & 3.9e-2 & \cellcolor[rgb]{ .859,  .859,  .859}1.5e-1 & 9.25e-3 & 2.51e-1 & 3.13e-2 & 1.62e-1 & 1.9e-2 & 2.37e-1 & 3.89e-2 & 2.12e-1 & 2.58e-2 & 3.69e-1 & 3.84e-2 & 6.38e-1 & 7.86e-2 \\
    \multirow{3}[0]{*}{MaF14} & \multicolumn{1}{c}{5} & \cellcolor[rgb]{ .859,  .859,  .859}3.5e-1 & 2.28e-2 & 3.85e-1 & 4.24e-2 & 7.14e-1 & 2.36e-1 & 1.44e0 & 4.96e-1 & 6.58e-1 & 2.87e-1 & 5.07e-1 & 6.83e-2 & 3.71e-1 & 3.38e-2 & 5.26e-1 & 8.67e-2 \\
          & \multicolumn{1}{c}{10} & 5.54e-1 & 5.09e-2 & 6.76e-1 & 8.35e-2 & 2.34e0 & 1.31e0 & 1.15e0 & 8.76e-2 & 9.78e-1 & 1.35e-1 & 1.36e0 & 8.1e-1 & \cellcolor[rgb]{ .859,  .859,  .859}5.36e-1 & 5.21e-2 & 7.22e-1 & 5.71e-2 \\
          & \multicolumn{1}{c}{15} & 6.19e-1 & 1.33e-1 & 6.04e-1 & 6.46e-2 & 1.32e0 & 2.61e-1 & 1.13e0 & 9.68e-2 & 1.11e0 & 1.76e-1 & 1.24e0 & 8.84e-1 & \cellcolor[rgb]{ .859,  .859,  .859}5.56e-1 & 1.22e-1 & 8.92e-1 & 1.3e-1 \\
    \multicolumn{1}{c}{\multirow{3}[1]{*}{MaF15}} & \multicolumn{1}{c}{5} & \cellcolor[rgb]{ .859,  .859,  .859}3.68e-1 & 4.02e-2 & 6.13e-1 & 2.57e-2 & 9.94e-1 & 9.4e-2 & 1.11e0 & 3.65e-2 & 8.86e-1 & 8.99e-2 & 1.16e0 & 9.25e-2 & 5.99e-1 & 1.54e-2 & 6.17e-1 & 4.06e-2 \\
          & \multicolumn{1}{c}{10} & 9.84e-1 & 4.72e-2 & 8.52e-1 & 5.04e-2 & 1.53e0 & 2.67e-1 & 3.37e0 & 1.53e0 & \cellcolor[rgb]{ .859,  .859,  .859}7.89e-1 & 6.42e-2 & 1.3e0 & 4.73e-2 & 1.01e0 & 8.98e-2 & 9.81e-1 & 6.8e-2 \\
          & \multicolumn{1}{c}{15} & \cellcolor[rgb]{ .859,  .859,  .859}1.04e0 & 4.51e-2 & 1.35e0 & 7.91e-2 & 3.89e0 & 2.03e0 & 1.67e0 & 1.39e0 & 1.3e0 & 8.51e-3 & 1.52e0 & 4.85e-2 & 1.14e0 & 3.58e-2 & 1.27e0 & 5.4e-2 \\
    \midrule
    \multicolumn{1}{c}{\multirow{4}[2]{*}{Friedman}} & \multicolumn{1}{c}{5} & \multicolumn{2}{c}{\textbf{1.27}} & \multicolumn{2}{c}{4.60} & \multicolumn{2}{c}{4.80} & \multicolumn{2}{c}{4.27} & \multicolumn{2}{c}{4.20} & \multicolumn{2}{c}{5.00} & \multicolumn{2}{c}{6.80} & \multicolumn{2}{c}{5.07} \\
          & \multicolumn{1}{c}{10} & \multicolumn{2}{c}{\textbf{3.13}} & \multicolumn{2}{c}{4.20} & \multicolumn{2}{c}{6.13} & \multicolumn{2}{c}{3.20} & \multicolumn{2}{c}{3.27} & \multicolumn{2}{c}{4.60} & \multicolumn{2}{c}{6.07} & \multicolumn{2}{c}{5.40} \\
          & \multicolumn{1}{c}{15} & \multicolumn{2}{c}{3.27} & \multicolumn{2}{c}{4.13} & \multicolumn{2}{c}{5.33} & \multicolumn{2}{c}{\textbf{2.73}} & \multicolumn{2}{c}{5.60} & \multicolumn{2}{c}{3.87} & \multicolumn{2}{c}{5.87} & \multicolumn{2}{c}{5.20} \\
          & all   & \multicolumn{2}{c}{\textbf{2.42}} & \multicolumn{2}{c}{4.36} & \multicolumn{2}{c}{5.71} & \multicolumn{2}{c}{3.49} & \multicolumn{2}{c}{3.96} & \multicolumn{2}{c}{4.56} & \multicolumn{2}{c}{6.22} & \multicolumn{2}{c}{5.29} \\
    \midrule
    \multicolumn{1}{c}{\multirow{4}[2]{*}{$t$-test}} & \multicolumn{1}{c}{5} & \multicolumn{2}{c}{\multirow{4}[2]{*}{~}} & \multicolumn{2}{c}{\textbf{15/0/0}} & \multicolumn{2}{c}{\textbf{15/0/0}} & \multicolumn{2}{c}{\textbf{15/0/0}} & \multicolumn{2}{c}{\textbf{11/2/2}} & \multicolumn{2}{c}{\textbf{14/0/1}} & \multicolumn{2}{c}{\textbf{14/1/0}} & \multicolumn{2}{c}{\textbf{13/1/1}} \\
          & \multicolumn{1}{c}{10} & \multicolumn{2}{c}{} & \multicolumn{2}{c}{\textbf{8/3/4}} & \multicolumn{2}{c}{\textbf{12/3/0}} & \multicolumn{2}{c}{7/1/7} & \multicolumn{2}{c}{7/1/7} & \multicolumn{2}{c}{\textbf{7/4/4}} & \multicolumn{2}{c}{\textbf{13/2/0}} & \multicolumn{2}{c}{\textbf{11/3/1}} \\
          & \multicolumn{1}{c}{15} & \multicolumn{2}{c}{} & \multicolumn{2}{c}{\textbf{9/2/4}} & \multicolumn{2}{c}{\textbf{1/4/0}} & \multicolumn{2}{c}{6/0/9} & \multicolumn{2}{c}{\textbf{9/2/4}} & \multicolumn{2}{c}{\textbf{7/2/6}} & \multicolumn{2}{c}{\textbf{14/0/1}} & \multicolumn{2}{c}{\textbf{12/1/2}} \\
          & all   & \multicolumn{2}{c}{} & \multicolumn{2}{c}{\textbf{32/5/8}} & \multicolumn{2}{c}{\textbf{38/7/0}} & \multicolumn{2}{c}{\textbf{28/1/16}} & \multicolumn{2}{c}{\textbf{27/5/13}} & \multicolumn{2}{c}{\textbf{28/6/11}} & \multicolumn{2}{c}{\textbf{41/3/1}} & \multicolumn{2}{c}{\textbf{36/5/4}} \\
    \bottomrule
    \bottomrule
    \end{tabular}%
  \label{tab:IGD}%
\end{table*}%

\begin{table*}[!t]
\setlength{\tabcolsep}{3.5pt}
\renewcommand\arraystretch{0.8}
  \centering
  \caption{Runtime averaged from 20 independent runs of all compared algorithms}
    \begin{tabular}{cccccccccccccccc}
    \toprule
    \toprule
          & \algoabbr{}  & \multicolumn{2}{c}{AR-MOEA} & \multicolumn{2}{c}{A-NSGA-III} & \multicolumn{2}{c}{Two\_Arch2} & \multicolumn{2}{c}{GrEA} & \multicolumn{2}{c}{KnEA} & \multicolumn{2}{c}{MOEA/DD} & \multicolumn{2}{c}{RVEA*} \\
    \midrule
    Objectives & Time  & Time  & Ratio & Time  & Ratio & Time  & Ratio & Time  & Ratio & Time  & Ratio & Time  & Ratio & Time  & Ratio \\
    M5/min & 6.27  & 21.57 & 3.44x & 4.41  & 0.70x & 53.90 & 8.60x & 29.20 & 4.66x & 5.60  & 0.89x & 96.74 & 15.43x & 4.01  & 0.64x \\
    M10/min & 9.98  & 155.88 & 15.62x & 9.12  & 0.91x & 320.37 & 32.10x & 78.10 & 7.83x & 9.97  & 1.00x & 247.01 & 24.75x & 8.65  & 0.87x \\
    M15/min & 18.05 & 274.18 & 15.19x & 15.79 & 0.87x & 622.79 & 34.50x & 127.04 & 7.04x & 19.33 & 1.07x & 380.09 & 21.06x & 14.62 & 0.81x \\
    All/h & 0.57  & 7.53  & 13.17x & 0.49  & 0.85x & 16.62 & 29.07x & 3.91  & 6.83x & 0.58  & 1.02x & 12.06 & 21.10x & 0.45  & 0.80x \\
    \bottomrule
    \bottomrule
    \end{tabular}%
  \label{tab:runtime}%
\end{table*}%

\begin{figure*}[!t]
\centering
\captionsetup{justification = centering}
\includegraphics[width=\textwidth]{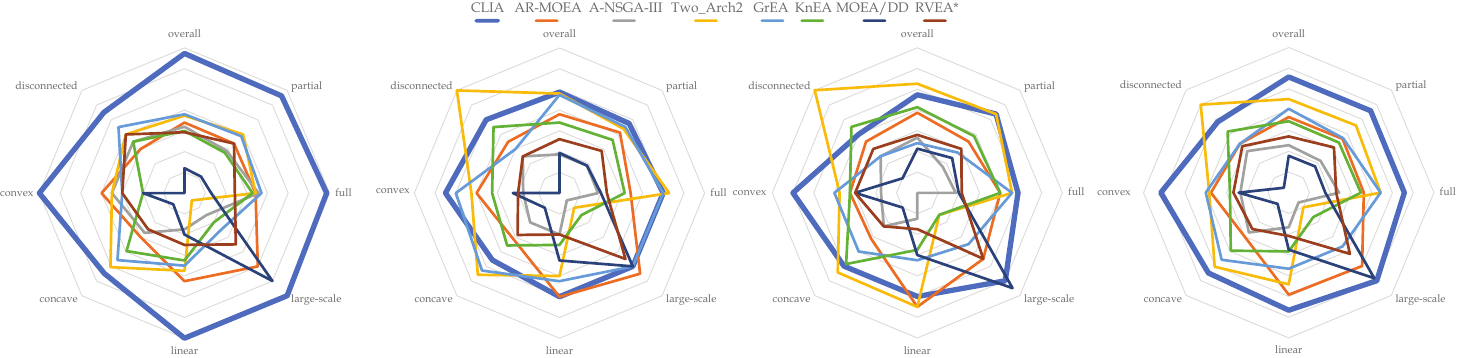}
\caption{The radar diagrams for the specified performance on each category of MaF problems. The first three radar charts are the categorized performance on $M=5$, $M=10$ and $M=15$ respectively. The fourth chart shows the overall performance. The data in each dimension is $|A|-r$, where $r$ is the Friedman mean rank for such category and $|A|$ is the number of all algorithms compared, $|A|=8$. $F_{partial}=\{f_{1}, f_{2}, f_{4}, f_{6}, f_{7}, f_{8}, f_{9}, f_{13}, f_{15}\}$. $F_{full}=\{f_{3}, f_{5}, f_{10}, f_{11}, f_{12}, f_{14}\}$. $F_{largescale}=\{f_{14}, f_{15}\}$. $F_{linear}=\{f_{1}, f_{8}, f_{9}, f_{14}\}$. $F_{concave}=\{f_{2}, f_{4}, f_{5}, f_{6}, f_{10}, f_{12}, f_{13}\}$. $F_{convex}=\{f_{3}, f_{10}, f_{11}, f_{15}\}$. $F_{disconnected}=\{f_{7}, f_{11}\}$. The true PF of $f_{10}$ is hybrid in shape with concave and convex parts.}
\label{fig:radar}
\end{figure*}
\begin{figure*}[!t]
\centering
\subfloat[MaF1: population]{
\captionsetup{justification = centering}
\includegraphics[width=0.18\textwidth]{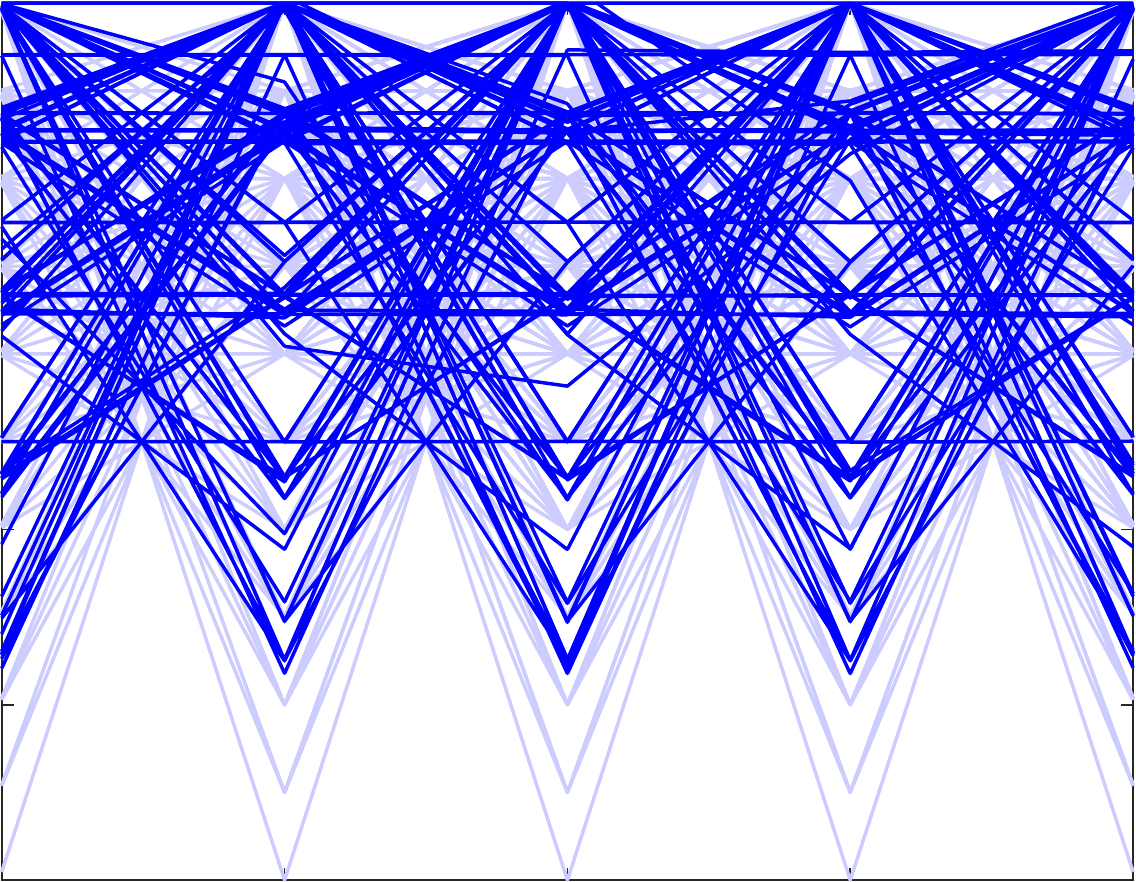}}
\hfill
\subfloat[MaF4: population]{
\captionsetup{justification = centering}
\includegraphics[width=0.18\textwidth]{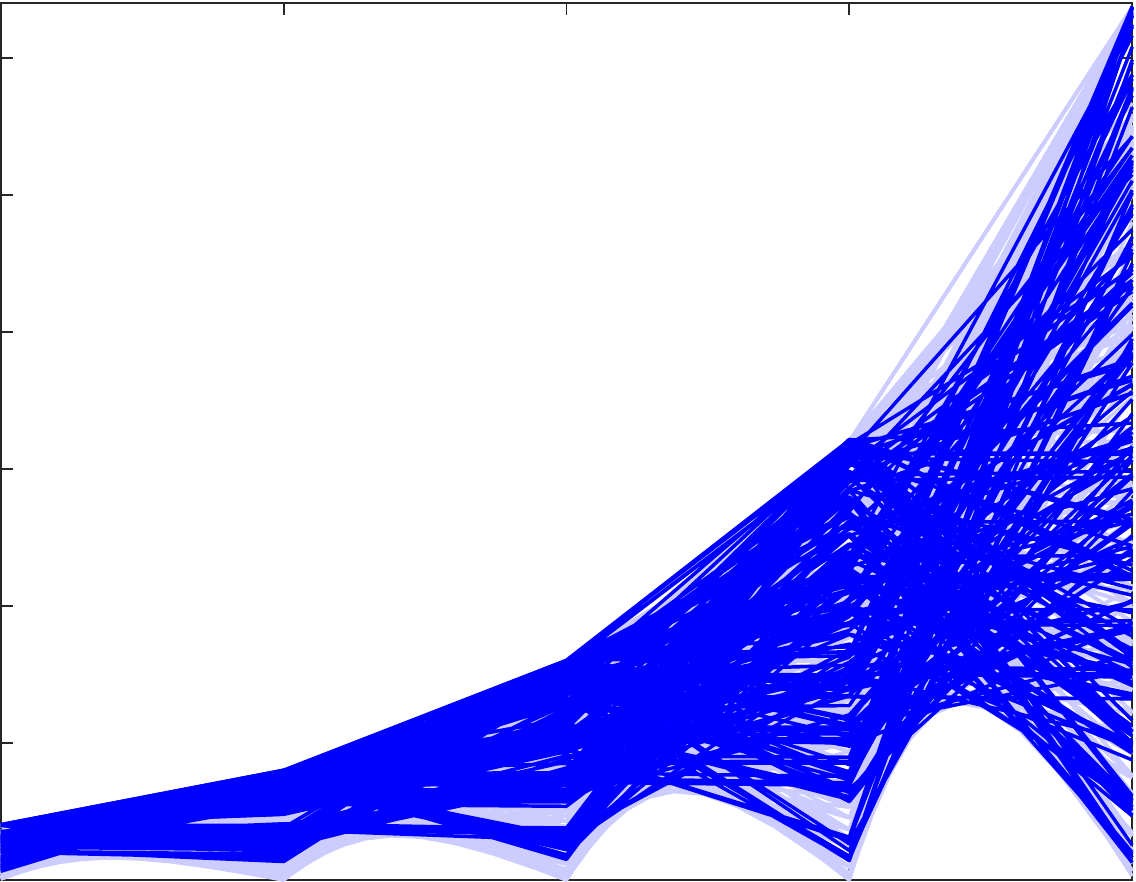}}
\hfill
\subfloat[MaF6: population]{
\captionsetup{justification = centering}
\includegraphics[width=0.18\textwidth]{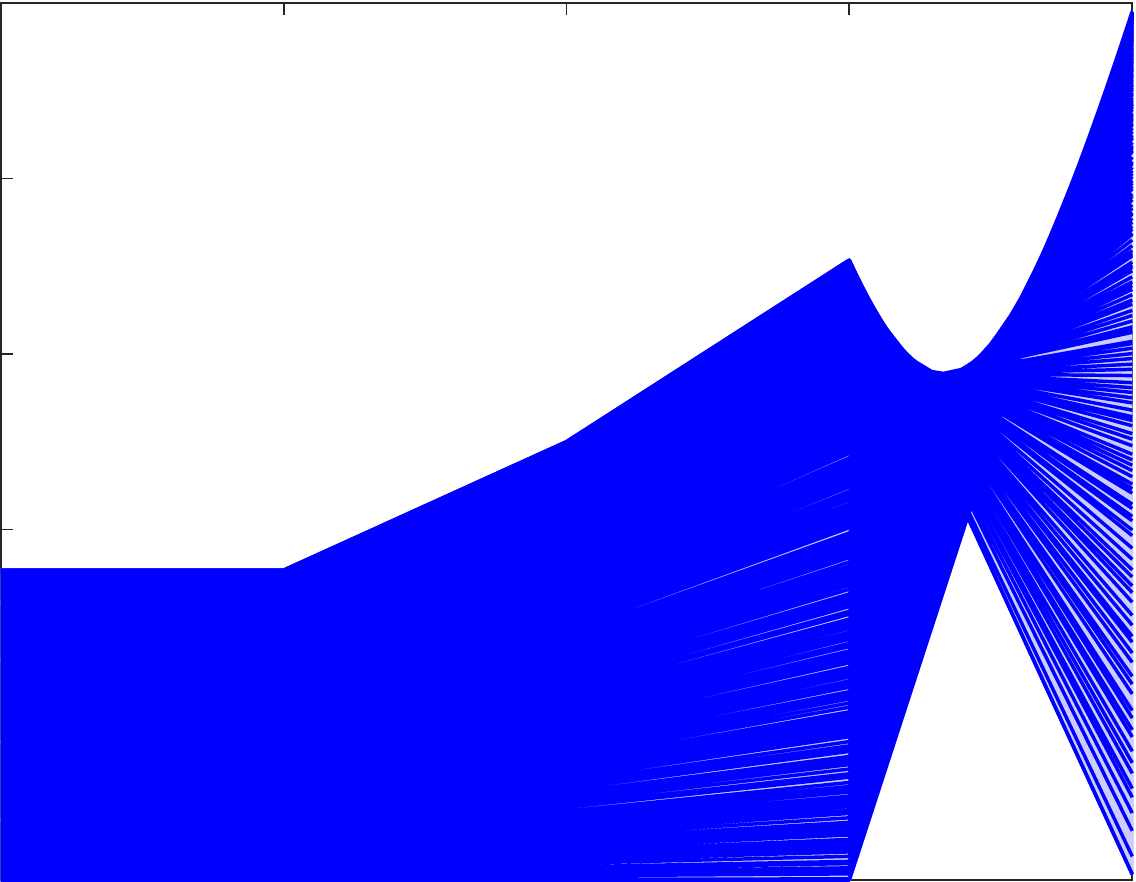}}
\hfill
\subfloat[MaF8: population]{
\captionsetup{justification = centering}
\includegraphics[width=0.18\textwidth]{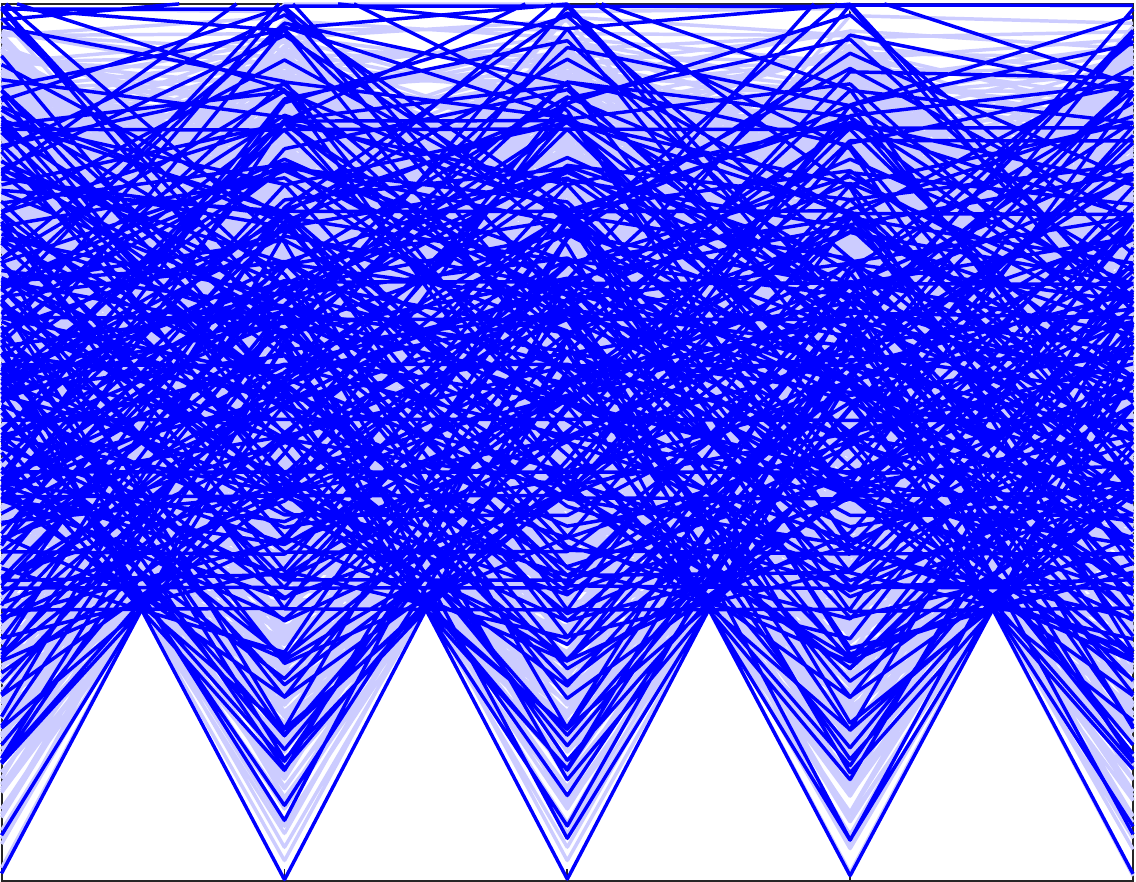}}
\hfill
\subfloat[MaF13: population]{
\captionsetup{justification = centering}
\includegraphics[width=0.18\textwidth]{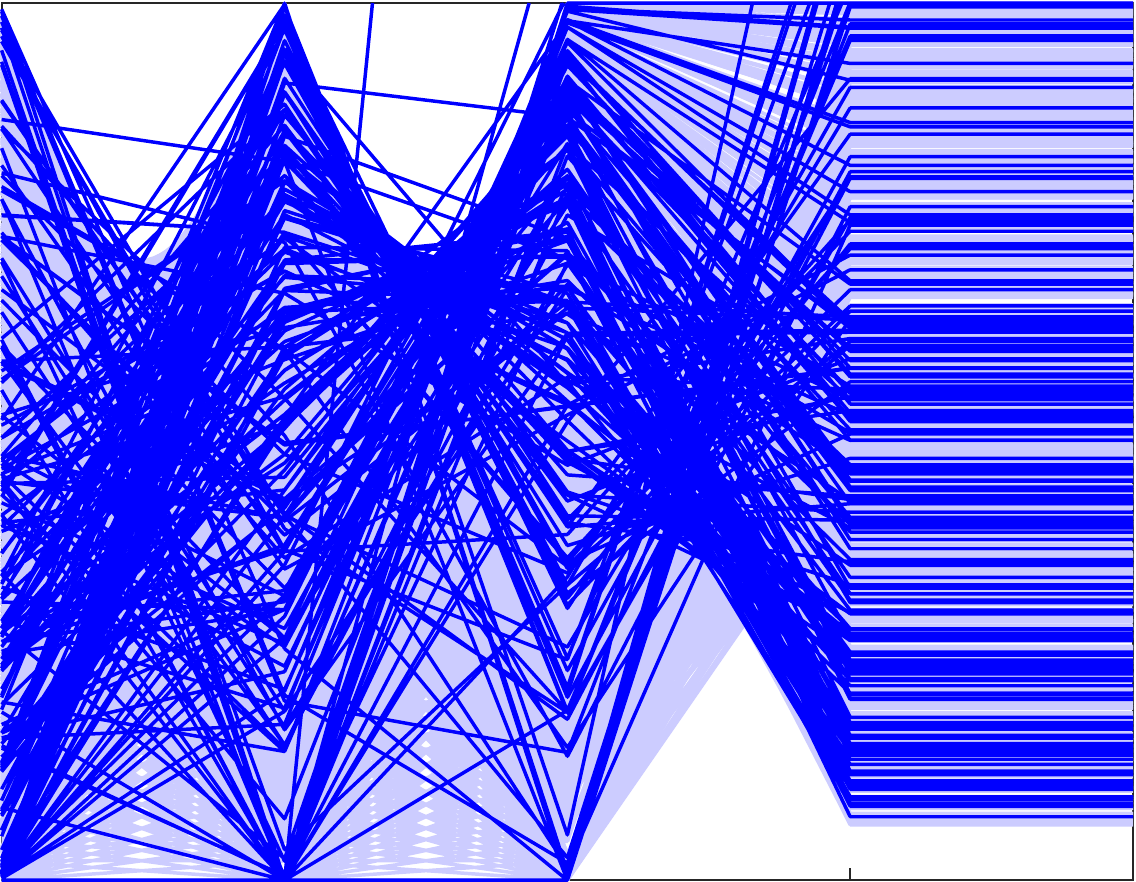}}

\subfloat[MaF1: reference points]{
\captionsetup{justification = centering}
\includegraphics[width=0.18\textwidth]{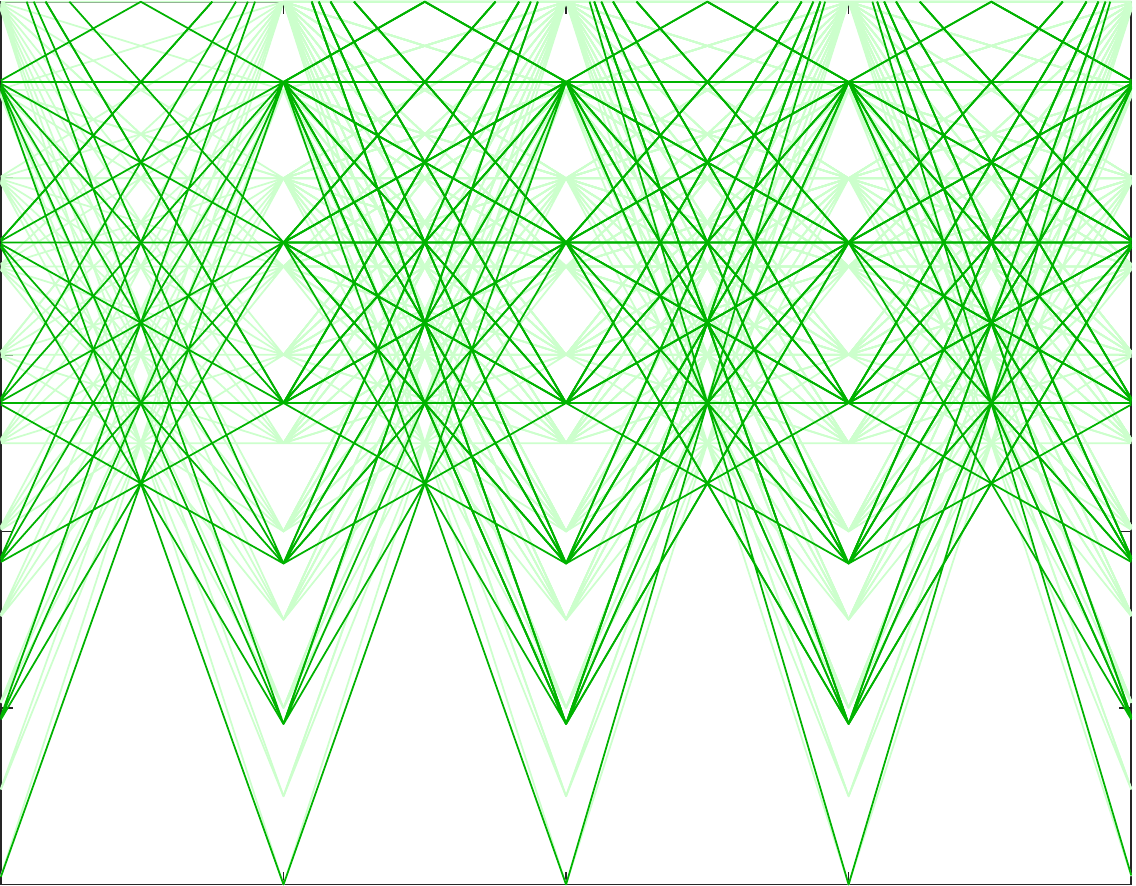}}
\hfill
\subfloat[MaF4: reference points]{
\captionsetup{justification = centering}
\includegraphics[width=0.18\textwidth]{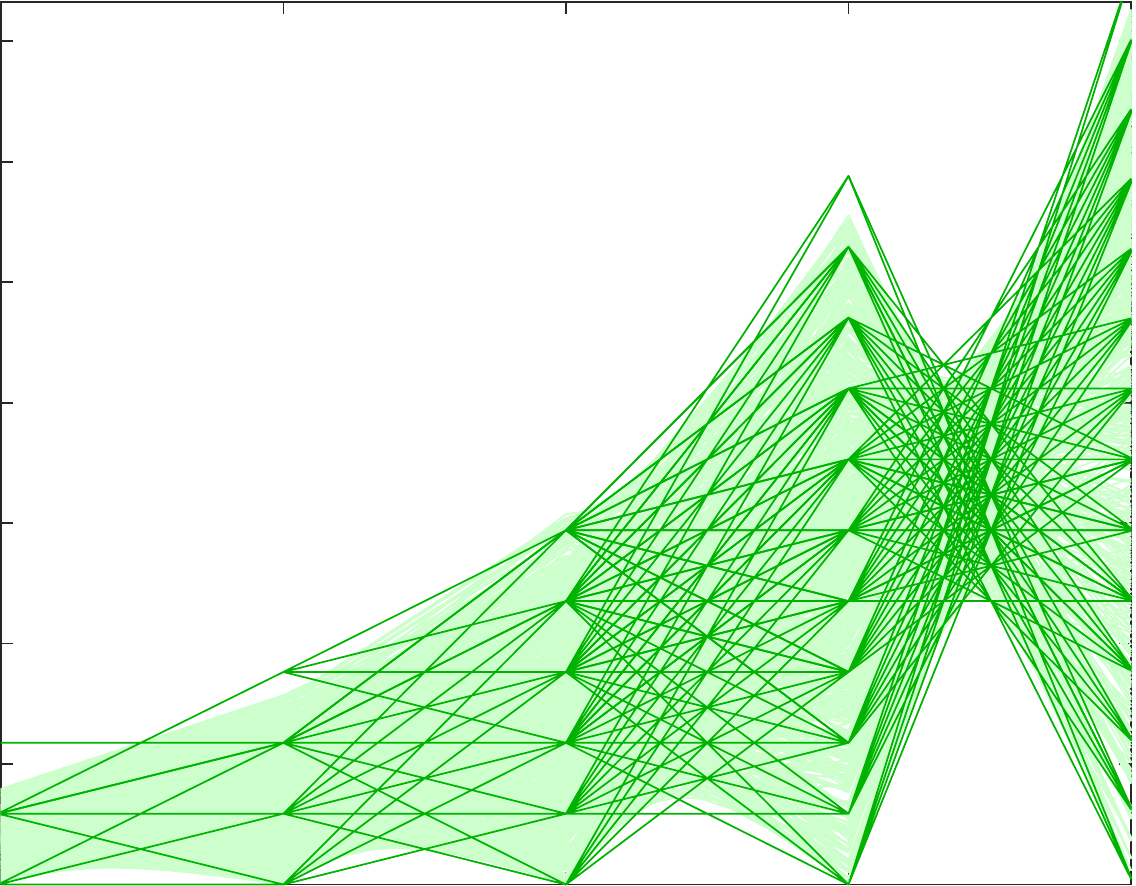}}
\hfill
\subfloat[MaF6: reference points]{
\captionsetup{justification = centering}
\includegraphics[width=0.18\textwidth]{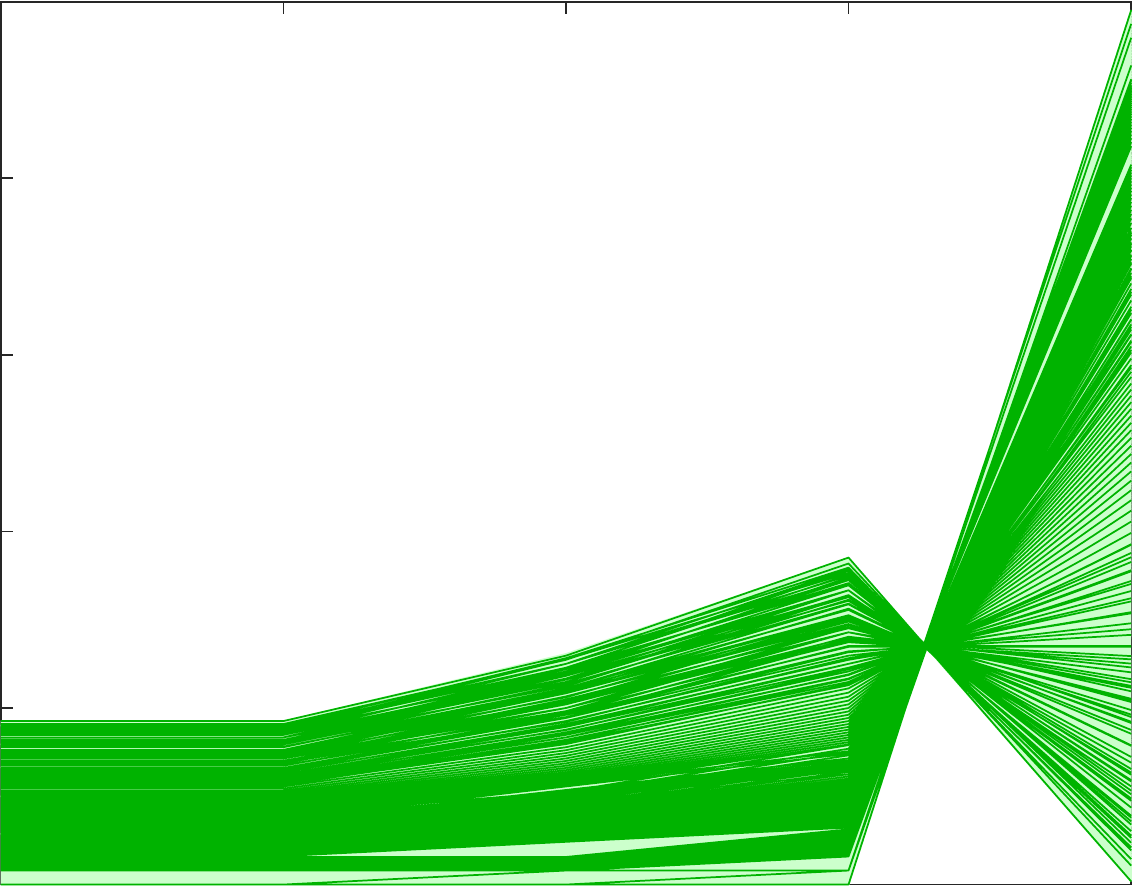}}
\hfill
\subfloat[MaF8: reference points]{
\captionsetup{justification = centering}
\includegraphics[width=0.18\textwidth]{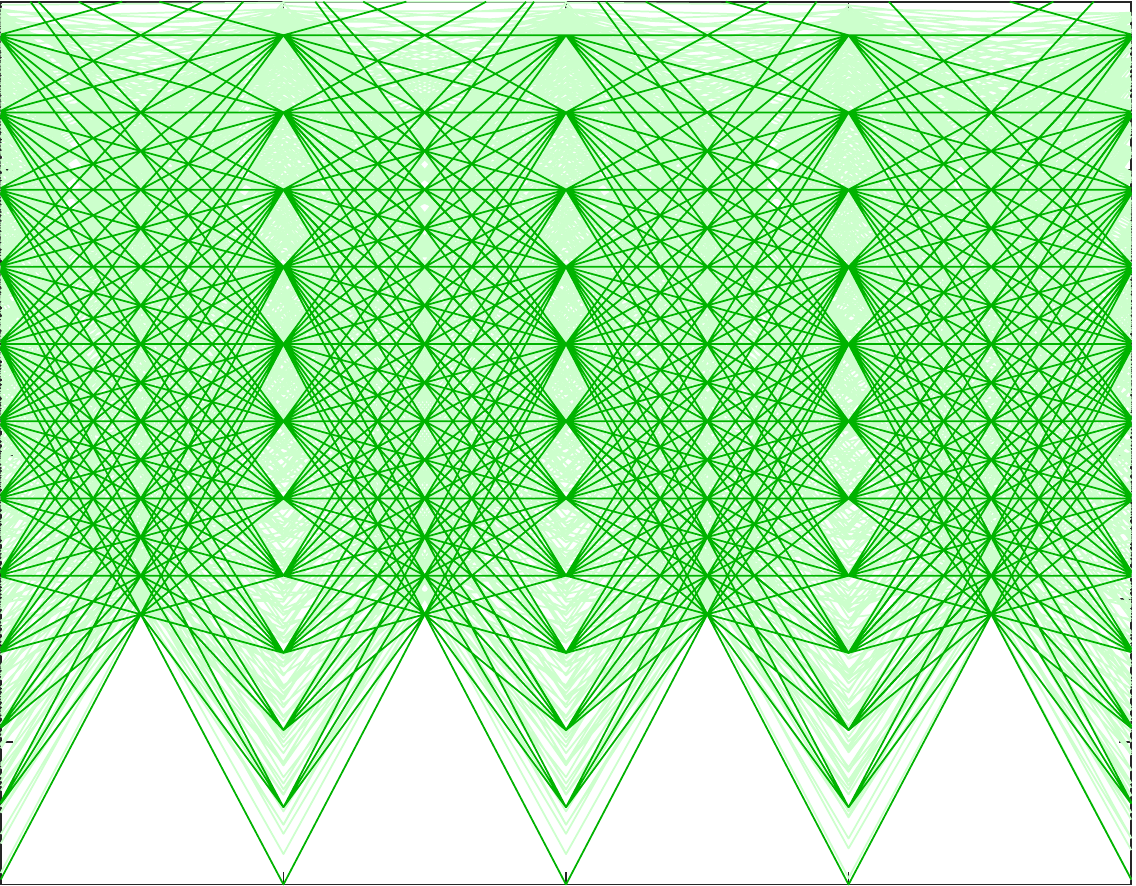}}
\hfill
\subfloat[MaF13: reference points]{
\captionsetup{justification = centering}
\includegraphics[width=0.18\textwidth]{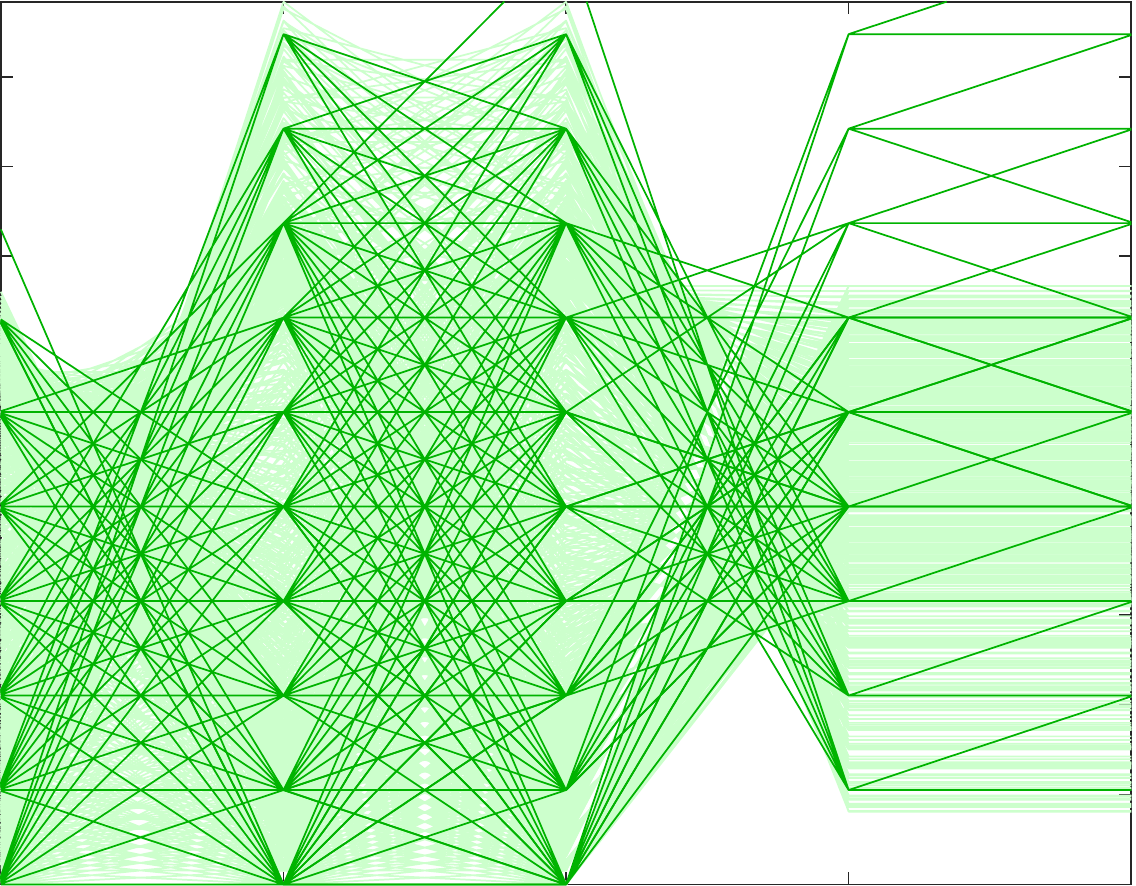}}

\caption{Coverage of the obtained population and the coverage of the obtained reference points.}
\label{fig:covering}
\end{figure*}
\par
In Table \ref{tab:IGD}, Friedman tests are conducted on the results, where the mean rankings and the detailed results are also presented. With the significance level of $\alpha=0.05$, $\chi^2=80.45$ and $p=1.12\times 10^{-14}$, the Friedman test shows that the respective mean rankings are effective for the significant differences in performance, where \algoabbr{} achieves the top overall ranking considering all 45 individual test cases by obtaining the best rankings on $M=5$ and $M=10$, and the second on $M=15$.
\par
Also, $t$-tests are conducted on the $7$ pairs of algorithms, each pair is \algoabbr{} and one compared algorithm. The results are presented as ``$l/u/g$,'' where $l$ represents the number of test cases where \algoabbr{} performs better statistically, $u$ represents the number of similar performances and $g$ represents the number of test cases where \algoabbr{} performs worse. From the $t$-test results, it can be observed that \algoabbr{} has outperformed the state-of-the-art algorithms in many test cases.
\subsubsection{Versatility for Diverse Characteristics}
The MaF suite is used to analyse the capabilities of algorithms handling diverse characteristics. Fig. \ref{fig:radar} presents the radar diagrams specifying the rankings of the performance of the algorithms on each type of problems. With these diagrams, we can compare the performances of the algorithms in an intuitive way.
\par
For $M=5$, the performance of \algoabbr{} is good, with the best performance on each category, demonstrating exciting capabilities to deal with diverse characteristics with a medium-scale number of objectives. For $M=10$, the deterioration in performance on some categories can be observed, though \algoabbr{} still achieves the top performance. For $M=15$, deterioration in performance made \algoabbr{} lose its leading position in the overall performance. We ascribe the shift of the leadership to Two\_Arch2 \cite{wang2015improved} to the utilization of the Diversity Archive (DA) that is able to maintain the diversity in such high-dimensional objective space. In the simulations, we find that even the appropriate number of uniformly distributed reference vectors intersecting with the true PF cannot lead to the IGDs better than the ones obtained by Two\_Arch2, which indicates that to have better performance in the higher-dimensional objective spaces, the adaptation should be able to adapt to the curvatures as well.
\par
Report \cite{cheng2017benchmark} indicates that for problems with partial true PFs, algorithms based on reference vectors should effectively adapt to the true PF characteristics to ensure the performance. On problems with partial PFs, where the other two reference vector adaptation-based algorithms struggle, \algoabbr{} obtains the best overall performance, with the best performance among the state-of-the-art algorithms when $M=5$ and $M=10$ and the second best when $M=15$. The deterioration in performance may be due to the difficulties for accurate learning in the high dimensional objective spaces and the sparse gaps between the reference vectors and the exact boundaries of the true PF, which is further investigated in the supplementary file. Though this kind of deterioration indicates that the incremental learning process has limits for scalability, the competitive performance achieved by the clustering-learning interactions is still encouraging. For a better visualization of the performance on partial PFs, in each of the first five diagrams of Fig. \ref{fig:covering}, the obtained population with the median IGD is presented with the true PF as background in the parallel coordinates. In each of the last five diagrams, the activated reference points of the corresponding population are presented with the projections of the true PF on the unit simplex (the effective areas) as background. The roughly evenly distributed population reflects the effectiveness of the clustering-learning interactions. Also, the evenly distributed active reference points over the effective areas show that the \MakeLowercase{\learningprocessfullname{}} can provide effective reference vectors without disturbing the uniformity. The addressing of the disturbance of the uniformity is further discussed in the supplementary file.
\par
Convex PFs may bring difficulties to the reference vector-based algorithms in the uniformity maintenance \cite{cheng2017benchmark, li2016pareto}. It is good to see that \algoabbr{} has the best performance on the problems with convex PFs, even on the problems with partial convex PFs. The good performance can be also observed in the later validation of PDM.
\par
On the large scale problems with hundreds of decision variables and huge amount of FEs provided, the good performance of \algoabbr{} demonstrates that the evolution pressure provided by the selection operator of \MakeLowercase{\clusteringprocessfullname{}} is satisfactory, even without additional efforts on the offspring generation that may greatly boost the qualities of the population \cite{he2017coordinated}.
\par
In the categorized analyses, \algoabbr{} demonstrates competitive performance on problems with diverse characteristics.
\subsubsection{Runtime Comparison and Analysis}
Table \ref{tab:runtime} provides the runtime of the compared algorithms as well as their runtime ratios to \algoabbr{}. The ``runtime'' sums the averaged total runtime for running all $15$ problems with certain $M$.
\par
The results show that \algoabbr{} has relatively fast overall runtime. The efficiency of \algoabbr{} can be highlighted, as high performance is obtained in fair runtime.
\subsubsection{Stability and Convergence Analyses Using IGD Curves}
\begin{figure*}[!t]
\centering
\subfloat[MaF3, $M=15$]{
\captionsetup{justification = centering}
\includegraphics[width=0.25\textwidth]{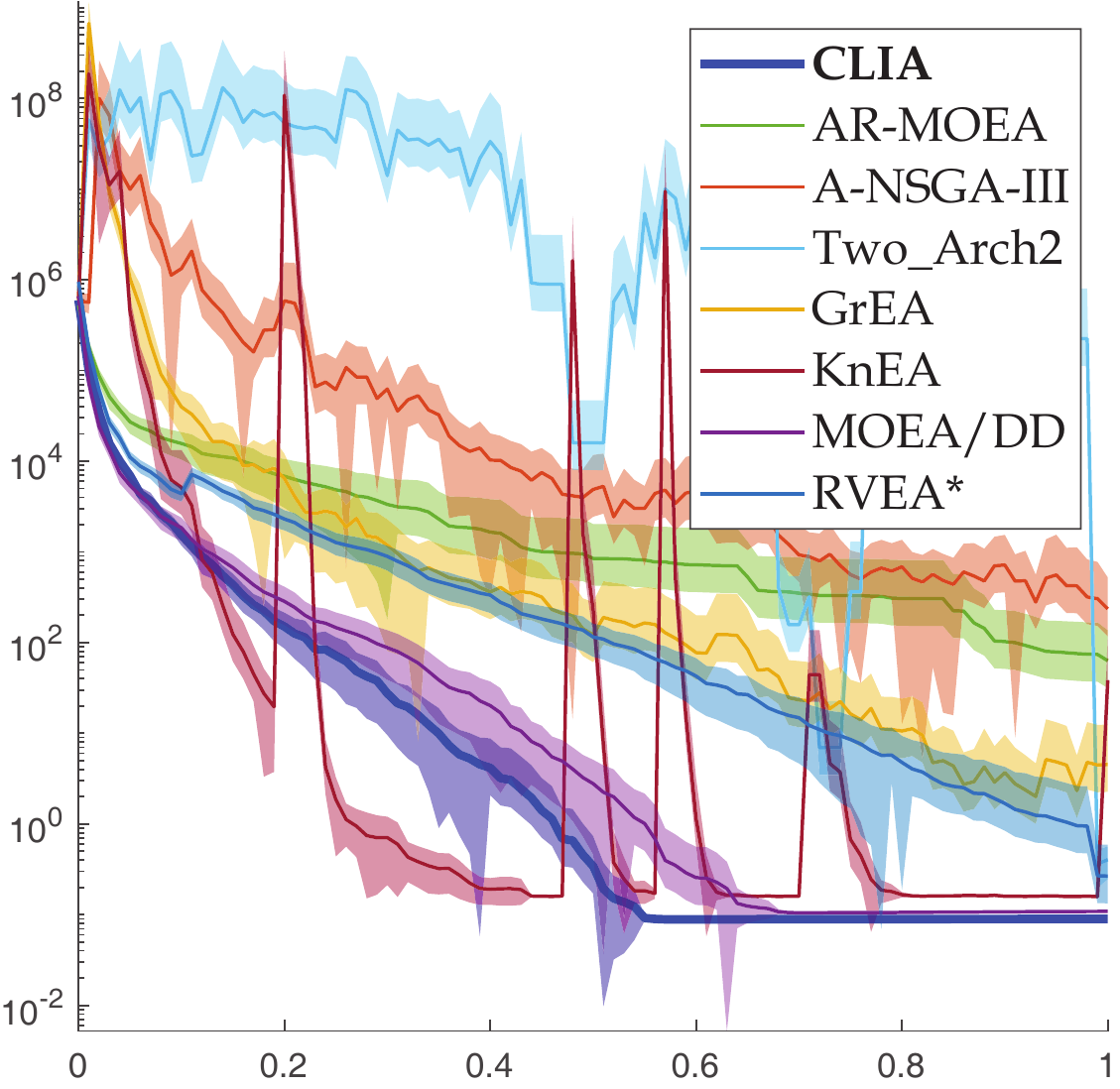}}
\hfill
\subfloat[MaF6, $M=15$]{
\captionsetup{justification = centering}
\includegraphics[width=0.25\textwidth]{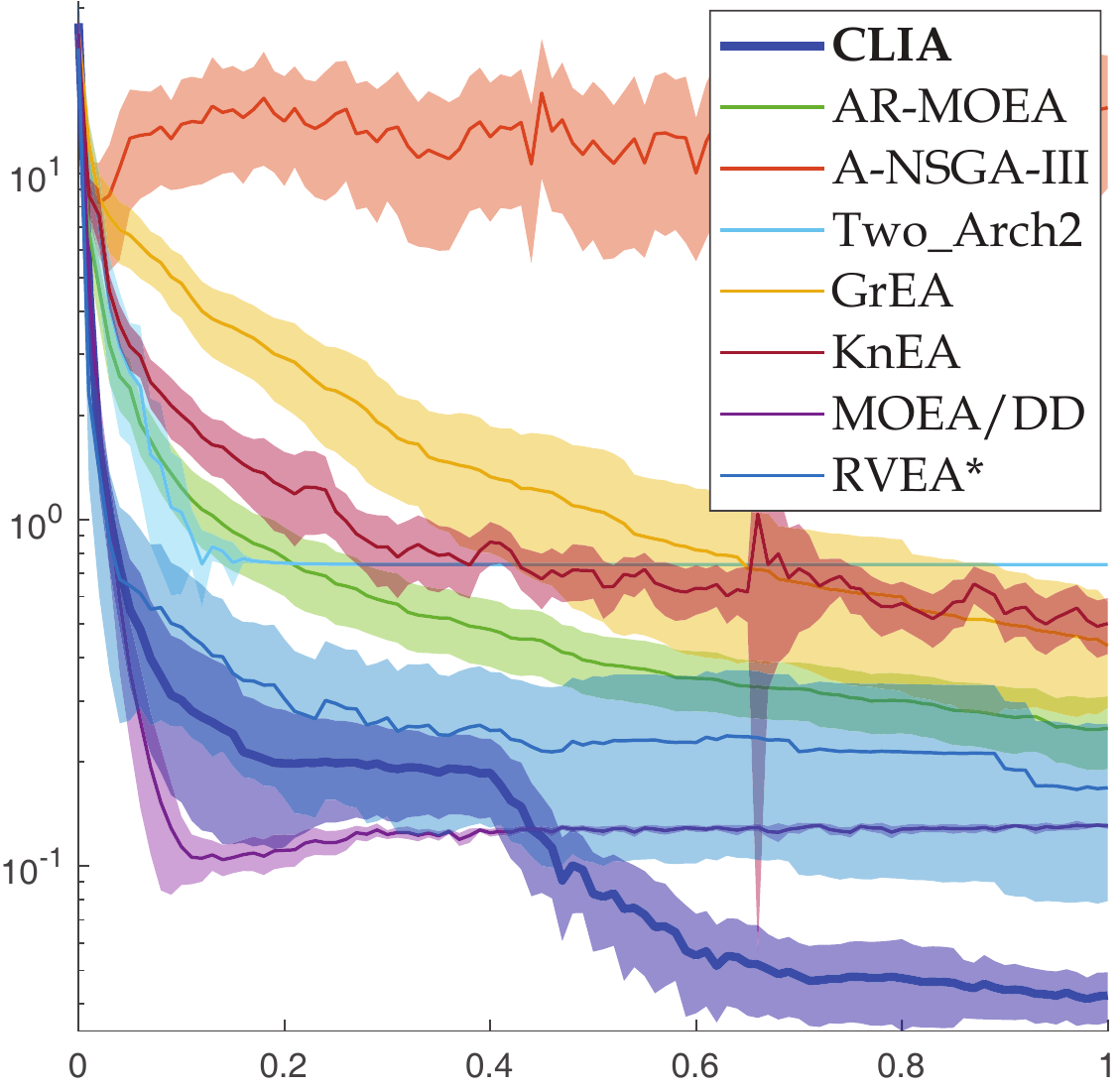}}
\hfill
\subfloat[MaF15, $M=15$]{
\captionsetup{justification = centering}
\includegraphics[width=0.25\textwidth]{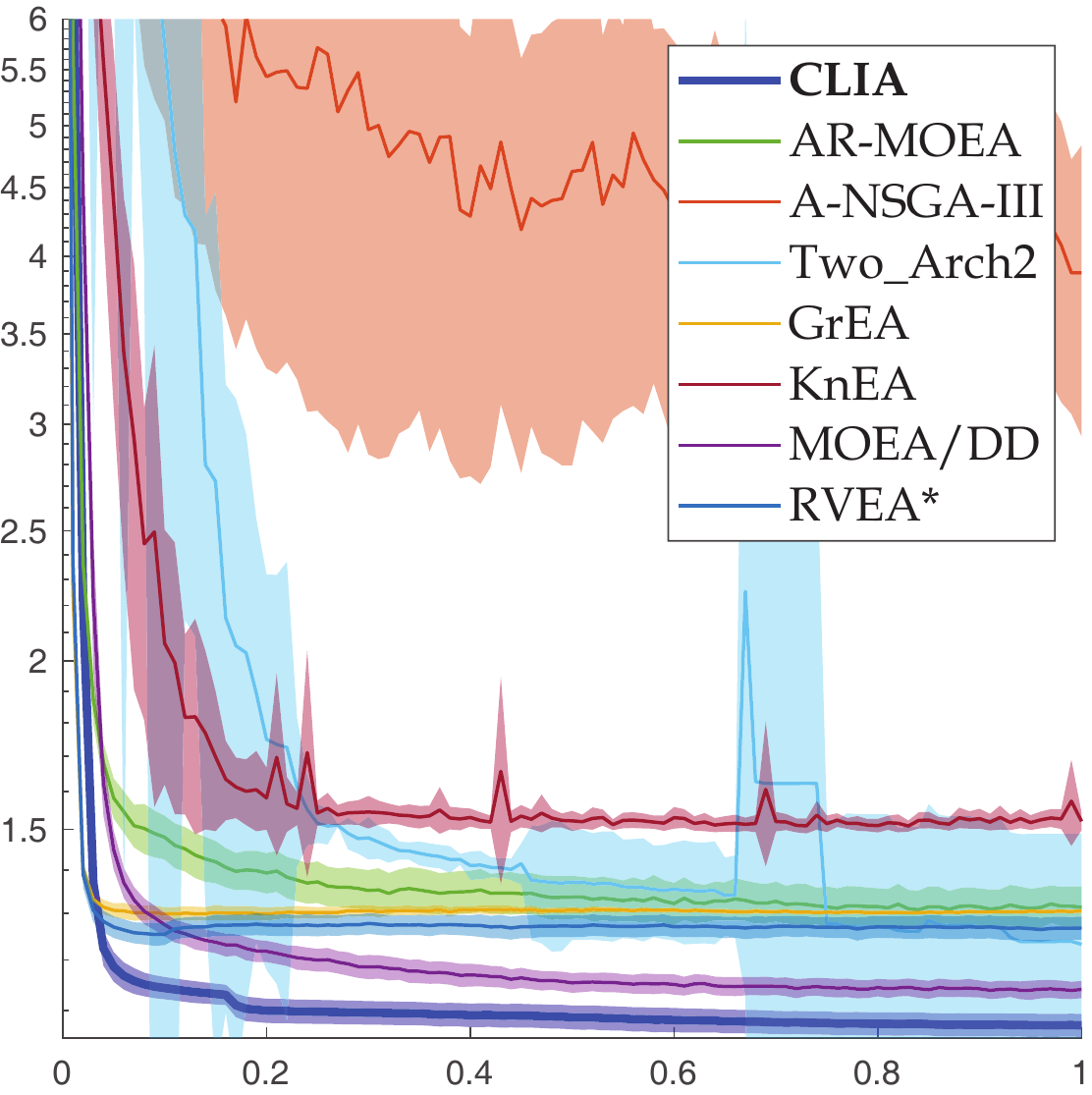}}

\caption{Real-time IGD data as mean curves, with confidence intervals. The abscissae are the percentage of consumed FEs. The minimum and maximum bands make the diagrams too messy and are therefore reduced.}
\label{fig:IGD_curves}
\end{figure*}

To examine the convergence pattern of \algoabbr{}, the collected IGD curves of the algorithms on some test cases are presented in Fig. \ref{fig:IGD_curves}. The abscissae show the percentage of FEs consumed.
\par
In Fig. \ref{fig:IGD_curves} (a), a fast and steady convergence of \algoabbr{} can be observed, where other MOEAs either fluctuate violently or converge slowly. Since the problem MaF3 has a full PF, the phenomena may indicate that \MakeLowercase{\clusteringprocessfullname{}} is able to provide sufficient evolution pressure toward proximity and diversity in the high-dimensional objective space; In Fig. \ref{fig:IGD_curves} (b), on MaF6 ($M=15$) with a degenerate PF, a step-by-step convergence trend of \algoabbr{} can be observed. It indicates that the incremental learning in \algoabbr{} managed to generate more useful reference vectors and provide better guidance for diversity. Generally, degenerate PFs are challenging for the reference adaptation mechanisms, since they are tiny compared to the feasible objective space. On this problem, it can be seen that \algoabbr{} has outperformed the other reference vector-based MOEAs and even all other compared ones; In Fig. \ref{fig:IGD_curves} (c), the curves depict the convergence patterns of the MOEAs on a problem with extremely-high dimension of the solution space, with huge amount of FEs provided. We can observe a turning point for the convergence curve of \algoabbr{}, where the status reached stable and the incremental learning provided better reference vectors to guide the evolution. These show that most of the compared algorithms have much slower convergence speeds with respect to the FEs consumed, while \algoabbr{} still managed to provide satisfactory evolution pressure.
\par
The IGD curves show that \algoabbr{} is relatively stable, as some algorithms have unexpected spring-backs. Also, its convergence is relatively fast. Combining with the competitive performance on such a complicated benchmark, its comprehensive capabilities are thoroughly demonstrated.
\subsection{Component Analyses}
\subsubsection{Characteristics of PDM}
To examine the differences in behavior of PDM and PBI on the problems with diverse PF curvatures, we have implemented two non-adaptive version of \algoabbr{} as PBICC and PDMCC, of which the only difference is the metrics of frontier individual evaluation, \ie{} PBI or PDM. We have chosen the full PF problems with diverse curvatures, including DTLZ1, DTLZ2, a convex version of DTLZ3 \cite{deb2005scalable} as well as WFG1 \cite{huband2006review}. The test is conducted on the test cases on these $4$ problems with $M=\{2,3,4,5,6,7,8,9,10\}$ for $20$ independent runs. The results are shown in Table \ref{tab:metric}.
\begin{table}[!t]
\tiny
\setlength{\tabcolsep}{0.5pt}
\renewcommand\arraystretch{0.8}
  \centering
  \caption{PDM vs. PBI on Different PF Curvatures}

    \begin{tabular}{cccccccccccccccc}
    \toprule
    \toprule
    \multirow{2}[2]{*}{Problem} & \multirow{2}[2]{*}{$M$} & \multirow{2}[2]{*}{$D$} & \multirow{2}[2]{*}{FEs} & \multicolumn{2}{c}{PDMCC} & \multicolumn{2}{c}{PBICC} & \multirow{2}[2]{*}{Problem} & \multirow{2}[2]{*}{$M$} & \multirow{2}[2]{*}{$D$} & \multirow{2}[2]{*}{FEs} & \multicolumn{2}{c}{PDMCC} & \multicolumn{2}{c}{PBICC} \\
          &       &       &       & Mean  & Std   & Mean  & Std   &       &       &       &       & Mean  & Std   & Mean  & Std \\
    \midrule
    \multirow{9}[1]{*}{DTLZ1} & 2     & 6     & 1.0e5 & \cellcolor[rgb]{ .859,  .859,  .859}7.50e-4 & 1.20e-5 & 7.50e-4 & 1.24e-5 & \multirow{9}[1]{*}{cDTLZ3} & 2     & 11    & 1.1e5 & \cellcolor[rgb]{ .859,  .859,  .859}3.66e-3 & 7.24e-4 & 4.16e-3 & 9.58e-4 \\
          & 3     & 7     & 1.0e5 & \cellcolor[rgb]{ .859,  .859,  .859}1.24e-2 & 2.22e-5 & 1.24e-2 & 2.13e-5 &       & 3     & 12    & 1.2e5 & \cellcolor[rgb]{ .859,  .859,  .859}2.61e-2 & 4.98e-4 & 3.62e-2 & 7.52e-4 \\
          & 4     & 8     & 1.0e5 & \cellcolor[rgb]{ .859,  .859,  .859}3.21e-2 & 4.64e-5 & 3.21e-2 & 4.26e-5 &       & 4     & 13    & 1.3e5 & \cellcolor[rgb]{ .859,  .859,  .859}4.82e-2 & 9.59e-4 & 7.50e-2 & 1.54e-3 \\
          & 5     & 9     & 1.0e5 & \cellcolor[rgb]{ .859,  .859,  .859}5.27e-2 & 3.97e-5 & 5.27e-2 & 4.97e-5 &       & 5     & 14    & 1.4e5 & \cellcolor[rgb]{ .859,  .859,  .859}6.42e-2 & 1.74e-3 & 1.06e-1 & 2.29e-3 \\
          & 6     & 10    & 1.0e5 & \cellcolor[rgb]{ .859,  .859,  .859}6.96e-2 & 1.66e-4 & 6.97e-2 & 1.93e-4 &       & 6     & 15    & 1.5e5 & \cellcolor[rgb]{ .859,  .859,  .859}7.81e-2 & 2.07e-3 & 1.21e-1 & 1.41e-3 \\
          & 7     & 11    & 1.1e5 & 8.20e-2 & 4.70e-4 & \cellcolor[rgb]{ .859,  .859,  .859}8.15e-2 & 3.34e-4 &       & 7     & 16    & 1.6e5 & \cellcolor[rgb]{ .859,  .859,  .859}7.47e-2 & 1.88e-3 & 1.26e-1 & 1.13e-3 \\
          & 8     & 12    & 1.2e5 & \cellcolor[rgb]{ .859,  .859,  .859}8.73e-2 & 1.54e-4 & 8.80e-2 & 3.11e-4 &       & 8     & 17    & 1.7e5 & \cellcolor[rgb]{ .859,  .859,  .859}8.53e-2 & 1.39e-3 & 1.30e-1 & 2.08e-3 \\
          & 9     & 13    & 1.3e5 & 1.04e-1 & 1.42e-3 & 9.96e-2 & 1.13e-3 &       & 9     & 18    & 1.8e5 & \cellcolor[rgb]{ .859,  .859,  .859}7.27e-2 & 3.31e-3 & 1.06e-1 & 9.14e-4 \\
          & 10    & 14    & 1.4e5 & \cellcolor[rgb]{ .859,  .859,  .859}1.30e-1 & 3.22e-3 & 1.33e-1 & 1.83e-3 &       & 10    & 19    & 1.9e5 & \cellcolor[rgb]{ .859,  .859,  .859}7.92e-2 & 1.53e-3 & 1.19e-1 & 1.24e-3 \\
    \multicolumn{4}{c}{$t$-test}    & \multicolumn{4}{c}{\textbf{3/4/2}} & \multicolumn{4}{c}{$t$-test}    & \multicolumn{4}{c}{\textbf{8/1/0}} \\
    \midrule
    \multirow{9}[1]{*}{DTLZ2} & 2     & 11    & 1.1e5 & 1.64e-3 & 1.17e-6 &\cellcolor[rgb]{ .859,  .859,  .859} 1.64e-3 & 5.96e-7 & \multirow{9}[1]{*}{WFG1} & 2     & 11    & 1.1e5 & \cellcolor[rgb]{ .859,  .859,  .859}1.32e-1 & 2.23e-2 & 1.33e-1 & 2.18e-2 \\
          & 3     & 12    & 1.2e5 & 3.28e-2 & 8.07e-6 & \cellcolor[rgb]{ .859,  .859,  .859}3.28e-2 & 4.85e-6 &       & 3     & 12    & 1.2e5 & \cellcolor[rgb]{ .859,  .859,  .859}1.79e-1 & 8.68e-3 & 1.92e-1 & 4.50e-2 \\
          & 4     & 13    & 1.3e5 & \cellcolor[rgb]{ .859,  .859,  .859}9.53e-2 & 1.03e-5 & 9.53e-2 & 1.13e-5 &       & 4     & 13    & 1.3e5 & \cellcolor[rgb]{ .859,  .859,  .859}3.23e-1 & 6.33e-3 & 3.49e-1 & 4.16e-2 \\
          & 5     & 14    & 1.4e5 & \cellcolor[rgb]{ .859,  .859,  .859}1.65e-1 & 2.09e-4 & 1.65e-1 & 1.63e-5 &       & 5     & 14    & 1.4e5 & \cellcolor[rgb]{ .859,  .859,  .859}5.23e-1 & 9.13e-3 & 6.08e-1 & 5.84e-2 \\
          & 6     & 15    & 1.5e5 & 2.27e-1 & 1.54e-4 & 2.27e-1 & 1.66e-4 &       & 6     & 15    & 1.5e5 & \cellcolor[rgb]{ .859,  .859,  .859}8.73e-1 & 1.18e-2 & 9.96e-1 & 8.12e-2 \\
          & 7     & 16    & 1.6e5 & \cellcolor[rgb]{ .859,  .859,  .859}2.65e-1 & 5.68e-4 & 2.65e-1 & 4.98e-4 &       & 7     & 16    & 1.6e5 & \cellcolor[rgb]{ .859,  .859,  .859}1.11e0 & 3.78e-2 & 1.17e0 & 7.09e-2 \\
          & 8     & 17    & 1.7e5 & \cellcolor[rgb]{ .859,  .859,  .859}3.03e-1 & 2.29e-4 & 3.03e-1 & 2.14e-4 &       & 8     & 17    & 1.7e5 & \cellcolor[rgb]{ .859,  .859,  .859}1.16e0 & 3.99e-2 & 1.31e0 & 1.37e-1 \\
          & 9     & 18    & 1.8e5 & 3.80e-1 & 2.14e-3 & 3.78e-1 & 2.05e-3 &       & 9     & 18    & 1.8e5 & \cellcolor[rgb]{ .859,  .859,  .859}1.29e0 & 3.78e-2 & 1.58e0 & 4.27e-2 \\
          & 10    & 19    & 1.9e5 & \cellcolor[rgb]{ .859,  .859,  .859}4.56e-1 & 5.25e-4 & 4.56e-1 & 4.63e-4 &       & 10    & 19    & 1.9e5 & \cellcolor[rgb]{ .859,  .859,  .859}1.60e0 & 3.91e-2 & 1.93e0 & 3.91e-2 \\
    \multicolumn{4}{c}{$t$-test}    & \multicolumn{4}{c}{0/8/1}     & \multicolumn{4}{c}{$t$-test}    & \multicolumn{4}{c}{\textbf{8/1/0}} \\
    \midrule
    \multicolumn{4}{c}{overall $t$-test} & \multicolumn{12}{c}{\textbf{19/14/2}} \\
    \bottomrule
    \bottomrule
    \end{tabular}%

  \label{tab:metric}%
\end{table}%

\begin{table}[!t]
  \centering
  \tiny
\setlength{\tabcolsep}{0.5pt}
\renewcommand\arraystretch{0.8}
  \caption{Comparison for Reference Vector based Selection Operators}
    \begin{tabular}{ccccccccccccccccc}
    \toprule
    \toprule
    \multirow{2}[2]{*}{Problem} & \multirow{2}[2]{*}{$M$} & \multicolumn{3}{c}{PDMCC} & \multicolumn{4}{c}{NSGA-III}   & \multicolumn{4}{c}{RVEA}      & \multicolumn{4}{c}{MOEA/DD} \\
          &       & Mean  & Std   & tpI  & Mean  & Std   & tpI  & ratio & Mean  & Std   & tpI  & ratio & Mean  & Std   & tpI  & ratio \\
    \midrule
    \multirow{9}[2]{*}{DTLZ1} & 2     & 7.5e-4 & 1.2e-5 & 2.e-2 & 7.58e-4 & 6.02e-5 & 4.e-2 & 204\% & 7.55e-4 & 4.14e-5 & 3.4e-2 & 173\% & \cellcolor[rgb]{ .859,  .859,  .859}7.41e-4 & 2.9e-6 & 1.4e0 & 7216\% \\
          & 3     & 1.24e-2 & 2.22e-5 & 2.3e-2 & 1.24e-2 & 5.14e-5 & 4.6e-2 & 202\% & 1.24e-2 & 2.55e-5 & 3.4e-2 & 150\% & \cellcolor[rgb]{ .859,  .859,  .859}1.23e-2 & 5.12e-6 & 1.4e0 & 6285\% \\
          & 4     & 3.21e-2 & 4.64e-5 & 2.4e-2 & 3.21e-2 & 1.12e-4 & 4.8e-2 & 197\% & \cellcolor[rgb]{ .859,  .859,  .859}3.21e-2 & 3.65e-5 & 3.4e-2 & 143\% & 3.21e-2 & 3.63e-5 & 1.5e0 & 6308\% \\
          & 5     & \cellcolor[rgb]{ .859,  .859,  .859}5.27e-2 & 3.97e-5 & 2.5e-2 & 5.28e-2 & 1.75e-4 & 5.2e-2 & 203\% & 5.27e-2 & 6.33e-5 & 3.6e-2 & 143\% & 5.27e-2 & 4.06e-5 & 1.6e0 & 6106\% \\
          & 6     & 6.96e-2 & 1.66e-4 & 2.7e-2 & 6.92e-2 & 1.49e-4 & 5.2e-2 & 191\% & 6.91e-2 & 1.01e-4 & 3.5e-2 & 129\% & \cellcolor[rgb]{ .859,  .859,  .859}6.91e-2 & 6.6e-5 & 1.5e0 & 5414\% \\
          & 7     & 8.2e-2 & 4.7e-4 & 2.7e-2 & 8.16e-2 & 7.15e-3 & 5.5e-2 & 204\% & 8.0e-2 & 2.39e-3 & 3.4e-2 & 127\% & \cellcolor[rgb]{ .859,  .859,  .859}7.91e-2 & 1.26e-4 & 1.7e0 & 6194\% \\
          & 8     & 8.73e-2 & 1.54e-4 & 2.7e-2 & 9.78e-2 & 1.29e-2 & 6.e-2 & 220\% & 8.71e-2 & 9.04e-4 & 3.5e-2 & 127\% & \cellcolor[rgb]{ .859,  .859,  .859}8.65e-2 & 1.62e-4 & 1.7e0 & 6240\% \\
          & 9     & 1.04e-1 & 1.42e-3 & 2.9e-2 & 1.08e-1 & 1.25e-2 & 6.2e-2 & 218\% & 9.78e-2 & 6.48e-4 & 3.5e-2 & 121\% & \cellcolor[rgb]{ .859,  .859,  .859}9.74e-2 & 1.89e-4 & 1.6e0 & 5675\% \\
          & 10    & \cellcolor[rgb]{ .859,  .859,  .859}1.3e-1 & 3.22e-3 & 2.9e-2 & 1.39e-1 & 2.5e-2 & 6.7e-2 & 229\% & 1.34e-1 & 1.41e-3 & 3.5e-2 & 120\% & 1.33e-1 & 1.0e-3 & 1.7e0 & 5853\% \\
    \midrule
    \multirow{9}[2]{*}{DTLZ2} & 2     & \cellcolor[rgb]{ .859,  .859,  .859}1.64e-3 & 1.17e-6 & 2.2e-2 & 1.64e-3 & 3.87e-7 & 4.6e-2 & 211\% & 2.03e-3 & 2.21e-4 & 3.8e-2 & 174\% & 1.65e-3 & 1.32e-5 & 1.5e0 & 6998\% \\
          & 3     & \cellcolor[rgb]{ .859,  .859,  .859}3.28e-2 & 8.07e-6 & 2.6e-2 & 3.28e-2 & 2.39e-5 & 5.3e-2 & 205\% & 3.28e-2 & 6.44e-5 & 3.8e-2 & 149\% & 3.28e-2 & 1.36e-4 & 1.6e0 & 6265\% \\
          & 4     & 9.53e-2 & 1.03e-5 & 2.7e-2 & \cellcolor[rgb]{ .859,  .859,  .859}9.53e-2 & 1.21e-5 & 5.6e-2 & 209\% & 9.53e-2 & 4.62e-6 & 3.8e-2 & 143\% & 9.53e-2 & 1.7e-6 & 1.7e0 & 6517\% \\
          & 5     & \cellcolor[rgb]{ .859,  .859,  .859}1.65e-1 & 2.09e-4 & 2.7e-2 & 1.65e-1 & 1.66e-5 & 5.5e-2 & 202\% & 1.65e-1 & 6.14e-6 & 3.8e-2 & 140\% & 1.65e-1 & 1.62e-6 & 1.7e0 & 6273\% \\
          & 6     & 2.27e-1 & 1.54e-4 & 2.9e-2 & 2.27e-1 & 3.5e-5 & 5.9e-2 & 207\% & 2.27e-1 & 1.74e-5 & 4.e-2 & 138\% & \cellcolor[rgb]{ .859,  .859,  .859}2.27e-1 & 6.84e-6 & 1.6e0 & 5578\% \\
          & 7     & 2.65e-1 & 5.68e-4 & 3.e-2 & 2.63e-1 & 7.26e-5 & 6.2e-2 & 208\% & 2.63e-1 & 4.44e-5 & 3.9e-2 & 130\% & \cellcolor[rgb]{ .859,  .859,  .859}2.63e-1 & 1.37e-5 & 1.9e0 & 6260\% \\
          & 8     & \cellcolor[rgb]{ .859,  .859,  .859}3.03e-1 & 2.29e-4 & 3.1e-2 & 3.42e-1 & 8.78e-2 & 6.6e-2 & 216\% & 3.04e-1 & 2.29e-4 & 3.8e-2 & 122\% & 3.04e-1 & 2.28e-5 & 1.9e0 & 6180\% \\
          & 9     & 3.8e-1 & 2.14e-3 & 3.1e-2 & 4.1e-1 & 8.24e-2 & 6.5e-2 & 209\% & \cellcolor[rgb]{ .859,  .859,  .859}3.7e-1 & 2.23e-4 & 3.8e-2 & 123\% & 3.7e-1 & 1.34e-4 & 1.7e0 & 5618\% \\
          & 10    & 4.56e-1 & 5.25e-4 & 3.2e-2 & 4.96e-1 & 6.76e-2 & 6.9e-2 & 216\% & \cellcolor[rgb]{ .859,  .859,  .859}4.53e-1 & 2.63e-4 & 3.9e-2 & 122\% & 4.54e-1 & 3.84e-4 & 1.8e0 & 5753\% \\
    \midrule
    \multirow{9}[2]{*}{cDTLZ3} & 2     & \cellcolor[rgb]{ .859,  .859,  .859}3.66e-3 & 7.24e-4 & 2.e-2 & 3.67e-3 & 1.2e-3 & 4.4e-2 & 215\% & 5.74e-2 & 9.17e-2 & 3.6e-2 & 177\% & 6.53e-3 & 3.47e-3 & 1.4e0 & 6888\% \\
          & 3     & \cellcolor[rgb]{ .859,  .859,  .859}2.61e-2 & 4.98e-4 & 2.2e-2 & 3.08e-2 & 7.01e-4 & 4.7e-2 & 210\% & 3.87e-2 & 1.09e-2 & 3.7e-2 & 164\% & 3.51e-2 & 1.1e-3 & 1.4e0 & 6277\% \\
          & 4     & \cellcolor[rgb]{ .859,  .859,  .859}4.82e-2 & 9.59e-4 & 2.4e-2 & 5.45e-2 & 1.97e-3 & 4.9e-2 & 204\% & 5.39e-2 & 1.57e-2 & 3.7e-2 & 153\% & 7.31e-2 & 2.54e-3 & 1.5e0 & 6068\% \\
          & 5     & \cellcolor[rgb]{ .859,  .859,  .859}6.42e-2 & 1.74e-3 & 2.6e-2 & 7.21e-2 & 3.36e-3 & 5.3e-2 & 205\% & 6.52e-2 & 5.8e-3 & 3.7e-2 & 144\% & 1.04e-1 & 3.79e-3 & 1.5e0 & 5751\% \\
          & 6     & \cellcolor[rgb]{ .859,  .859,  .859}7.81e-2 & 2.07e-3 & 2.8e-2 & 1.59e-1 & 2.73e-1 & 5.8e-2 & 210\% & 8.86e-2 & 1.72e-2 & 3.8e-2 & 137\% & 1.2e-1 & 2.16e-3 & 1.4e0 & 5131\% \\
          & 7     & \cellcolor[rgb]{ .859,  .859,  .859}7.47e-2 & 1.88e-3 & 2.8e-2 & 5.51e0 & 8.57e0 & 6.1e-2 & 220\% & 8.65e-2 & 1.57e-2 & 3.7e-2 & 134\% & 1.22e-1 & 2.51e-3 & 1.6e0 & 5768\% \\
          & 8     & \cellcolor[rgb]{ .859,  .859,  .859}8.53e-2 & 1.39e-3 & 2.9e-2 & 2.2e3 & 1.09e4 & 6.6e-2 & 229\% & 1.03e-1 & 1.37e-2 & 3.9e-2 & 134\% & 1.16e-1 & 4.78e-3 & 1.6e0 & 5684\% \\
          & 9     & \cellcolor[rgb]{ .859,  .859,  .859}7.27e-2 & 3.31e-3 & 4.4e-2 & 2.0e5 & 6.56e5 & 7.e-2 & 158\% & 8.32e-2 & 1.42e-2 & 3.6e-2 & 82\%  & 1.02e-1 & 2.07e-3 & 1.6e0 & 3539\% \\
          & 10    & \cellcolor[rgb]{ .859,  .859,  .859}7.92e-2 & 1.53e-3 & 4.2e-2 & 1.38e5 & 3.52e5 & 7.5e-2 & 177\% & 9.29e-2 & 1.35e-2 & 3.8e-2 & 91\%  & 1.16e-1 & 2.6e-3 & 1.7e0 & 3932\% \\
    \midrule
    \multirow{9}[2]{*}{WFG1} & 2     & 1.31e-1 & 2.23e-2 & 2.5e-2 & 1.38e-1 & 1.25e-2 & 4.5e-2 & 185\% & 3.02e-1 & 5.28e-2 & 3.8e-2 & 157\% & \cellcolor[rgb]{ .859,  .859,  .859}7.81e-2 & 5.04e-2 & 1.7e0 & 7009\% \\
          & 3     & 1.66e-1 & 8.68e-3 & 2.8e-2 & \cellcolor[rgb]{ .859,  .859,  .859}1.32e-1 & 2.99e-2 & 5.3e-2 & 187\% & 2.52e-1 & 3.38e-2 & 3.5e-2 & 125\% & 1.5e-1 & 1.45e-2 & 1.8e0 & 6382\% \\
          & 4     & 3.36e-1 & 6.33e-3 & 2.9e-2 & \cellcolor[rgb]{ .859,  .859,  .859}2.83e-1 & 3.19e-2 & 5.7e-2 & 194\% & 2.9e-1 & 1.82e-2 & 3.6e-2 & 123\% & 4.12e-1 & 6.22e-2 & 1.9e0 & 6411\% \\
          & 5     & 5.19e-1 & 9.13e-3 & 3.e-2 & 4.19e-1 & 3.37e-2 & 5.5e-2 & 182\% & \cellcolor[rgb]{ .859,  .859,  .859}3.94e-1 & 1.61e-2 & 3.6e-2 & 118\% & 5.69e-1 & 5.92e-2 & 1.9e0 & 6124\% \\
          & 6     & 8.25e-1 & 1.18e-2 & 3.2e-2 & \cellcolor[rgb]{ .859,  .859,  .859}5.62e-1 & 2.11e-2 & 5.7e-2 & 177\% & 5.82e-1 & 4.8e-2 & 3.7e-2 & 113\% & 1.08e0 & 1.73e-1 & 1.8e0 & 5602\% \\
          & 7     & 9.58e-1 & 3.78e-2 & 3.2e-2 & \cellcolor[rgb]{ .859,  .859,  .859}6.77e-1 & 4.25e-2 & 6.e-2 & 188\% & 7.3e-1 & 8.32e-2 & 3.6e-2 & 111\% & 1.03e0 & 1.12e-1 & 2.e0 & 6300\% \\
          & 8     & 1.16e0 & 3.99e-2 & 3.3e-2 & \cellcolor[rgb]{ .859,  .859,  .859}8.49e-1 & 4.13e-2 & 6.6e-2 & 196\% & 1.04e0 & 1.12e-1 & 3.6e-2 & 109\% & 1.65e0 & 2.62e-1 & 2.1e0 & 6171\% \\
          & 9     & 1.33e0 & 3.78e-2 & 3.4e-2 & \cellcolor[rgb]{ .859,  .859,  .859}9.1e-1 & 5.39e-2 & 6.6e-2 & 193\% & 1.05e0 & 6.67e-2 & 3.8e-2 & 111\% & 1.5e0 & 1.2e-1 & 1.9e0 & 5555\% \\
          & 10    & 1.6e0 & 3.91e-2 & 3.5e-2 & \cellcolor[rgb]{ .859,  .859,  .859}1.12e0 & 6.11e-2 & 6.7e-2 & 192\% & 1.32e0 & 5.99e-2 & 3.6e-2 & 104\% & 1.94e0 & 1.02e-1 & 2.e0 & 5760\% \\
    \midrule
    \multicolumn{4}{c}{t-test / tpI (ms) / ratio} & 2.9e-2 & \multicolumn{2}{c}{\textbf{14/12/10}} & 5.7e-2 & 202\% & \multicolumn{2}{c}{\textbf{14/10/12}} & 3.7e-2 & 132\% & \multicolumn{2}{c}{\textbf{19/6/11}} & 1.7e0 & 5975\% \\
    \bottomrule
    \bottomrule
    \end{tabular}%
  \label{tab:engine}%
\end{table}%

\par
As we can see from the $t$-test results, PDMCC outperforms PBICC on DTLZ1 with a linear PF, convex cDTLZ3 with a convex PF and WFG1 with a PF of hybrid curvatures. On DTLZ2, due to the preference for concave PFs, PBICC achieved better performance. These results concur with our analyses about PBI and PDM in Section III, that PBI's natural preference for concave PFs may cause the deterioration in versatility and the compromised design of PDM will perform more versatilely on the PF curvatures.
\subsubsection{Effectiveness and Efficiency of \clusteringprocessfullname{}}
In this part, we examine \MakeLowercase{\clusteringprocessfullname{}} by comparing its performance with the counterparts in RVEA \cite{cheng2016reference}, NSGA-III \cite{deb2013evolutionary} and MOEA/DD \cite{li2015evolutionary}. With mutual predefined reference vectors, these tests are conducted with the same settings for PDM analysis. The results are presented in Table \ref{tab:engine}.
\par
The $t$-tests show that PDMCC has slightly better performance than the compared selection operators and thus the effectiveness of \MakeLowercase{\clusteringprocessfullname{}} is demonstrated. 
\begin{figure*}[!t]
\centering
\subfloat[MaF2, $M=5$]{
\captionsetup{justification = centering}
\includegraphics[width=0.2\textwidth]{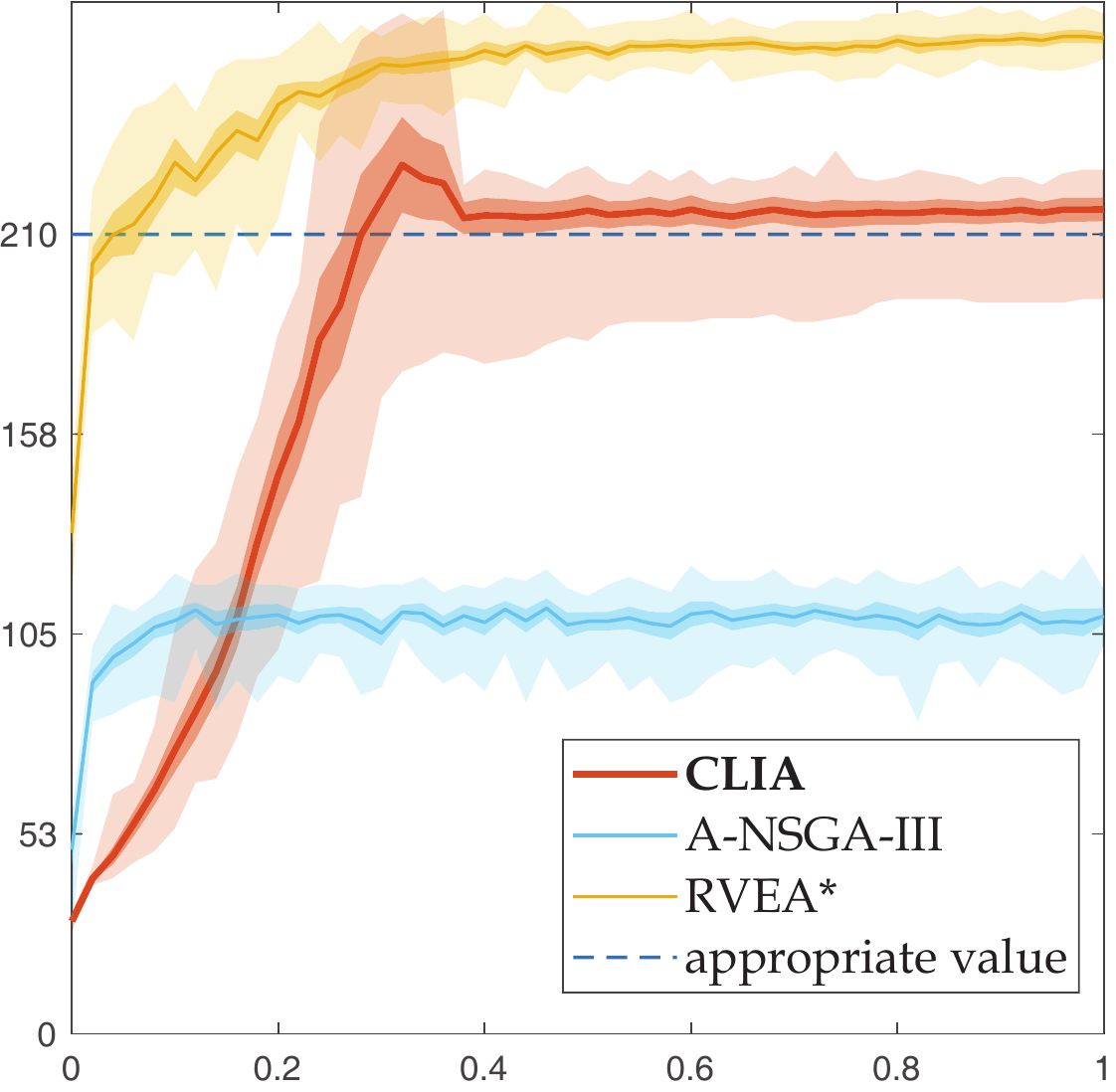}}
\hfill
\subfloat[MaF6, $M=5$]{
\captionsetup{justification = centering}
\includegraphics[width=0.2\textwidth]{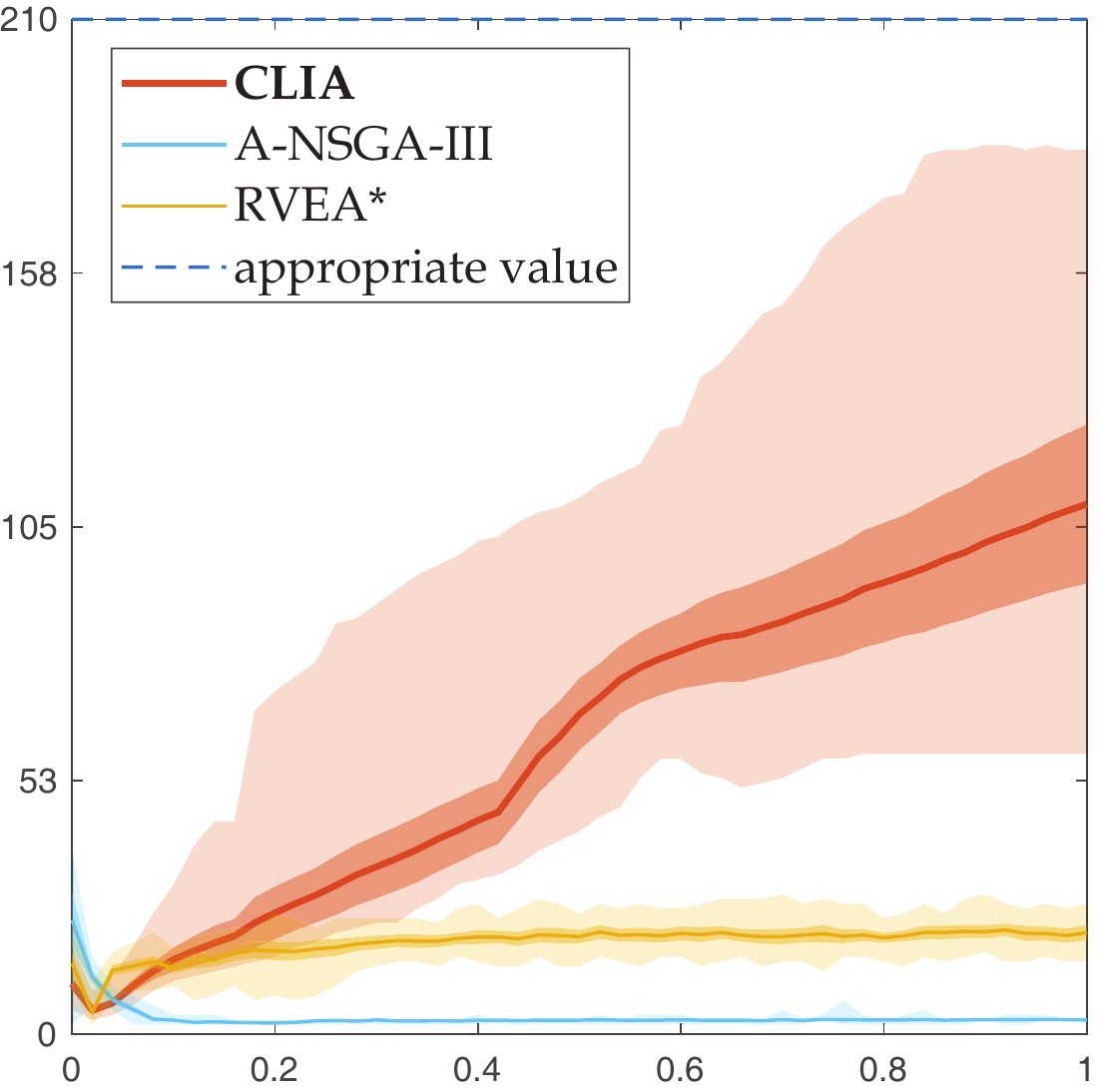}}
\hfill
\subfloat[MaF7, $M=5$]{
\captionsetup{justification = centering}
\includegraphics[width=0.2\textwidth]{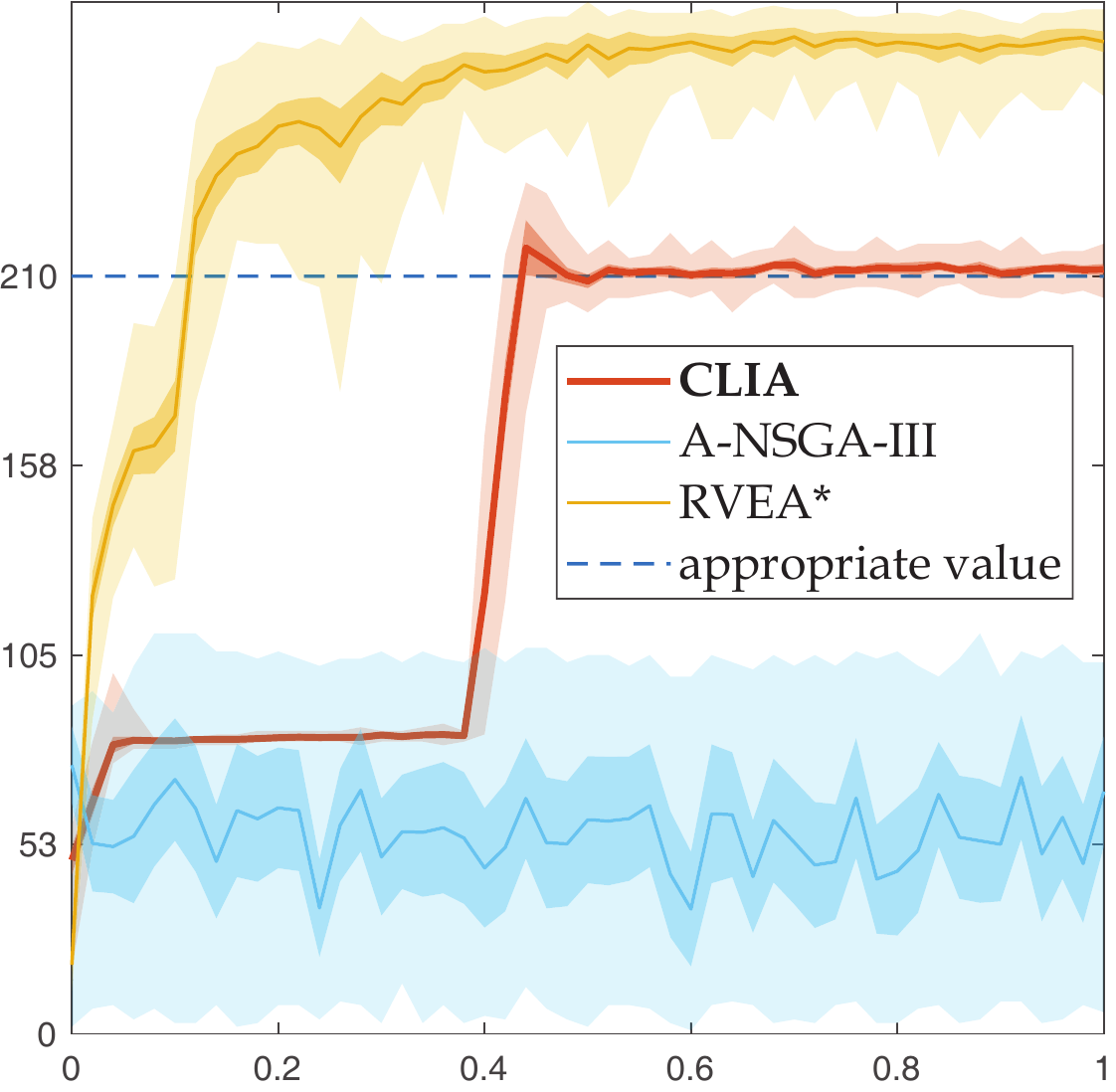}}
\hfill
\subfloat[MaF13, $M=15$]{
\captionsetup{justification = centering}
\includegraphics[width=0.2\textwidth]{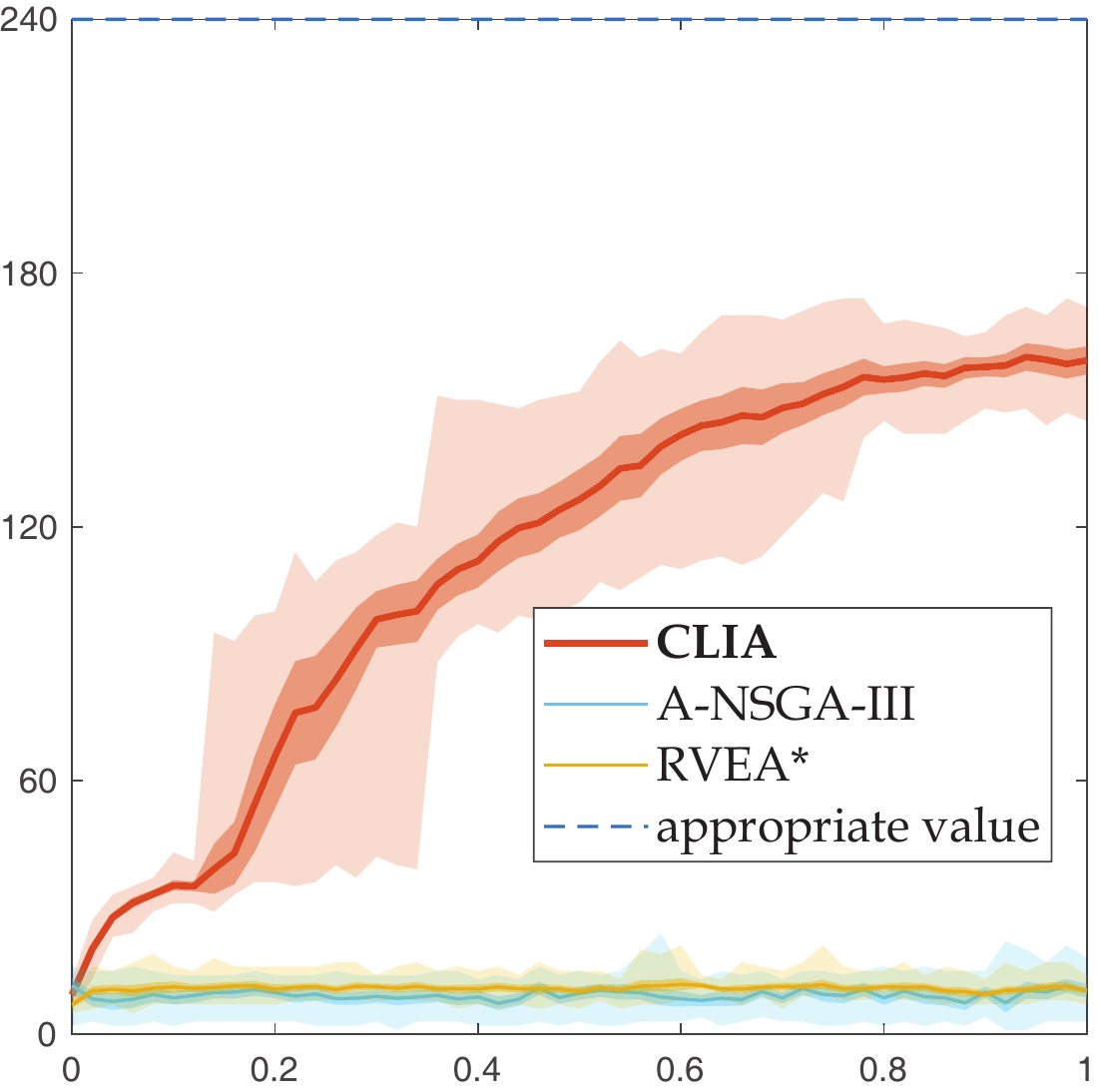}}

\caption{The real-time changes of the activity of reference vectors for problems with partial PFs. The light semi-transparent areas show the minimum and the maximum, the dark areas show the confidence interval with $\alpha = 0.05$, and the thick curves show the mean. The abscissae are the consumed FEs percentage. Appropriate values equal to corresponding population sizes. These data are gathered in the $20$ runs for the MaF suite.}
\label{fig:reference_curves}
\end{figure*}
\begin{figure*}[!t]
\centering
\subfloat[MaF1, $M=5$]{
\captionsetup{justification = centering}
\includegraphics[width=0.2\textwidth]{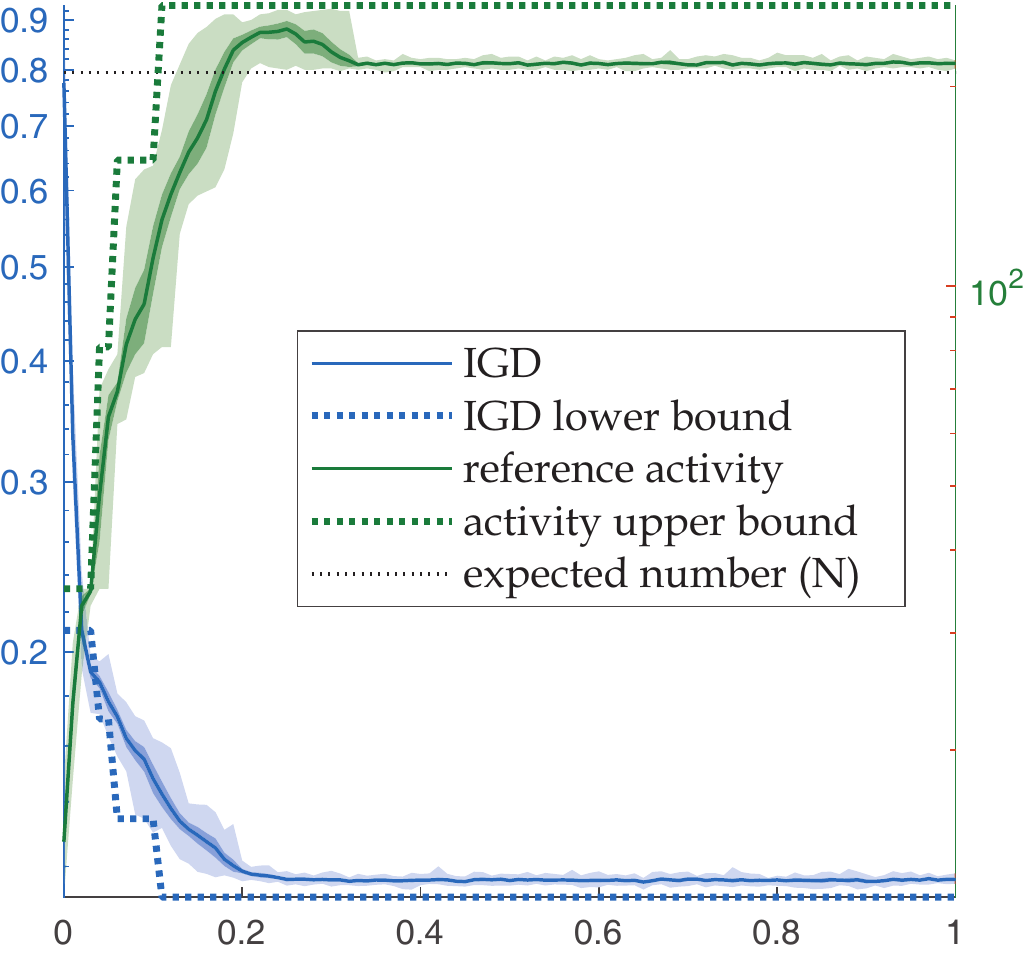}}
\hfill
\subfloat[MaF7, $M=5$]{
\captionsetup{justification = centering}
\includegraphics[width=0.2\textwidth]{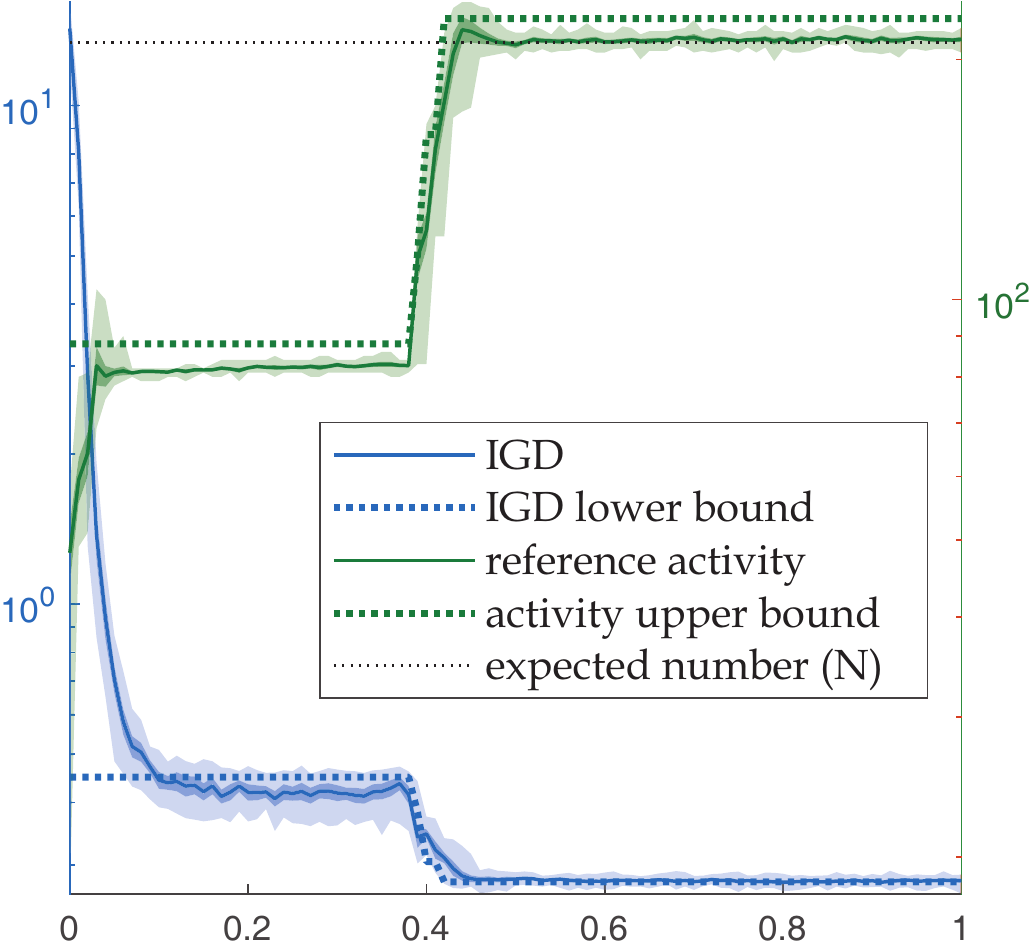}}
\hfill
\subfloat[MaF8, $M=5$]{
\captionsetup{justification = centering}
\includegraphics[width=0.2\textwidth]{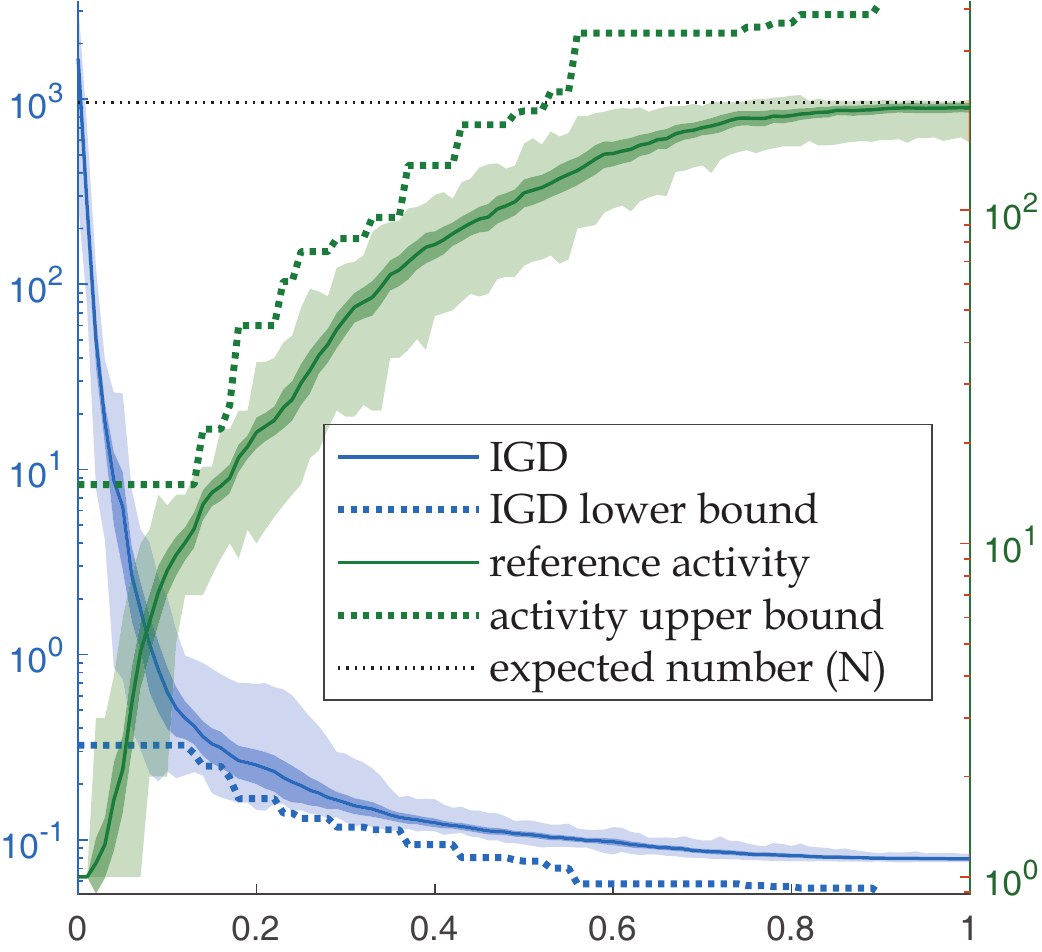}}
\hfill
\subfloat[MaF13, $M=5$]{
\captionsetup{justification = centering}
\includegraphics[width=0.2\textwidth]{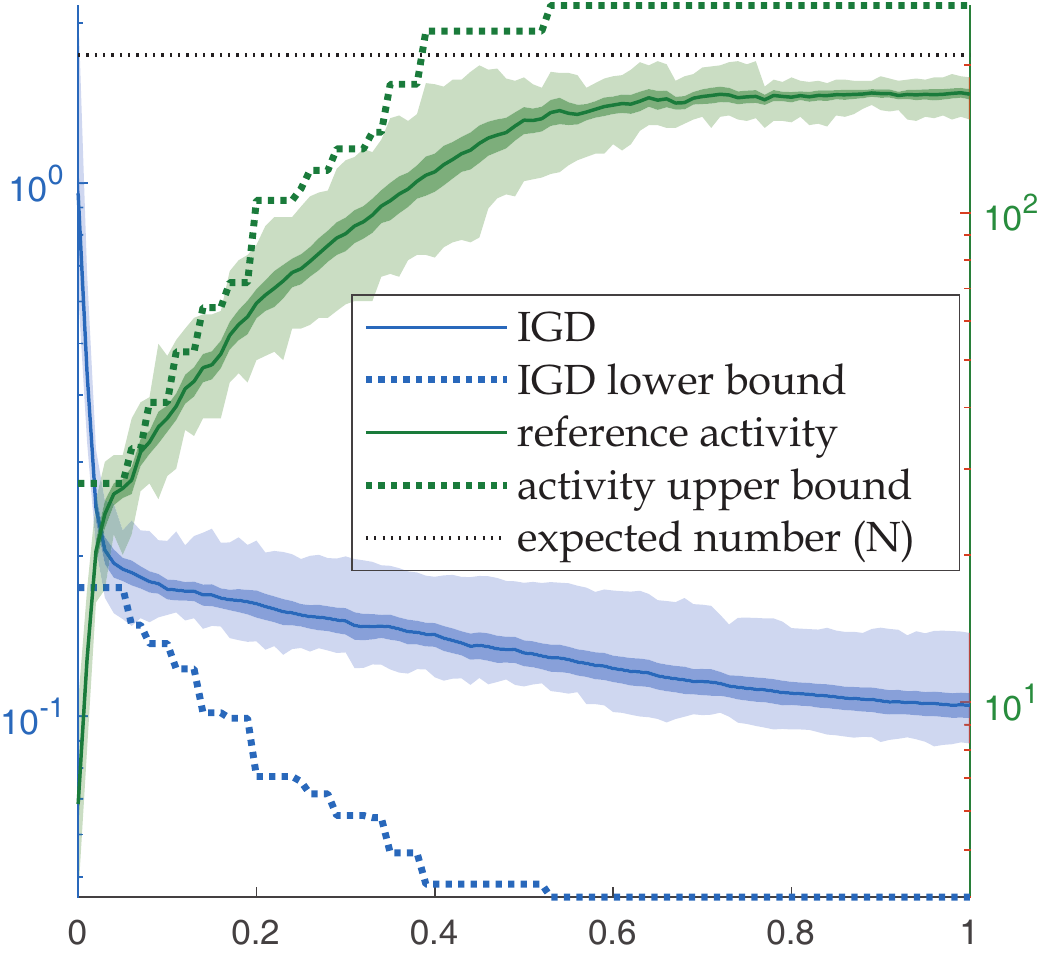}}

\caption{Comprehensive analysis diagrams for the clustering-learning interactions. The IGD and reference activity are presented with thick curves as the real-time mean value, dark bands as confidence interval with $\alpha= 0.05$ and light band as the variance intervals. The curves of the upper bound for activity and the lower bound for IGD are plotted with the mean values. The upper bound and the lower bound change synchronously each time incremental learning is initiated, providing better distributed reference vectors and thus resulting in the lifting of the reference activity and the diving of IGD.}
\label{fig:interaction_curves}
\end{figure*}
\par
NSGA-III has achieved superior performance on WFG1, since it is built with normalization scheme that can deal with the badly-scaled objectives. To cooperate with the learning process, \MakeLowercase{\clusteringprocessfullname{}} cannot adopt normalization, in case the unequal scaling on each objective disturbs the learning for the distribution of the true PF in the objective space. However, for the full PFs, we can assemble normalization scheme with \MakeLowercase{\clusteringprocessfullname{}}. In the supplementary file, the results of CC with normalization will be given. With normalization, CC outperforms the compared counterparts.
\par
The averaged time costs for picking one individual (tpI) are presented in Table \ref{tab:engine}. The ratios of the costs of the compared operators to PDMCC are also presented. Among them, RVEA and PDMCC are able to achieve high performance with fast speed. It should also be noticed that the utilization of dominance make the advantage of the efficiency decrease with the increase of the objectives, compared with RVEA.
\subsubsection{Effectiveness for \learningprocessfullname{}}
To validate the performance of the proposed \MakeLowercase{\learningprocessfullname{}} mechanism, Fig. \ref{fig:reference_curves} shows the changes in the numbers of active reference vectors (abbreviated as ``activity'') along the runs, in comparison with the adaptation mechanisms in A-NSGA-III\cite{deb2013evolutionary} and RVEA* \cite{cheng2016reference}.
\par
With the proposed \MakeLowercase{\learningprocessfullname{}}, the activities of \algoabbr{} go higher step by step. In Fig. \ref{fig:reference_curves}(a), the activity of A-NSGA-III stagnates around $0.5N$ and the activity of RVEA* is excessively more than the appropriate value. The excesses of activity are harmful for diversity, thus the activities are expected to be close to the appropriate value. In \algoabbr{}, the adaptation will be disabled when there are sufficient active reference vectors. In (b), the activity of \algoabbr{} keeps increasing while the other two are relatively low. For \algoabbr{}, if more FEs are given, the activity will lift up to be close to the appropriate value though facing the difficulties of a degenerate PF; In Fig. \ref{fig:reference_curves}(c), \algoabbr{} reaches the appropriate activity via a single sampling-learning-reducing procedure; In Fig. \ref{fig:reference_curves}(d), where the dimension of the objective space is very high, \algoabbr{} is still able to achieve satisfactory performance. These behaviors validate the effectiveness of \MakeLowercase{\learningprocessfullname{}}.
\subsubsection{Behaviors for Clustering-Learning Interactions}
In this part, we will investigate the behaviors of the clustering-learning interactions in \algoabbr{}.
\par
In Fig. \ref{fig:interaction_curves}, the performance patterns of \algoabbr{} averaged from $20$ runs are illustrated with two sets of curves. The first set shows the number of active reference vectors (labeled ``reference activity'') and the ``activity upper bound,'' which is the maximum number of active reference vectors under the corresponding generation density without the reductions. This set of curves shows the results of incremental learning in comparison with the ideal cases, where the differences show the inaccuracies of the scoring-based reduction. The second set contains two curves. The first is the curve for actual IGD values. The second curve, labeled ``IGD lower bound,'' shows the ideal IGD values for the corresponding density. Specifically, they are the IGD values when the individuals coincide on the intersection points of the true PF and the reference vectors corresponding to the ``activity upper bound.''
\par
From the curves of the activity upper bound and the IGD lower bound, the effectiveness of the \MakeLowercase{\learningprocessfullname{}} is validated, where the data show that if the newly denser generated reference vectors are appropriately selected, the diversity of the proximate populations will be enhanced. With the comparison with the actual curves and the boundaries, we observe that generally the interactions of the two processes have effectively improved the activities of the reference vectors and the IGD performance. Though the scoring-based reductions are not totally accurate, the incremental learning-based reference vector adaptation can provide many useful and uniformly distributed reference vectors that significantly improves the performance.
\section{Conclusion}
This paper proposes an MOEA with two interacting processes: \MakeLowercase{\clusteringprocessfullname{}} and \MakeLowercase{\learningprocessfullname{}}. In \algoabbr{}, CC operates as the selection operator, providing evolution pressure for proximity and diversity using the reference vectors from the adaptation process. The reference vector adaptation mechanism based on \MakeLowercase{\learningprocessfullname{}} provides gradually adjusted reference vectors using the feedbacks from the selection operator. The experimental results show that \algoabbr{} has competitive performance on diverse problems including high efficiency and stable convergence patterns. Besides, the effectiveness of its components is thoroughly examined. It can be concluded that \algoabbr{} is effective, efficient and stable in dealing with the diverse MaOPs.
\par
In future research, we will investigate into more effective mechanisms for the selection operator. Also, we will put efforts into the combination of incremental learning and mechanisms that adapts to the curvatures of the PF so that the adaptation can be further enhanced.

\bibliographystyle{IEEEtran}
\bibliography{IEEEabrv,reference}

\begin{IEEEbiography}
[{\includegraphics[width=1in,height=1in,clip,keepaspectratio]{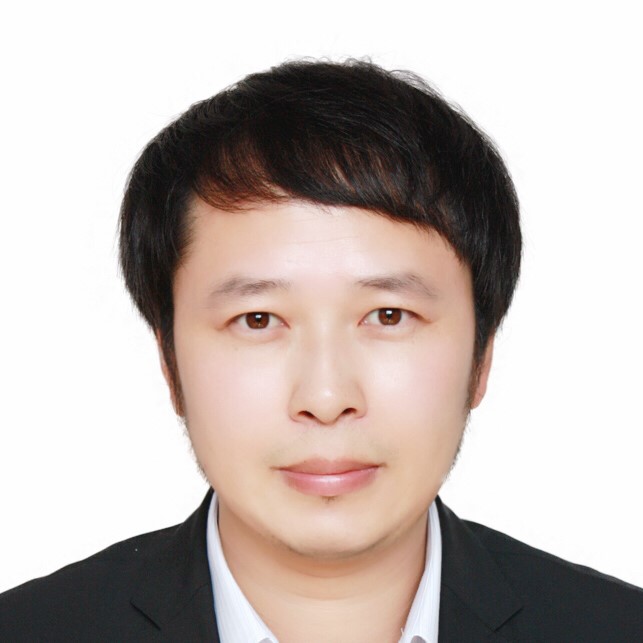}}]{Hongwei Ge}
received B.S. and M.S. degrees in mathematics from Jilin University, Jilin, China, and the Ph.D. degree in computer application technology from Jilin University, in 2006.
\par
He is currently a professor in the College of Computer Science and Technology, Dalian University of Technology, Dalian, China. His main research interests are machine learning, computational intelligence, optimization and modeling. His research was featured in the IEEE Transactions on Cybernetics, IEEE Transactions on Evolutionary Computation, Pattern Recognition, etc.
\end{IEEEbiography}
\vskip -2\baselineskip plus -1fil
\begin{IEEEbiography}
[{\includegraphics[width=1in,height=1in,clip,keepaspectratio]{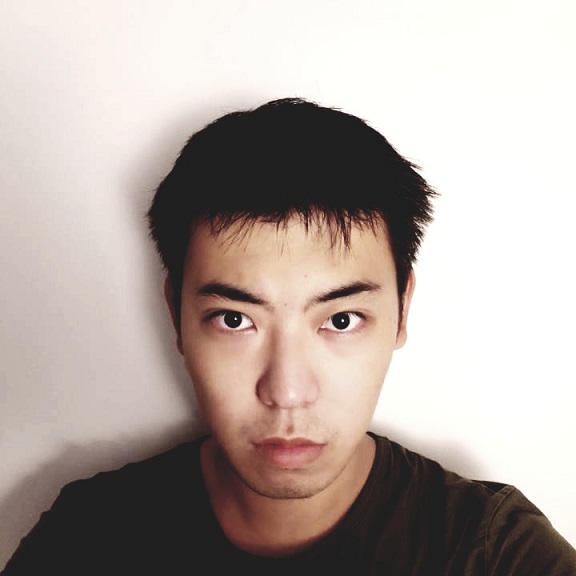}}]{Mingde Zhao}
received B.S. degree in Computer Science and Technology from Dalian University of Technology, Dalian, China, in 2018.
\par
He is currently a researcher at Mila (Montr\'eal Institute of Learning Algorithms, Qu\'ebec AI Institute) and a master student in Computer Science in School of Computer Science, McGill University, Montr\'eal, Canada. His research interests focus on meta-learning and reinforcement learning.
\end{IEEEbiography}
\vskip -2\baselineskip plus -1fil
\begin{IEEEbiography}
[{\includegraphics[width=1in,height=1in,clip,keepaspectratio]{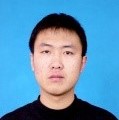}}]{Liang Sun}
received B.S. degree in computer science and technology from Xidian University, Xi¡¯an, China. He received double Dr. degree from Kochi University and Jilin University respectively, in 2012.
\par
He is currently with the College of Computer Science and Technology, Dalian University of Technology, Dalian, China. His main research interests include machine learning, swarm intelligence, deep learning and computer vision.
\end{IEEEbiography}
\vskip -2\baselineskip plus -1fil
\begin{IEEEbiography}
[{\includegraphics[width=1in,height=1in,clip,keepaspectratio]{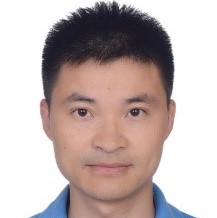}}]{Zhen Wang}
received Ph.D degree in school of Mathematical Sciences from Dalian University of Technology, Dalian, China, in 2008.
\par
He is currently a professor in School of Mathematical Sciences, Dalian University of Technology. His research interests include artificial intelligence, machine learning and computational methods.
\end{IEEEbiography}
\vskip -2\baselineskip plus -1fil
\begin{IEEEbiography}
[{\includegraphics[width=1in,height=1in,clip,keepaspectratio]{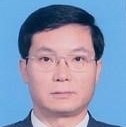}}]{Guozhen Tan}
received B.S. degree from the Shenyang University of Technology, Shenyang, China, and Ph.D. degree from the Dalian University of Technology, Dalian, China.
\par
He is currently a Professor of the College of Computer Science and Technology, Dalian University of Technology, Dalian, China. His research interests are cyber-physical systems and network optimization.
\end{IEEEbiography}
\vskip -2\baselineskip plus -1fil
\begin{IEEEbiography}
[{\includegraphics[width=1in,height=1in,clip,keepaspectratio]{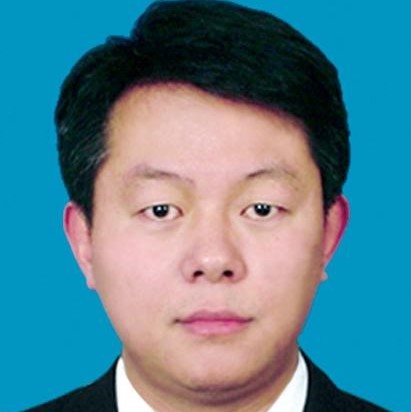}}]{Qiang Zhang}
received the B.S. degree in electronic engineering and M.S. and Ph.D. degrees in circuits and systems from the School of Electronic Engineering, Xidian University, Xi'an, China, in 1994, 1999, and 2002, respectively.
\par
He is currently a professor in the College of Computer Science and Technology, Dalian University of Technology, Dalian, China. His research interests are neural networks, artificial intelligence, DNA computing, optimization and intelligent robots.
\end{IEEEbiography}
\vskip -2\baselineskip plus -1fil
\begin{IEEEbiography}
[{\includegraphics[width=1in,height=1in,clip,keepaspectratio]{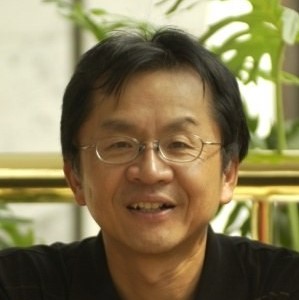}}]{C. L. Philip Chen}
received the M.S. degree in electrical engineering from the University of Michigan, Ann Arbor, in 1985, and Ph.D. degree in electrical engineering from Purdue University, West Lafayette, in 1988.
\par
He was a Tenured Professor, the Department Head, and an Associate Dean in two different universities in U.S. for 23 years. He is currently the Dean of the Faculty of Science and Technology, University of Macau, Macau, China, and a Chair Professor with the Department of Computer and Information Science. He was the President of the IEEE Systems, Man, and Cybernetics Society from 2012 to 2013. His current research interests include systems and cybernetics.
\end{IEEEbiography}

\end{document}